\pdfoutput=1

\documentclass[11pt]{article}

\usepackage[final]{acl}

\usepackage{times}
\usepackage{latexsym}

\usepackage[T1]{fontenc}

\usepackage[utf8]{inputenc}

\usepackage{microtype}

\usepackage{inconsolata}

\usepackage{graphicx}

\usepackage[utf8]{inputenc} 
\usepackage[T1]{fontenc}    
\usepackage{hyperref}       
\usepackage{url}            
\usepackage{booktabs}       
\usepackage{amsfonts}       
\usepackage{nicefrac}       
\usepackage{microtype}      
\usepackage{xcolor}         

\usepackage{caption}               
\usepackage{subcaption}
\usepackage{graphicx}
\usepackage{threeparttable}        
\usepackage{wrapfig}               
\usepackage{booktabs}
\usepackage{multirow}
\usepackage{xcolor}
\usepackage{bbding}                
\usepackage{lipsum}                
\usepackage[cjk]{kotex}            
\usepackage{amsmath}               
\usepackage{graphicx}              
\usepackage{accents}               
\usepackage{multirow}              
\usepackage{mathtools}             
\usepackage{footnote}
\usepackage{tablefootnote}
\usepackage{subfiles}
\usepackage{colortbl}              
\usepackage{xcolor}                
\usepackage{wrapfig}               
\usepackage{amssymb}               
\usepackage{comment}               
\usepackage{cleveref}              
\usepackage{pifont}
\usepackage{tocbibind}             
\usepackage{appendix}              
\usepackage{fancyhdr}

\definecolor{darkergreen}{RGB}{21, 152, 56}
\definecolor{red2}{RGB}{252, 54, 65}
\definecolor{Gray}{gray}{0.6}
\definecolor{LavenderBlush}{rgb}{1.0, 0.94, 0.96}

\newcommand{\yesmark}{\textcolor{darkergreen}{\ding{52}}}
\newcommand{\nomark}{\textcolor{red2}{\ding{56}}}
\newcommand{\gainp}[1]{\textcolor{teal}{$^{\texttt{(#1)}}$}}

\usepackage{titletoc}    

\def\addcontentsline#1#2#3{}

\hypersetup{
    colorlinks=true
}
\usepackage{url}
\usepackage{cuted}                 

\usepackage{tabularx}    

\title{3D-Aware Vision-Language Models Fine-Tuning\\with Geometric Distillation}

\author{
    Seonho~Lee\thanks{Equal contribution} ,~
    Jiho~Choi\footnotemark[1] ,~
    Inha~Kang,~
    Jiwook~Kim,~
    Junsung~Park,~
    Hyunjung~Shim\thanks{Corresponding author} \\
    Graduate School of Artificial Intelligence, KAIST, Republic of Korea \\
    {\tt\small {\{glanceyes, jihochoi, rkswlsj13, tom919, jshackist, kateshim\}@kaist.ac.kr}} \\
}

\begin{document}

\maketitle

\thispagestyle{fancy}
\fancyhf{}
\fancyfoot[L]{\footnotesize Published as a conference paper at Findings of EMNLP 2025.}
\renewcommand{\headrulewidth}{0pt} 
\renewcommand{\footrulewidth}{0pt} 

\begin{strip}
    \begin{minipage}{\textwidth}
        \centering
        \vspace{-6em}
        \includegraphics[width=0.95\linewidth]{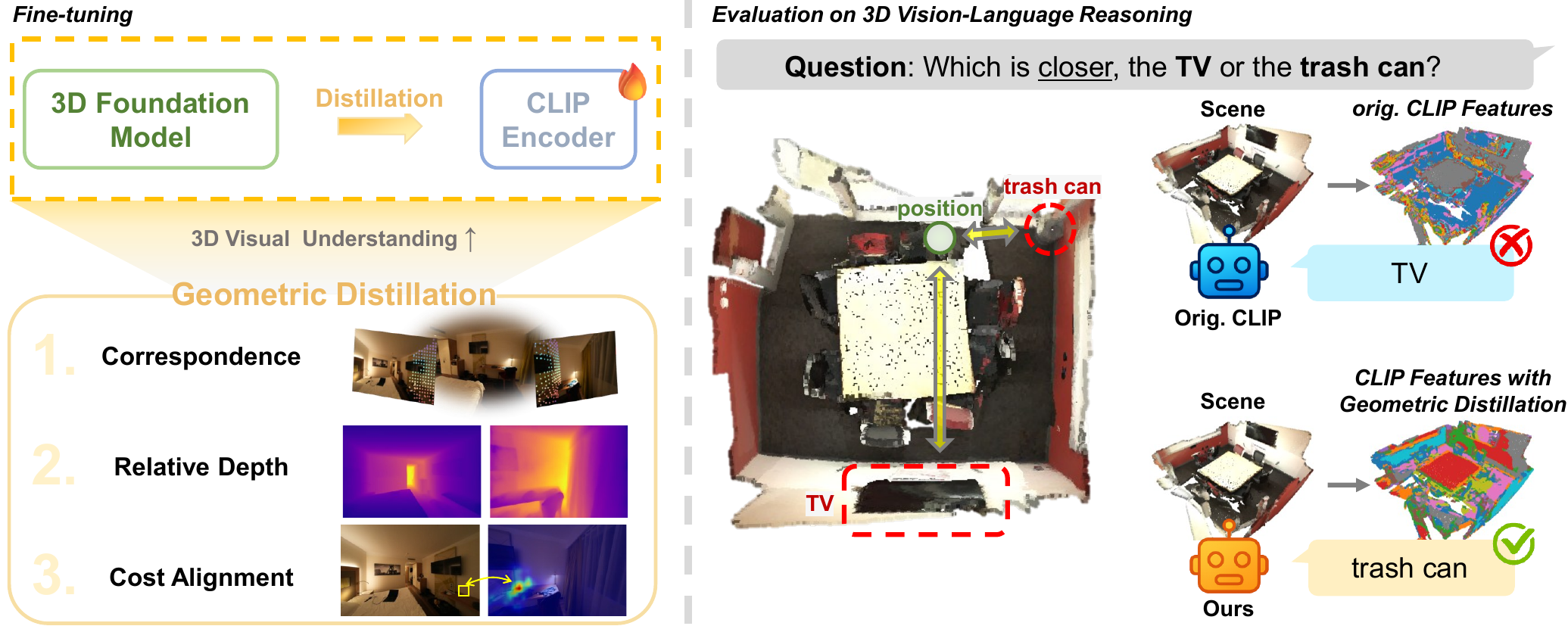}


        \vspace{-0.5em}
        \captionof{figure}{
            \textbf{Geometric Distillation enhances 3D spatial reasoning in vision-language models.} By distilling geometric cues such as correspondences, relative depth, and cost alignment from 3D foundation models, our method improves 3D visual understanding and enables accurate reasoning in tasks like answering which object is closer.
        }
        \label{fig:teaser}
        \vspace{-0.5em}
    \end{minipage}
\end{strip}

\begin{abstract}
    Vision-Language Models (VLMs) have shown remarkable performance on diverse visual and linguistic tasks, yet they remain fundamentally limited in their understanding of 3D spatial structures.
    We propose \textit{Geometric Distillation}, a lightweight, annotation-free fine-tuning framework that injects human-inspired geometric cues into pretrained VLMs without modifying their architecture.
    By distilling (1) sparse correspondences, (2) relative depth relations, and (3) dense cost volumes from off-the-shelf 3D foundation models (e.g., MASt3R, VGGT), our method shapes representations to be geometry-aware while remaining compatible with natural image–text inputs.
    Through extensive evaluations on 3D vision-language reasoning and 3D perception benchmarks, our method consistently outperforms prior approaches, achieving improved 3D spatial reasoning with significantly lower computational cost.
    Our work demonstrates a scalable and efficient path to bridge 2D-trained VLMs with 3D understanding, opening up wider use in spatially grounded multimodal tasks.
\end{abstract}

\vspace{-1em}
\begin{figure*}[ht]
    \centering

    \begin{subfigure}[t]{0.162\textwidth}
        \centering
        \includegraphics[width=\textwidth]{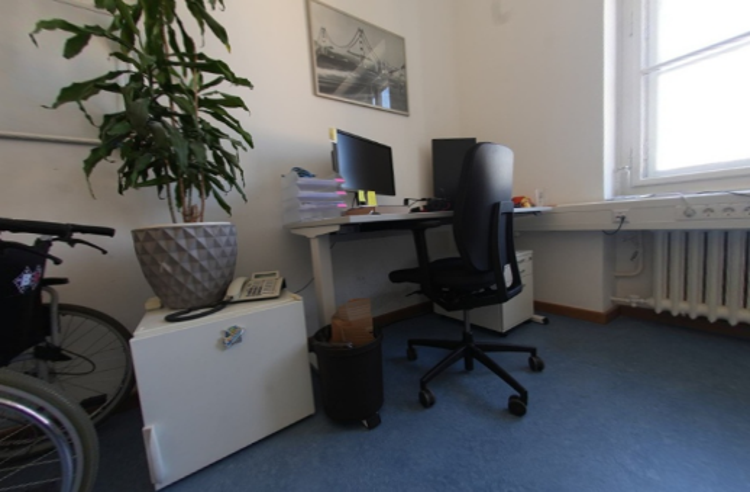}
        \captionsetup{skip=4pt}
    \end{subfigure}
    \begin{subfigure}[t]{0.162\textwidth}
        \centering
        \includegraphics[width=\textwidth]{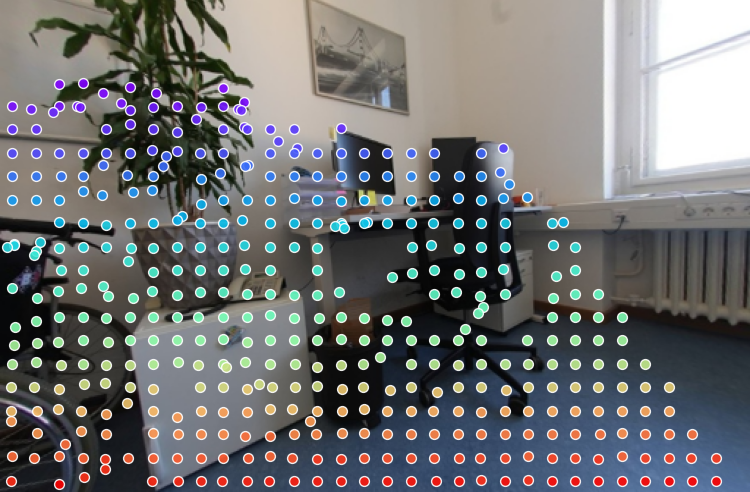}
        \captionsetup{skip=4pt}
    \end{subfigure}
    \begin{subfigure}[t]{0.162\textwidth}
        \centering
        \includegraphics[width=\textwidth]{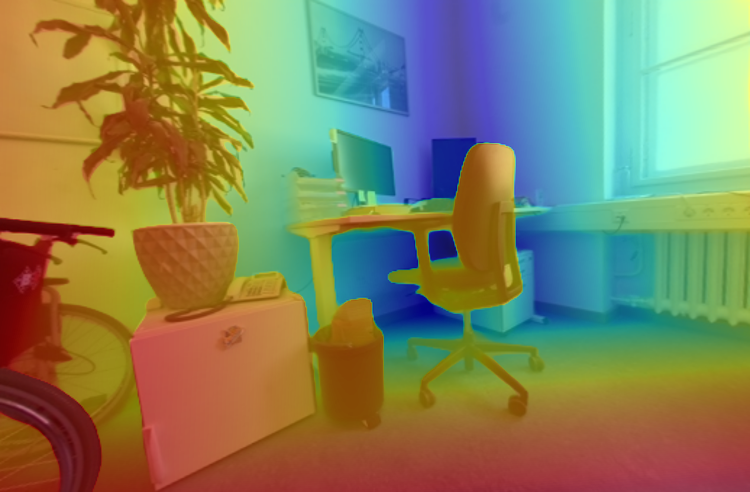}
        \captionsetup{skip=4pt}
    \end{subfigure}
    \begin{subfigure}[t]{0.162\textwidth}
        \centering
        \includegraphics[width=\textwidth]{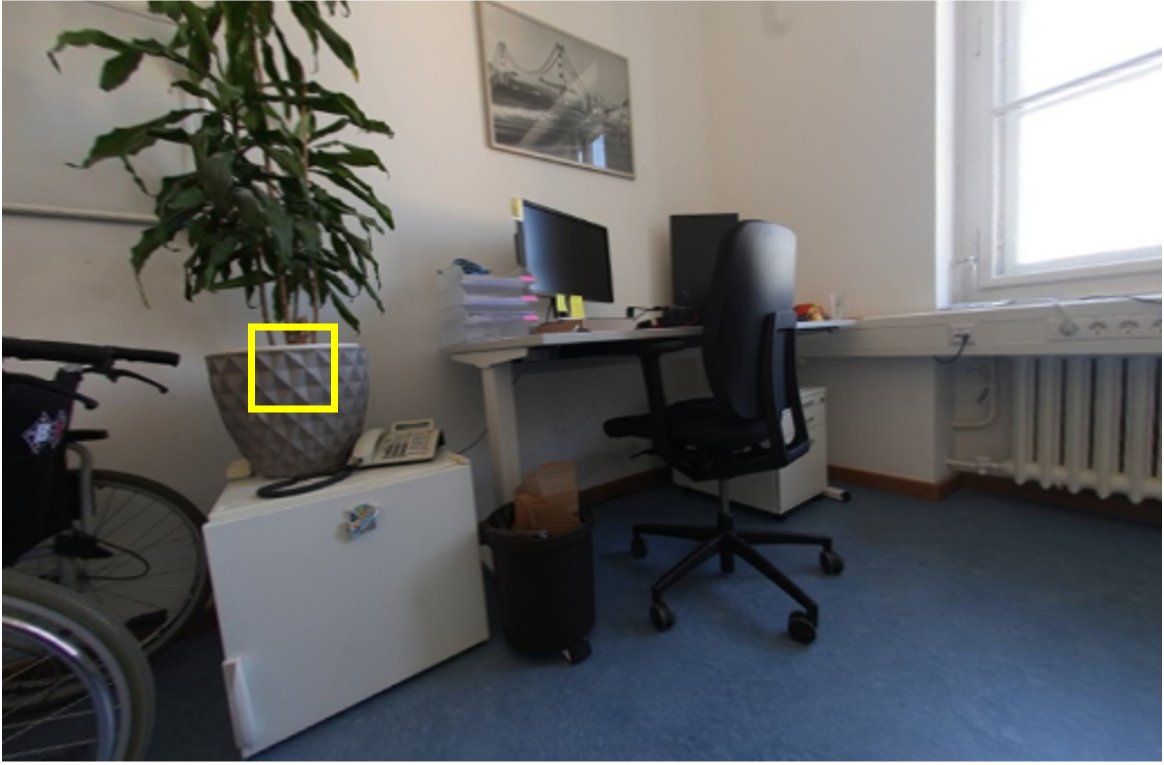}
        \captionsetup{skip=4pt}
    \end{subfigure}
    \begin{subfigure}[t]{0.156\textwidth}
        \centering
        \includegraphics[width=\textwidth]{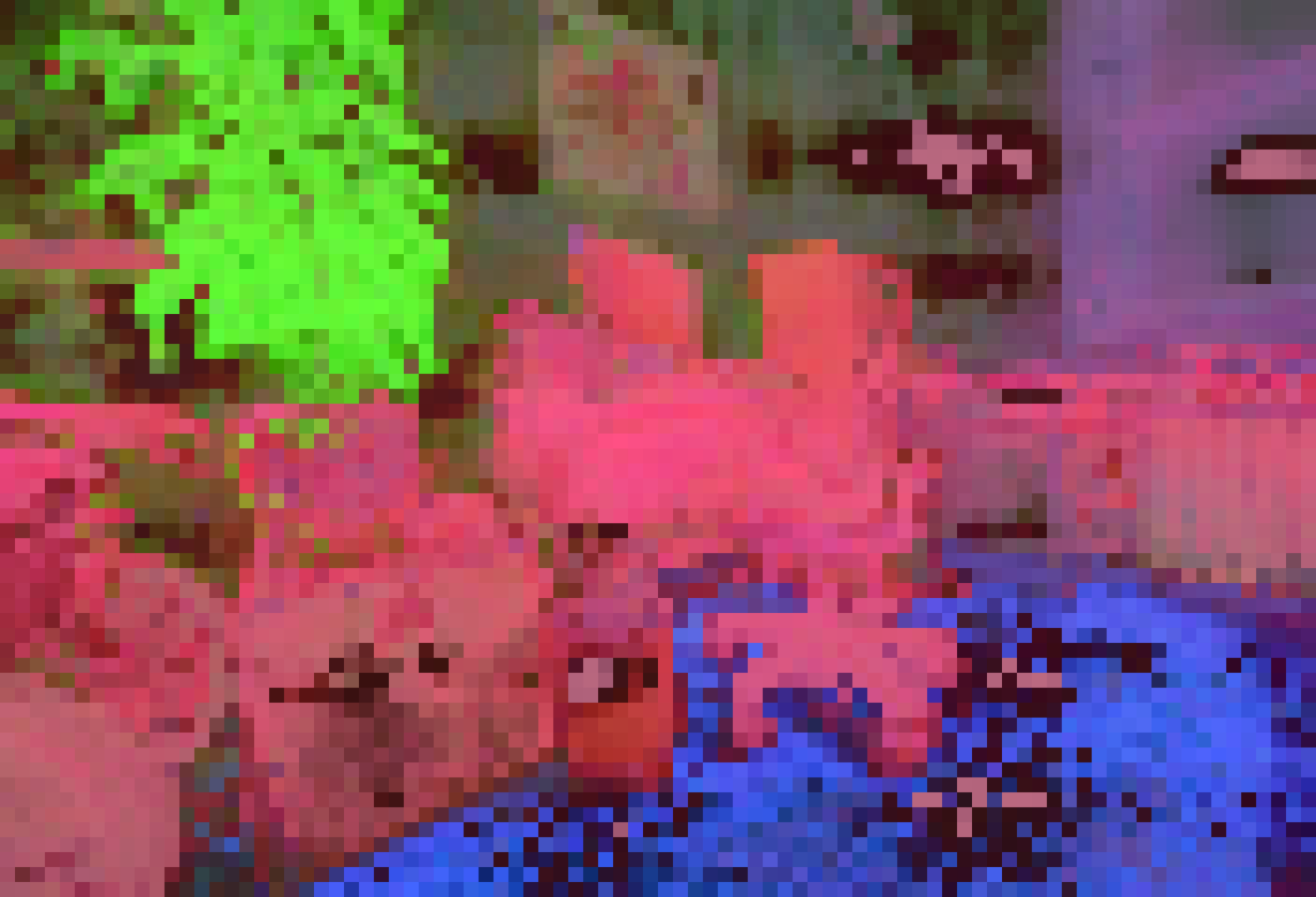}
        \captionsetup{skip=4pt}
    \end{subfigure}
    \begin{subfigure}[t]{0.156\textwidth}
        \centering
        \includegraphics[width=\textwidth]{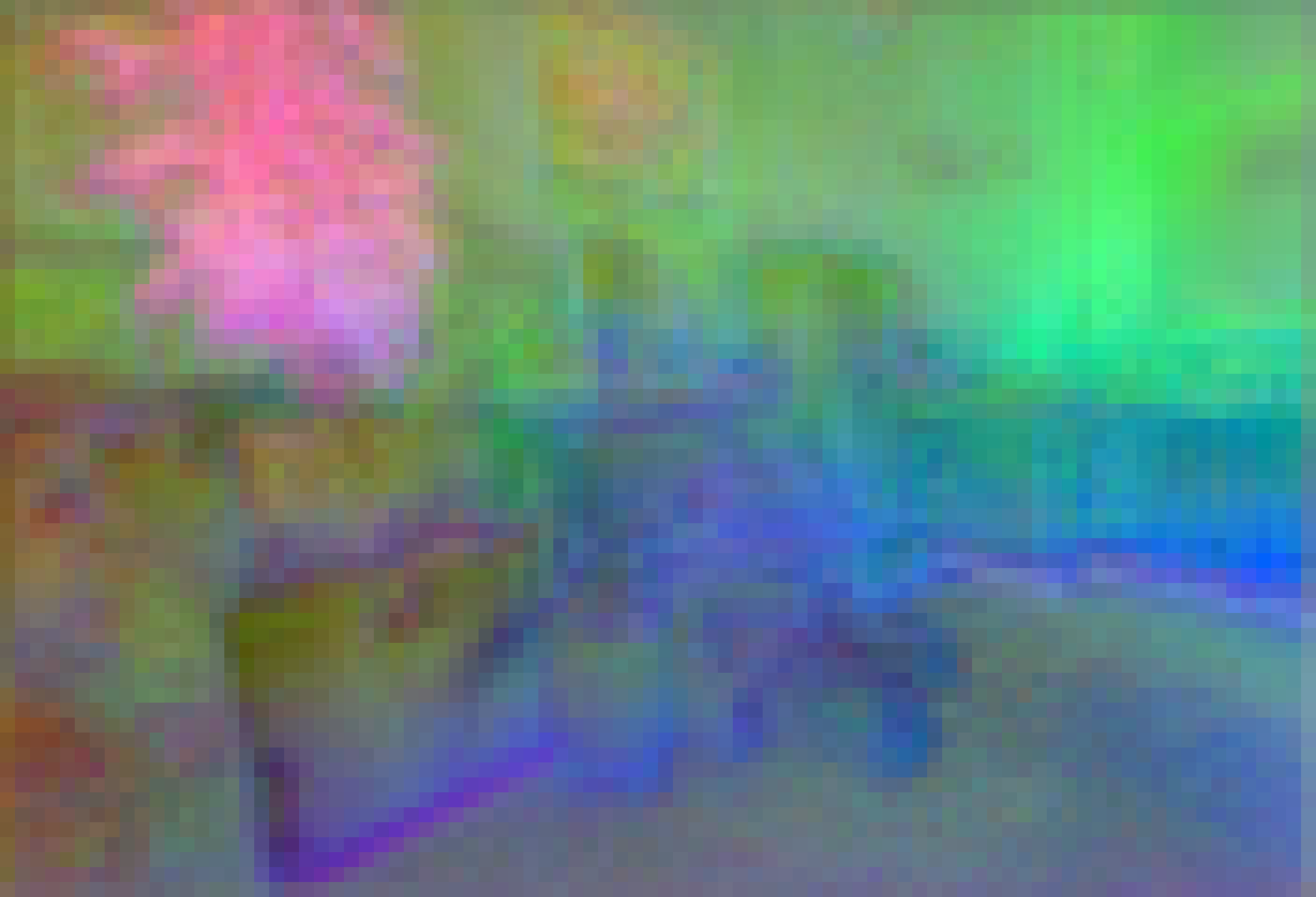}
        \captionsetup{skip=4pt}
    \end{subfigure}
    \\

    \begin{subfigure}[t]{0.162\textwidth}
        \centering
        \includegraphics[width=\textwidth]{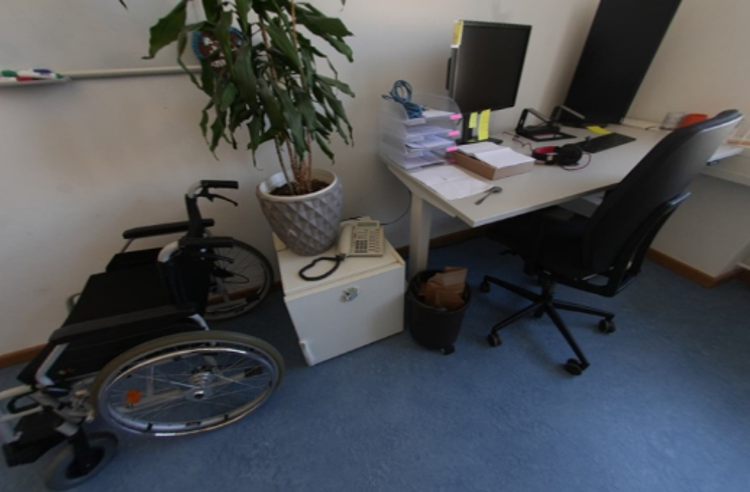}
        \captionsetup{skip=4pt}
        \caption{Multi-view}
    \end{subfigure}
    \begin{subfigure}[t]{0.162\textwidth}
        \centering
        \includegraphics[width=\textwidth]{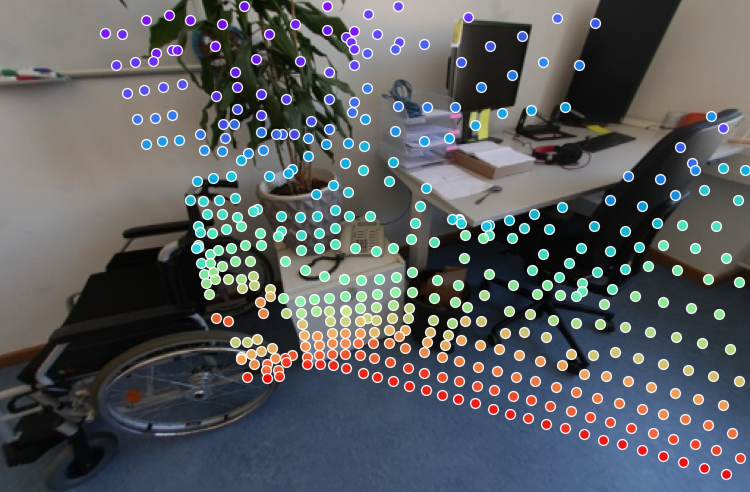}
        \captionsetup{skip=4pt}
        \caption{Correspondences}
    \end{subfigure}
    \begin{subfigure}[t]{0.162\textwidth}
        \centering
        \includegraphics[width=\textwidth]{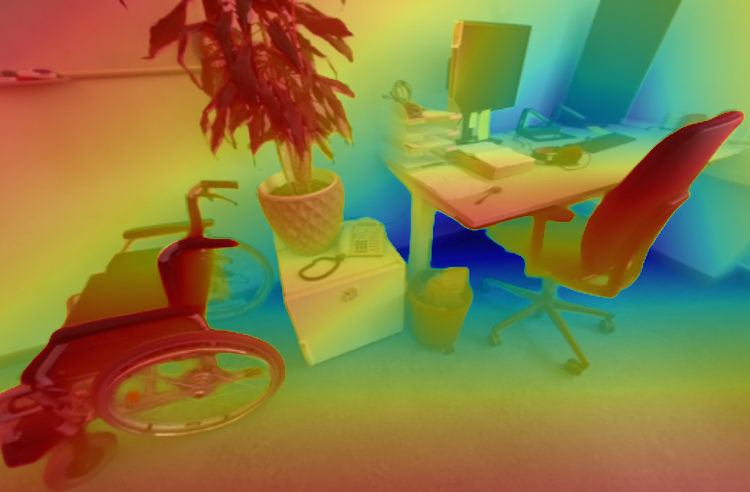}
        \captionsetup{skip=4pt}
        \caption{Depth Maps}
    \end{subfigure}
    \begin{subfigure}[t]{0.162\textwidth}
        \centering
        \includegraphics[width=\textwidth]{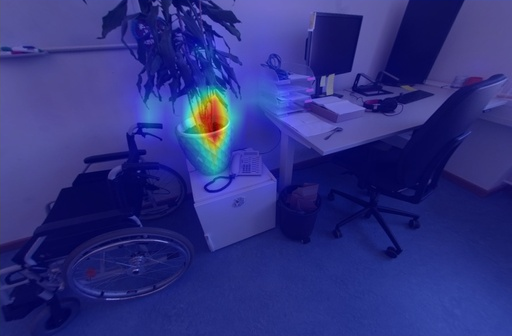}
        \captionsetup{skip=4pt}
        \caption{Cost Matching}
    \end{subfigure}
    \begin{subfigure}[t]{0.156\textwidth}
        \centering
        \includegraphics[width=\textwidth]{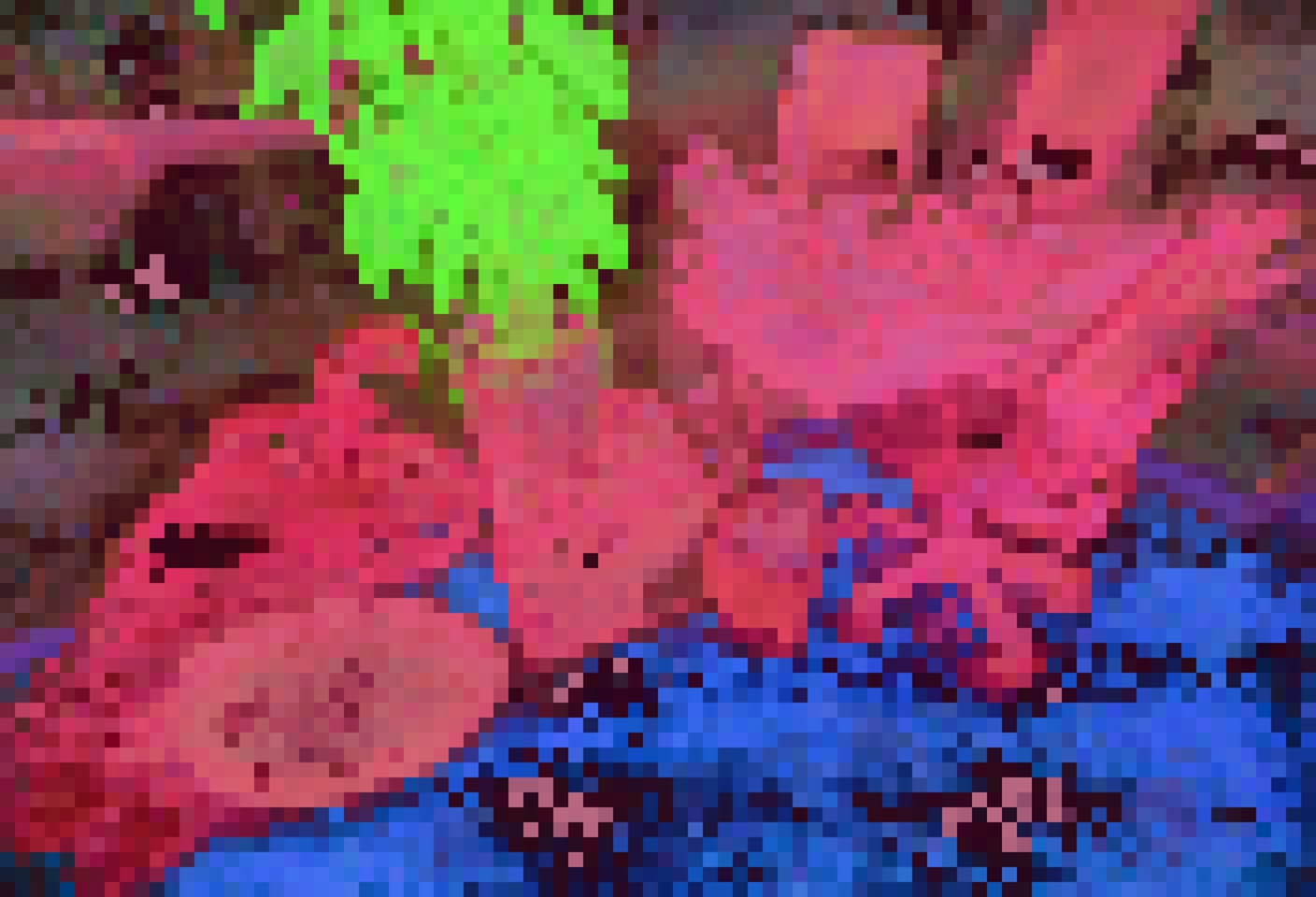}
        \captionsetup{skip=4pt}
        \caption{Before Tuning}
    \end{subfigure}
    \begin{subfigure}[t]{0.156\textwidth}
        \centering
        \includegraphics[width=\textwidth]{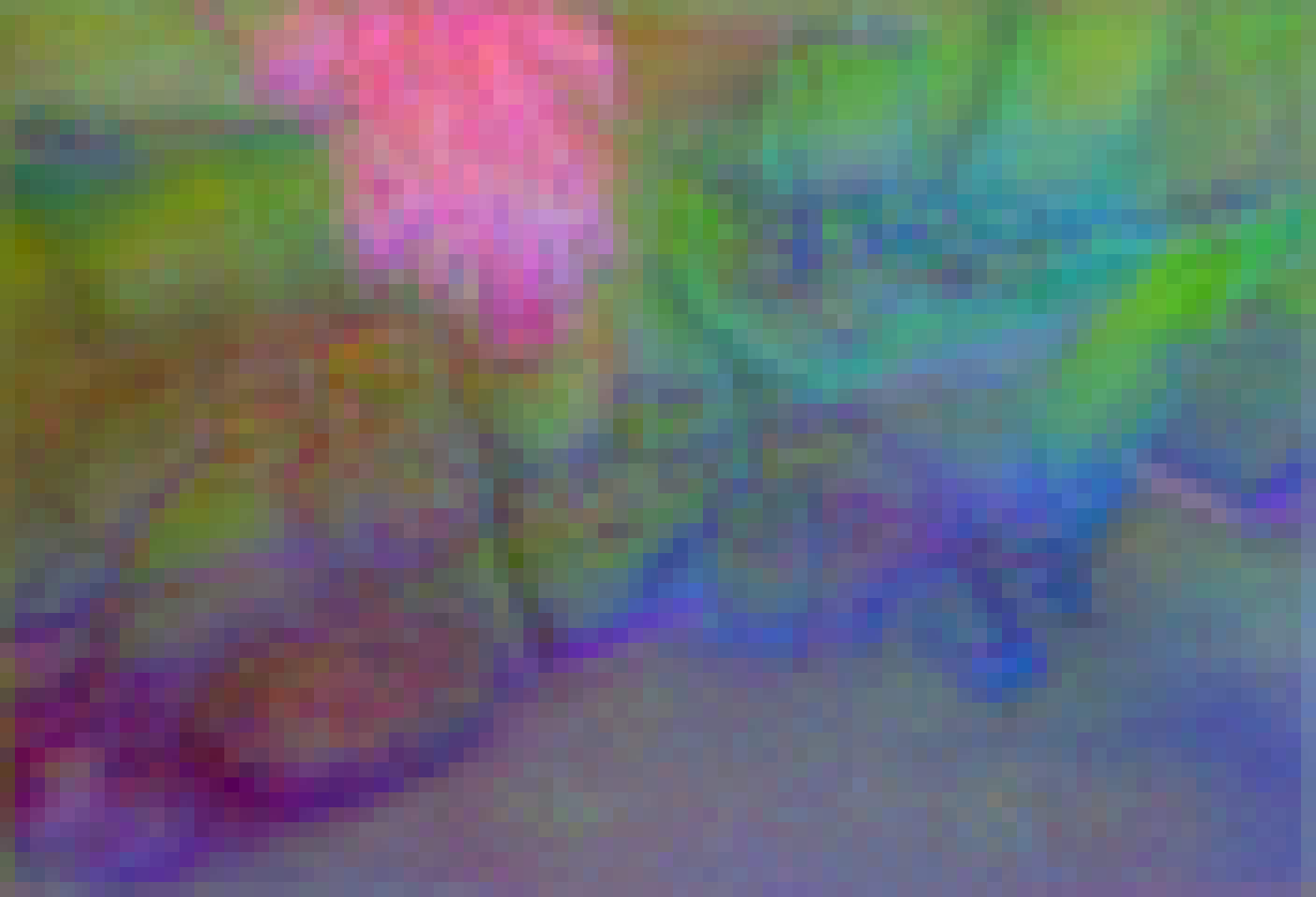}
        \captionsetup{skip=4pt}
        \caption{After Tuning}
    \end{subfigure}
    \\
    \vspace{-0.5em}
    \caption{
        Geometric cues and PCA visualization of feature transformation through geometric distillation.
    }
    \label{fig:geometric_cues}
    \vspace{-1.5em}
\end{figure*}

\section{Introduction}
\label{sec:introduction}

Vision-Language Models (VLMs) (e.g., CLIP~\cite{radford2021learning_CLIP}, ALIGN~\cite{jia2021scaling_ALIGN}, and BLIP~\cite{li2022blip, li2023blip_2}), trained on large-scale image-text datasets, have demonstrated competitive performance on diverse multimodal tasks~\cite{li2021align_ALBEF, gao2024clip_adapter, lee2022uniclip}.
Despite their progress, these models struggle with understanding 3D spatial structures~\cite{el2024probing3D, man2024lexicon3d, chen2024spatialvlm, danier2024_depthcues, li2024erroneous_aggreements, kamath2023s_whats_up, qiu2025refining_CLIP_VCP_spatial}.
Specifically, VLMs remain limited in grounded spatial reasoning tasks such as depth ordering, occlusion, or object layout in a scene~\cite{el2024probing3D, chen2024spatialvlm, kamath2023s_whats_up}.
This limitation stems from their reliance on 2D projections, which lack depth cues and multi-view supervision~\cite{eigen2014depth_ill_posed, tulsiani2017multi_ill_posed, qin2019monogrnet_geometric_info_loss}.
It is illustrated in~\Cref{fig:teaser}, where features of standard VLMs like CLIP incorrectly predict relative depth due to their limited 3D awareness.
These shortcomings greatly hinder applications requiring spatial reasoning, including navigation, scene understanding, and robotic planning~\cite{peng2023openscene, shridhar2022cliport, hong2023_3d_llm}.



To address this, recent work has explored injecting 3D priors into VLMs.
FiT3D~\cite{yue2024improving_FiT3D} reconstructs 3D scenes from multi-view images using Gaussian splatting~\cite{kerbl20233d_gs_gaussian_splatting}, and then aligns VLM's features with those rendered 3D views.
Multiview Equivariant Finetuning (MEF)~\cite{you2024multiview_ME} improves 3D equivariance by reinforcing feature consistency across rendered views of the same object.
SpatialVLM~\cite{chen2024spatialvlm} improves its spatial reasoning abilities by generating billions of synthetic spatial question–answer pairs to train VLMs.

Despite these advancements, existing methods suffer from notable drawbacks.
FiT3D incurs a high computational cost and suffers from semantic degradation due to its reliance on explicit 3D reconstruction. MEF depends on 3D object-centric datasets, which restricts its generalizability to real-world scenes.
SpatialVLM requires extensive synthetic data generation and task-specific tuning, making it resource-intensive and less flexible.
These limitations motivate the need for more efficient and generalizable approaches to endow VLMs with robust 3D awareness.

We propose \textit{Geometric Distillation}, a lightweight and annotation-free fine-tuning framework that enriches 3D spatial understanding in VLMs.
Our approach introduces supervision signals aligned with human perceptual strategies, derived from pretrained 3D foundation models such as MASt3R~\cite{leroy2024mast3r} and VGGT~\cite{wang2025_vggt} as in~\Cref{fig:geometric_cues} (a) - (d).
First, we supervise the VLM to align features at sparse correspondences that are visually stable and semantically meaningful regions, such as object corners or room boundaries, derived from pretrained 3D foundation models without any explicit 3D annotations.
These locations provide strong geometric anchors across views, and feature-level matching at these points encourages the model to learn consistent and viewpoint-invariant representations.
Second, we supervise relative depth reasoning through ordinal comparisons both within and across views.
This reflects the human tendency to reason in relative terms and aligns with the way spatial relationships are expressed in language~\cite{zhang2022can_DepthCLIP, auty2023_clip_MDE_depth_estimation}.
Lastly, we incorporate dense cost volume alignment, which captures soft correspondences across views by fully exploiting the geometric priors and warping relationships~\cite{weinzaepfel2022_croco, an2024cross_zeroco} provided by 3D foundation models, thereby enabling the model to learn fine-grained geometric consistency.
These signals collectively reshape the visual representations into a geometry-aware space that better supports grounded spatial reasoning and improves VLM performance on 3D-aware tasks as shown in~\Cref{fig:geometric_cues} (e), (f).
Additionally, since our approach operates without modifying the VLM's architecture and retains compatibility with natural image-text inputs, it preserves the strong generalization capabilities of the original model.

To overcome these limitations, we draw inspiration from human 3D spatial perception.
Humans infer depth and structure from sparse relational cues such as occlusions, relative size, and perspective, rather than absolute measurements~\cite{todd2003visual_human, howard1995binocular_human, landy1995measurement_human}.
In addition, spatial relationships are often expressed in language using relative terms (e.g., ``next to the table'', ``behind the sofa'') rather than absolute metric units, suggesting that the reasoning is both perceptually and linguistically grounded.
These observations suggest that incorporating human-inspired geometric cues into VLM can enhance their spatial reasoning abilities.

Our approach enhances the model’s ability to infer spatial relations, such as object proximity, without explicit 3D labels or costly reconstruction.
We demonstrate consistent improvements across a range of 3D-aware tasks, including semantic correspondence, depth estimation, and 3D visual question answering.
Our method outperforms strong baselines on benchmarks such as PF-PASCAL, TAP-Vid, and ScanQA, illustrating both the effectiveness and scalability of our approach.

\begin{color}{green}


\end{color}

\section{Related Work}
\label{sec:related_work}


\noindent \textbf{Fine-tuning VFMs and VLMs.}
Various attempts have been made to integrate 3D information into Visual Foundation Models (VFMs) or Vision-Language Models (VLMs)~\cite{yue2024improving_FiT3D, you2024multiview_ME}.
FiT3D~\cite{yue2024improving_FiT3D} lifts 2D visual features into a 3D Gaussian representation and re-renders them from multiple viewpoints.
By fine-tuning, this approach guides the original 2D features to align with re-rendered features, which enhances 3D perception in VFMs.
However, its dense L1 optimization introduces noise, which potentially leads to semantic information loss and significant computational overhead.
Multiview Equivariance Finetuning (MEF)~\cite{you2024multiview_ME} enhances 3D correspondence understanding by maximizing cross-view feature equivariance within pretrained foundation models.
This allows them to improve on tasks such as camera pose estimation, object tracking, and semantic transfer.
Nevertheless, MEF requires explicit 3D annotations and does not provide direct supervision for depth understanding.
SpatialVLM~\cite{chen2024spatialvlm} generates extensive 3D spatial QA corpora using pretrained detectors, depth estimators, and segmentation models.
Training on this large-scale data strengthens the spatial question-answering capabilities of VLMs.
However, the reliance on massive synthetic datasets limits their practicality.
In our work, we address these limitations by introducing a lightweight and annotation-free fine-tuning method that efficiently enhances 3D spatial reasoning in VLMs.

\noindent \textbf{3D Foundation Models.}
Recently, geometry-based models have emerged as foundation models for 3D vision. 
CroCo~\cite{weinzaepfel2022_croco, weinzaepfel2023_croco_v2} performs self-supervised cross-view completion by reconstructing one view from another, which allows the model to acquire multi-view consistent features. 
Based on CroCo pretraining, DUSt3R~\cite{wang2024_dust3r} introduces a unified approach to directly estimate scene point maps from two or more images taken from different viewpoints. 
DUSt3R effectively simplifies the Structure-from-Motion (SfM) pipeline.
MASt3R~\cite{leroy2024mast3r} further extends these approaches by incorporating a global matching head that aligns partial reconstructions and predicts dense 3D correspondences.
These models inherently provide 3D perceptual priors by learning scene geometry without explicit supervision or accurate dense reconstructions from limited views.
Additionally, VGGT~\cite{wang2025_vggt} introduces a large transformer-based model to jointly estimate camera poses, depth maps, and point clouds from a few images. 
Training VGGT on large-scale 3D datasets enables accurate depth prediction even from a single image, which significantly improves 3D downstream tasks. 
Consequently, these models embed critical 3D knowledge that is beneficial for robust 3D understanding.
In our work, we propose a method to effectively inject these rich 3D priors into VLMs.

\begin{figure*}
    \centering
    \includegraphics[width=0.93\linewidth]{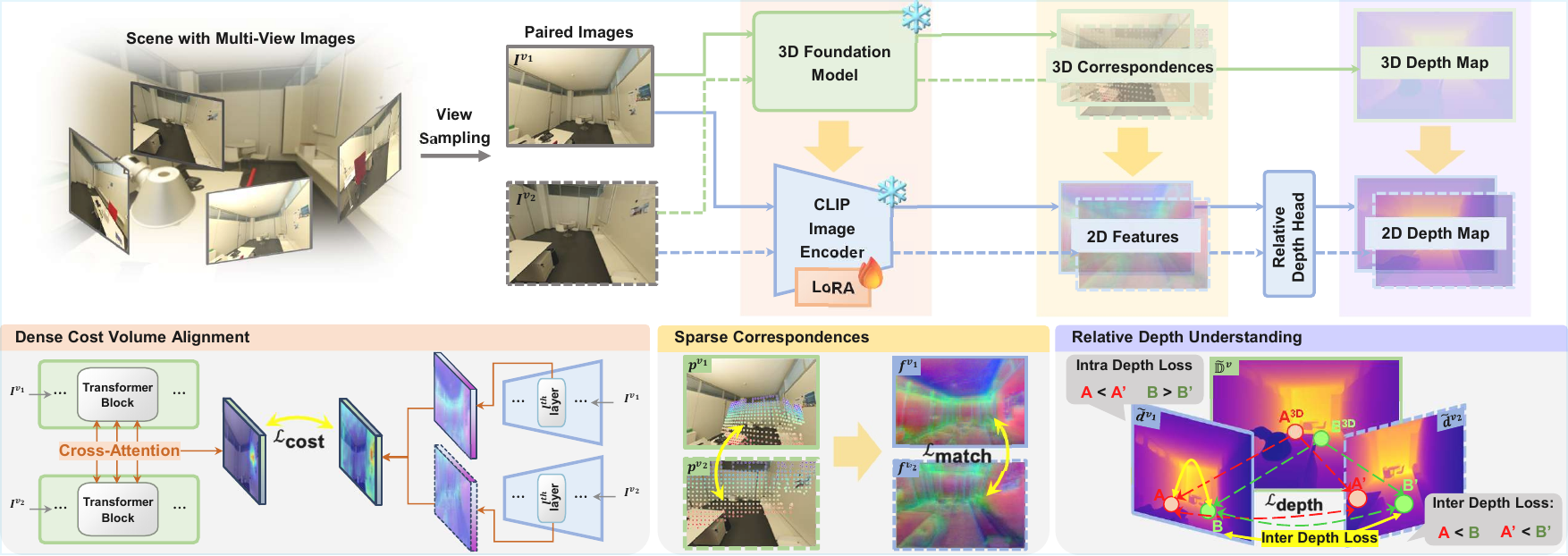}
    \vspace{-.8em}
    \caption{
        \textbf{Overview of Geometric Distillation Architecture.}
        A 3D foundation model extracts geometric cues including (1) sparse correspondences, (2) depth maps, and (3) dense cost volumes from multi-view inputs.
        These cues supervise a frozen CLIP image encoder with a lightweight adapter (LoRA) via three loss branches: $\mathcal{L}_\text{match}$, $\mathcal{L}_\text{depth}$, and $\mathcal{L}_\text{cost}$.
        The distillation enables the VLM to acquire 3D spatial awareness without explicit 3D annotations.
    }
    \label{fig:architecture}
    \vspace{-1.0em}
\end{figure*}


\noindent \textbf{Bridging VLMs and 3D Understanding.}
Recent studies have explored analyzing or improving vision-language representations to better understand 3D scenes.
Lexicon3D~\cite{man2024lexicon3d} evaluates various vision foundation encoders across vision–language reasoning tasks and identifies their strengths and limitations.  
Notably, image-text alignment supervised models~\cite{qiu2025refining_CLIP_VCP_spatial, auty2023_clip_MDE_depth_estimation, radford2021learning_CLIP, jia2021scaling_ALIGN} still exhibit substantial weaknesses in complex 3D spatial reasoning and language-driven question answering tasks.  
This suggests that vision–language pretraining alone may not sufficiently capture comprehensive 3D concepts. 
These observations underscore the necessity of incorporating explicit 3D signals or specialized training strategies into VLMs.  
To address these limitations, various approaches have been proposed.
Some studies~\cite{hegde2023clip} extend CLIP via prompt tuning by prepending learnable tokens to the vision encoder and training it contrastively on rendered 3D object images paired with textual labels.
Other notable efforts include PointCLIP~\cite{zhang2022pointclip, zhu2023pointclip}, which aligns 3D point clouds with CLIP's textual embedding space, and methods designed to enhance text–image alignment in 3D contexts~\cite{kim2023extending, zeng2021multi_MG_VLP}.
Collectively, these studies introduce additional representations or strategies to enrich 3D understanding within VLMs.
In contrast, our work directly injects robust 3D knowledge into 2D VLMs using multi-view images.
This enables leveraging their inherent rich 2D vision-language priors without relying on explicit supervision from other 3D data modalities such as point clouds or 3D Gaussians.


\section{Proposed Method}
\label{Proposed Method: Geometric Distillation}

We propose a geometric knowledge distillation framework that transfers 3D spatial understanding from high-performance 3D foundation models such as MASt3R~\cite{leroy2024mast3r} and VGGT~\cite{wang2025_vggt} into a pretrained vision-language model (VLM)~\cite{radford2021learning_CLIP,jia2021scaling_ALIGN} without requiring any ground truth 3D annotations.
Inspired by human perception, which infers spatial structure by integrating visual cues from multiple viewpoints, our method uses paired images, $\{I^{v_1}, I^{v_2}\}$, of the same scene captured from different perspectives $v_1$ and $v_2$.
From these image pairs, we extract geometric signals including sparse correspondences, ordinal depth relations, and viewpoint-induced disparities, which guide the VLM to learn geometry-aware representations.
An overview of our framework is illustrated in \Cref{fig:architecture}.

Our framework obtains these geometric cues using a teacher model that generates pseudo-3D supervision from image pairs.
Specifically, we utilize the following information provided by 3D foundation models: (i) sparse correspondences $\mathbb{P}^{{v_1}, {v_2}} = \{(p_i^{v_1}, p_i^{v_2})\}_{i=0}^{|\mathbb{P}^{{v_1}, {v_2}}|}$ for matching 3D points across views, (ii) estimated depth maps $\tilde{\mathbb{D}}^{v_1}, \tilde{\mathbb{D}}^{v_2}$ for each viewpoint, and (iii) a dense cost volume, $\mathbb{C}^{v_1 \to v_2}$, representing patch-level features similarity between two viewpoints.
These heterogeneous signals serve as supervision for three complementary objectives: sparse correspondence matching, relative depth learning using both intra-view and inter-view comparisons, and alignment of dense feature similarity.
Combined, they enrich the model’s multimodal representations and facilitate 3D-aware reasoning in complex scenes.

\subsection{Sparse Correspondences}
\label{sec:Sparse_Correspondences}

\textbf{Background.}
Humans often rely on sparse but stable visual features, such as corners or edges, to estimate spatial layout.
In a similar way, sparse correspondences across views serve as geometric anchors that help enforce cross-view consistency and identify matching 3D points.
These signals are essential for enforcing consistency across viewpoints~\cite{leroy2024mast3r, wang2025_vggt} and have been widely adopted in multi-view geometry~\cite{weinzaepfel2022_croco,weinzaepfel2023_croco_v2,an2024cross_zeroco} as well as recent representation learning methods such as MEF~\cite{you2024multiview_ME}.
To exploit these correspondences, we adopt a feature-matching objective that promotes accurate feature-level alignment between image pairs.
Given a set of pseudo correspondence pairs $\mathbb{P}^{{v_1}, {v_2}} = \{(p_i^{v_1}, p_i^{v_2})\}_{i=1}^{|\mathbb{P}^{{v_1}, {v_2}}|}$ generated by a geometric teacher, we extract local image features $\{(f_i^{v_1}, f_i^{v_2})\}_{i=1}^{|\mathbb{P}^{{v_1}, {v_2}}|}$ and intermediate patch features $\{h^{v_*}\}$ from each viewpoint.
We adopted a matching-based loss~\cite{brown2020smoothap, you2024multiview_ME} that encourages high retrieval performance by maximizing the Smooth Average Precision (SmoothAP)~\cite{brown2020smoothap}, computed within a spatial neighborhood.
For a query feature $f_i$, the SmoothAP is calculated using positive matches $\mathbb{P}^{{v_1}, {v_2}}$ and negative matches (non-matches), $\mathcal{N}(i)$, of point $p_i$ as:

{\small
\vspace{-1.5em}
\begin{align}
    \textstyle
    \mathrm{SmoothAP}_{\scriptstyle v_1 \to v_2} = \nonumber \\  
    \frac{1}{|\mathbb{P}^{v_1, v_2}|}
    \sum_{i \in \mathbb{P}^{v_1, v_2}}
    & \frac{1 + \sigma(D_{ii})}
         {1 + \sigma(D_{ii}) + 
         \sum_{j \in \mathcal{N}(i)} \sigma(D_{ij})},
\end{align}
}where $D_{ij} = f_j^{v_2} \cdot f_i^{v_1} - f_i^{v_1} \cdot f_i^{v_1}$ measures the difference in similarity between features, and $\sigma(x)$ denotes the sigmoid function.
This objective promotes higher similarity for true matches than for non-matches, thereby incorporating relative similarity into the training. To ensure symmetry across views, we apply the objective in both matching directions and define the final loss as:

{\small
\vspace{-1.0em}
\begin{equation}
    \textstyle
    \mathcal{L}_{\texttt{match}}
    =
    1-\frac{1}{2}
    \{
    \mathrm{SmoothAP}_{\scriptstyle v_1 \to v_2}
    +\mathrm{SmoothAP}_{\scriptstyle v_2 \to v_1}
    \}
    .
\end{equation}
}

\begin{color}{red}

\end{color}

\begin{color}{gray}

\end{color}

\subsection{Relative Depth Understanding}
\label{subsec:Relative_Depth_Understanding}

To complement sparse correspondences, we enhance the VLM’s geometric reasoning by supervising its understanding of relative depth.
Unlike absolute depth estimation, which is fundamentally ambiguous in monocular settings due to scale uncertainty, relative depth reasoning (i.e., determining which of two points is closer) is intuitive and practically robust across domains~\cite{todd2003visual_human, howard1995binocular_human, landy1995measurement_human}.
Numerous studies~\cite{fu2018deep_ordinal, chen2016single_relative_depth, xian2020structure_depth, zoran2015learning_depth} show that models trained with ordinal depth constraints generalize better to diverse scenes and produce sharper depth maps with preserved structure.

Inspired by this, we leverage the outputs of high-capacity 3D foundation models (e.g., MASt3R~\cite{leroy2024mast3r}, VGGT~\cite{wang2025_vggt}) to construct pseudo ground-truth relative depth labels.
This approach allows us to inject 3D awareness into VLMs without explicit 3D supervision or reconstruction.
The learning proceeds on two levels: \textit{intra-view} and \textit{inter-view}, capturing both local monocular cues and multi-view disparities, akin to human depth perception mechanisms.

\noindent \textbf{Intra-view Relative Depth.}
Given an image $I^v$, we sample point pairs $(x, y) \in \mathcal{P}^{v}$ and determine their ordinal pseudo ground-truth relation using the depth map $\tilde{\mathbb{D}}^v$ provided by a 3D foundation model (e.g., MASt3R, VGGT).
The relative depth ordering is defined as:
\begin{equation}
    \mathrm{s}_{xy} = \operatorname{sign}(\tilde{d}_x - \tilde{d}_y) \in \{-1, +1\},
\end{equation}
where $\tilde{d}_x$ and $\tilde{d}_y$ denote the estimated depths of points $x$ and $y$ from viewpoint $v$, respectively.
The VLM predicts a scalar depth ranking score $\hat{\mathrm{s}}_{x y}$ for each pair based on its encoded features, and is trained with a logistic ranking loss~\cite{chen2009ranking_loss, fu2018deep_ordinal}:

{\small
\vspace{-1.0em}
\begin{equation}
    \mathcal{L}_{\texttt{intra\_depth}} =
    \frac{1}{|\mathcal{P}^{v}|^2}
    \sum_{(x, y) \in \mathcal{P}^{v}}
    \log \left(1 + \exp \left[ -{\mathrm{s}}_{xy} \cdot \hat{\mathrm{s}}_{xy} \right] \right).
\end{equation}
} This loss encourages correct ordinal predictions without relying on metric depth values, allowing the model to learn scale-invariant depth cues from local monocular structure.

\begin{figure}[t!]
    \centering
    \subcaptionbox{Anchor}[0.24\linewidth]{%
      \includegraphics[width=\linewidth]{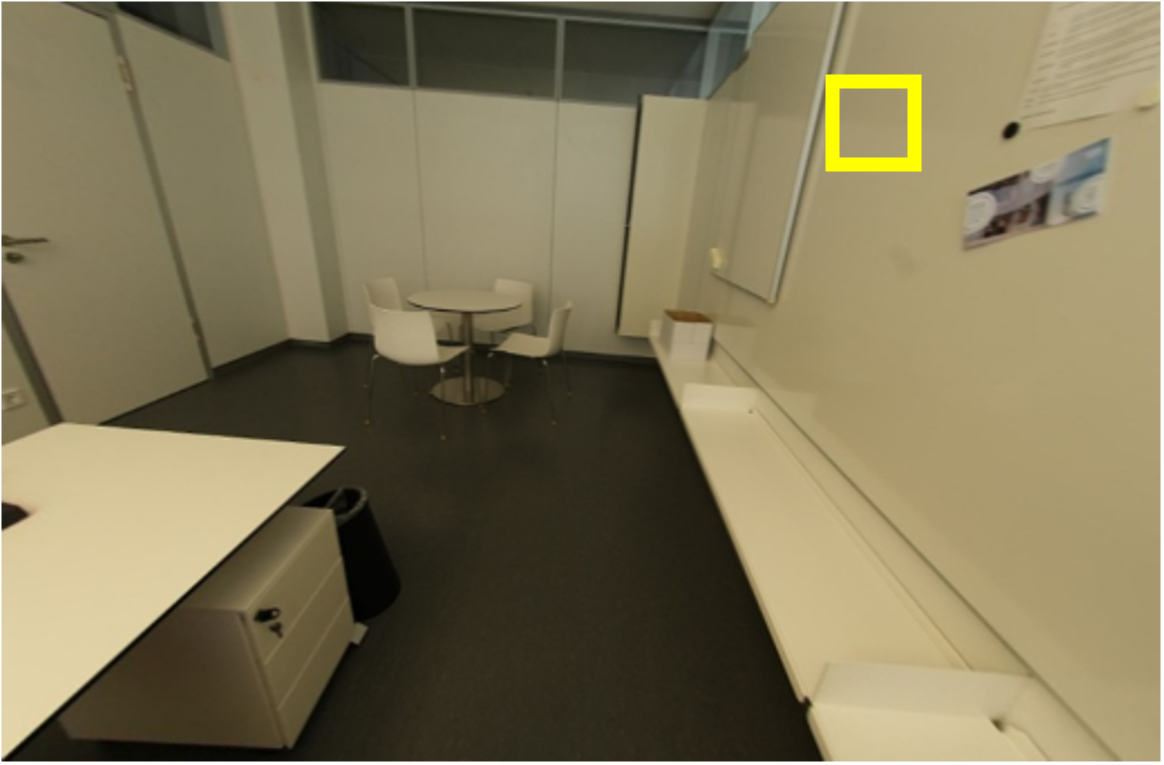}\par}
    \hspace{0.1pt}%
    \subcaptionbox{MASt3R}[0.24\linewidth]{%
      \includegraphics[width=\linewidth]{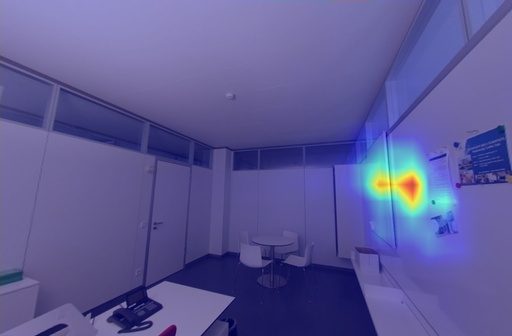}\par}
    \hspace{0.1pt}%
    \subcaptionbox{Vanilla}[0.24\linewidth]{%
      \includegraphics[width=\linewidth]{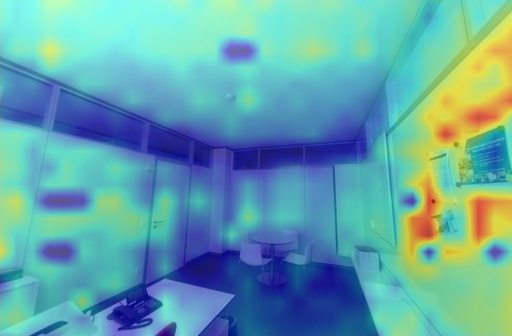}\par}
    \hspace{0.1pt}%
    \subcaptionbox{Ours}[0.24\linewidth]{%
      \includegraphics[width=\linewidth]{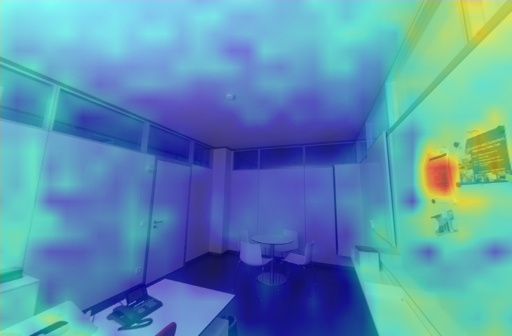}}
    \vspace{-0.6em}
    \caption{
        \textbf{Visualization of cost volume}.
        (a) Anchor view with query location (yellow box). Cost volume heatmaps from (b) the teacher (MASt3R), (c) the vanilla CLIP, and (d) after geometric distillation. The proposed method better captures localized geometric similarity, closely aligning with the teacher’s output.
    }
    \label{fig:cost_volume_alignment}
    \vspace{-1.5em}
\end{figure}

\noindent \textbf{Inter-view Relative Depth.}
To further infuse geometric awareness, we supervise relative depth relationships across multiple views, as absolute depth values may differ due to scale variations between viewpoints.
Unlike intra-view supervision, which assumes a consistent scale within a single image, inter-view supervision requires the model to reason about depth differences under potential scale shifts.

Given a correspondence pair $(p_i^{v_1}, p_i^{v_2}) \in \mathbb{P}^{{v_1}, {v_2}}$ that observes the same 3D point from views $v_1$ and $v_2$, we extract the pseudo ground-truth depths $\tilde{d}_i^{v_1}$ and $\tilde{d}_i^{v_2}$ from the teacher model's depth maps $\tilde{\mathbb{D}}^{v_1}$ and $\tilde{\mathbb{D}}^{v_2}$, respectively.
To mitigate the effect of absolute scale mismatch, we define a bounded signed depth difference using the $\tanh$ function as $\delta^*_i = \tanh(\tilde{d}_i^{v_1} - \tilde{d}_i^{v_2})$.
The model is trained to regress this value using a lightweight MLP head, which is applied to the feature representations of each view.
The loss is defined as:
\begin{equation}
    \mathcal{L}^{v_1, v_2}_{\texttt{inter\_depth}}
    =
    \frac{1}{|\mathbb{P}^{{v_1}, {v_2}}|}
    \sum_{i \in \mathbb{P}^{{v_1}, {v_2}}}
    \left| \hat{\delta}_i - \delta^*_i \right|.
\end{equation}

This supervision encourages the model to be sensitive to viewpoint-induced disparities and relative geometry, even in the absence of explicit camera calibration or metric consistency.
To jointly capture both local (intra-view) and cross-view (inter-view) depth relationships, we define the final relative depth loss as a combination of the two components: $\mathcal{L}_{\texttt{depth}} =
    \sum_{p} \{ \mathcal{L}_{\texttt{intra\_depth}}^{v_p} + 
    \sum_{q} \mathcal{L}_{\texttt{inter\_depth}}^{v_p, v_q} \}$.
By unifying intra-view ordinal supervision with inter-view relative regression, the model learns to infer consistent and structurally-aware depth relationships. This multi-scale depth reasoning framework fosters a more human-like, scale-invariant understanding of 3D geometry, enhancing the generalization ability of vision-language models across diverse visual domains.

\begin{table*}[t!]
    \caption{Comparison of zero-shot semantic correspondence on PF-PASCAL.}
    \label{tab:main_semantic_corr}
    \vspace{-1.0em}
    \centering
    \begin{small}
        \scalebox{0.85} {
            \begin{threeparttable}
            \begin{tabular}{@{}lccccccccc@{}}
                \toprule
                \multicolumn{1}{c}{\multirow{2}{*}{\textbf{Method}}} & \multicolumn{1}{c}{\multirow{2}{*}{\textbf{Dataset}}} & \multicolumn{3}{c}{\textbf{Different Views}} & \multicolumn{3}{c}{\textbf{Same Views}} \\ \cmidrule(l){3-5} \cmidrule(l){6-8} 
                &  & PCK@0.05 & PCK@0.10 & PCK@0.15 & PCK@0.05 & PCK@0.10 & PCK@0.15  \\ \midrule
                \multirow{1}{*}{(Vanilla) CLIP} & -- &  16.61  &   26.96  &   37.64  &  18.23  &  32.27  &  43.01 \\
                \multirow{1}{*}{FiT3D~\cite{yue2024improving_FiT3D}} & ScanNet++                 &  15.90 &   23.40  &  30.34  &  14.93  &  26.52  &  34.56 \\
                MEF~\cite{you2024multiview_ME} & Objaverse          &  21.18  &  33.54  &  43.58  &  25.94 &  43.33  &  53.87 \\
                \midrule
                Ours & Objaverse                                                    & \underline{25.87}   &  \underline{39.85}   &  \underline{50.21}  &  \underline{36.77}  &  \underline{56.61}  &  \underline{67.93} \\
                Ours & ScanNet++                                                    &  \textbf{28.48}  &  \textbf{43.07}  &  \textbf{53.55}  &  \textbf{42.16}  & \textbf{61.57}  &  \textbf{72.16} \\
                \vspace{-4pt} & & \gainp{+11.87} & \gainp{+16.11} &  \gainp{+15.91}  & \gainp{+23.93} & \gainp{+29.30} & \gainp{+29.15} \\
                
                \bottomrule
            \end{tabular}
            \begin{tablenotes}
                \item[1] The best score is \textbf{bold} and the second-best score is \underline{underlined}. These are the same for all experiments.
            \end{tablenotes}
            \end{threeparttable}
        }
    \end{small}
    \vspace{-0.5em}
\end{table*}

\begin{table*}[t!]
    \caption{Comparison of video tracking on TAP-Vid and pose estimation on OnePose-LowTexture.}
    \label{tab:main_video_track_and_pose_est}
    \vspace{-1.0em}
    \centering
    \begin{small}
        \scalebox{0.85} {
            \begin{threeparttable}
            \begin{tabular}{@{}lcccccccc@{}}
                \toprule
                \multicolumn{1}{c}{\multirow{2}{*}{\textbf{Method}}} & \multicolumn{1}{c}{\multirow{2}{*}{\textbf{Dataset}}} & \multicolumn{2}{c}{\textbf{Video Tracking}} & \multicolumn{3}{c}{\textbf{Pose Estimation}} \\ \cmidrule(l){3-4} \cmidrule(l){5-7} 
                &  & Avg. Jaccard Index & Avg. Position Accuracy & 1cm-1deg & 3cm-3deg & 5cm-5deg  \\ \midrule
                \multirow{1}{*}{(Vanilla) CLIP} & -- &  27.73  &   42.59  &  2.50  &  19.32  &  33.11 \\
                \multirow{1}{*}{FiT3D \cite{yue2024improving_FiT3D}} & ScanNet++                 &  28.45  &   43.51  &  2.86  &  20.14 &  34.75 \\
                MEF~\cite{you2024multiview_ME} & Objaverse          &  34.61  &  50.58  &  6.32 &  36.00  &  52.33 \\
                \midrule
                Ours & Objaverse                                                    & \underline{35.60}   &  \underline{54.65}  &  \underline{8.50}  &  \underline{39.30}  &  \underline{57.68} \\
                Ours & ScanNet++                                                    &  \textbf{40.09}   &  \textbf{57.75}  &  \textbf{10.96}  & \textbf{44.93}  &  \textbf{63.65} \\
                \vspace{-4pt} &  & \gainp{+12.36} &  \gainp{+15.16}  & \gainp{+8.46}  & \gainp{+25.61} & \gainp{+30.54} \\
                
                \bottomrule
            \end{tabular}
            \end{threeparttable}
        }
    \end{small}
    \vspace{-1.0em}
\end{table*}

\subsection{Dense Cost Volume Alignment}
\label{subsec:Dense_Cost_Volume_Alignment}

Beyond sparse matching and relative depth supervision, we introduce a dense cost volume alignment method to extract richer geometric cues from intermediate features of 3D foundation models.
This alignment is further enhanced by leveraging geometric priors from cross-view completion models such as CroCo~\cite{weinzaepfel2022_croco, weinzaepfel2023_croco_v2}, and transformer-based models using cross-attention mechanisms across multiple views like VGGT~\cite{wang2025_vggt}.
Recent findings from ZeroCo~\cite{an2024cross_zeroco} show that cross-attention maps learned through cross-view completion pretext tasks encode high-quality dense correspondences, effectively acting as self-supervised cost volumes.
These maps inherently learn to warp source features to reconstruct masked target views by estimating correspondences across views. 
By treating these attention-derived correspondences as pseudo ground-truth warping functions, we can supervise the VLM's dense feature similarity to better reflect geometric consistency, thereby enhancing its capacity for dense 3D-aware reasoning.

To enforce dense geometric consistency across entire feature maps, we align the feature similarities produced by a vision-language model with geometrically grounded predictions from a 3D foundation model as in~\Cref{fig:cost_volume_alignment}.
Given two views $v_1$ and $v_2$, we construct a 4D cost volume that encodes normalized feature similarity between all spatial positions (patch index) across the views:
{
\vspace{-0.5em}
\begin{equation}
    \mathbb{C}_{\scriptstyle v_1 \to v_2}(i,j)
    =
    \frac
        {h^{v_1}_i \cdot h^{v_2}_j}
        {\lVert h^{v_1}_{i} \rVert \lVert h^{v_2}_{j} \rVert},
\end{equation}
}where $h^{v_*}_i \in \mathbb{R}$ denotes the intermediate feature vector at patch index $i$ in view $v_1$, and $j$ is a corresponding patch index in view $v_2$.
This similarity matrix captures the VLM's inherent geometric understanding between all patch pairs across views. We convert this cost volume into a probability distribution using temperature-scaled softmax as:
\begin{equation}
    P_{\scriptstyle v_1 \to v_2}(j\mid i)
      = \mathrm{softmax}_{j} \bigl( \mathbb{C}_{\scriptstyle v_1 \to v_2}(i,j) / \tau \bigr),
\label{eq:softmax_annealing_cost}
\end{equation}
where temperature $\tau$ controls the sharpness of the matching distribution. The geometric teacher provides target distributions $\tilde{P}_{\scriptstyle v_1 \to v_2}$ derived from its robust 3D understanding. Our alignment loss minimizes the Jensen-Shannon Divergence~\cite{menendez1997jensen_shannon_divergence} as:

{\small
\vspace{-1.0em}
\begin{equation}
    \textstyle
    \mathcal{L}_{\texttt{cost}}
    =\frac{1}{2}
    \{
        D_{\mathrm{KL}} \bigl( \tilde{P}_{\scriptstyle v_1 ^{\to} v_2} \Vert P_{\scriptstyle v_1 ^{\to} v_2} \bigr)
        + D_{\mathrm{KL}} \bigl( \tilde{P}_{\scriptstyle v_2 ^{\to} v_1} \Vert P_{\scriptstyle v_2 ^{\to} v_1} \bigr)
    \}.
\end{equation}}
This dense supervision compels the VLM's feature similarities to mirror the teacher's geometrically grounded predictions, enforcing subpixel-level geometric awareness.


\subsection{Overall Objective}
\label{subsec:overall_objective}

To jointly train the vision-language model with rich geometric supervision, we combine the proposed loss components into a single objective function.
Given a pair of images $(I^{v_1}, I^{v_2})$ from the same scene, the total loss is defined as:
\begin{equation}
\textstyle
\mathcal{L}_{\texttt{total}} = 
    \lambda_{\texttt{match}} \mathcal{L}_{\texttt{match}} + 
    \lambda_{\texttt{depth}} \mathcal{L}_{\texttt{depth}} +
    \lambda_{\texttt{cost}} \mathcal{L}_{\texttt{cost}}.
\end{equation}
where $\lambda_{\texttt{match}}$, $\lambda_{\texttt{depth}}$, and $\lambda_{\texttt{cost}}$ are hyperparameters for balancing each loss term.

\section{Experiments}
\label{sec:experiments}

\subsection{Experimental Setups}
\label{subsec:experimetnal_setups}


\textbf{Datasets.}
We evaluate our method in two main sets of downstream tasks to examine the effectiveness of our 3D-aware VLM representations: 3D visual understanding and vision-language understanding tasks. Specifically, to measure the 3D correspondence understanding, we conduct experiments on three downstream benchmarks introduced by~\cite{you2024multiview_ME}: (1) semantic correspondence on PF-PASCAL~\cite{ham2016proposal_pf_pascal}, (2) video tracking on TAP-Vid~\cite{doersch2022tap}, and (3) object pose estimation on the OnePose-LowTexture dataset~\cite{he2022onepose++}. Additionally, we perform experiments on downstream tasks for dense scene understanding via linear probing as in FiT3D~\cite{yue2024improving_FiT3D}, including semantic segmentation on ADE20K~\cite{zhou2019semantic_ade20k} and VOC2012~\cite{everingham2015pascal_voc2012}, and monocular depth estimation on ScanNet++~\cite{yeshwanth2023scannet++} and KITTI~\cite{geiger2013vision_kitti}. Furthermore, we assess improvements in 3D vision-language understanding by evaluating our method on the 3D visual question-answering benchmarks SQA3D~\cite{ma2022sqa3d} and ScanQA~\cite{azuma2022scanqa}. 

\noindent \textbf{Implementation Details.}
We fine-tune the ViT-based CLIP model for up to 500 epochs on either Objaverse~\cite{deitke2023objaverse} or ScanNet++.
We perform parameter-efficient fine-tuning through LoRA~\cite{hu2022lora}, adopting settings similar to those used in MEF~\cite{you2024multiview_ME}. Our method primarily leverages MASt3R~\cite{leroy2024mast3r} as a pretrained 3D foundation teacher during geometric distillation.
Further implementation details, including experiments with VGGT~\cite{wang2025_vggt}, are provided in the appendix.

\subsection{Experimental Results}
\label{subsec:experimental_results}

\newcommand{\zoomwidecrop}[1]{%
  \includegraphics[width=\linewidth,
    trim=12pt 96pt 12pt 96pt,clip]{#1}}

\begin{figure}[t!]
  \centering
  \begin{minipage}[t]{0.49\textwidth}
    \centering
    \subcaptionbox{Source}[0.322\linewidth]{%
      \zoomwidecrop{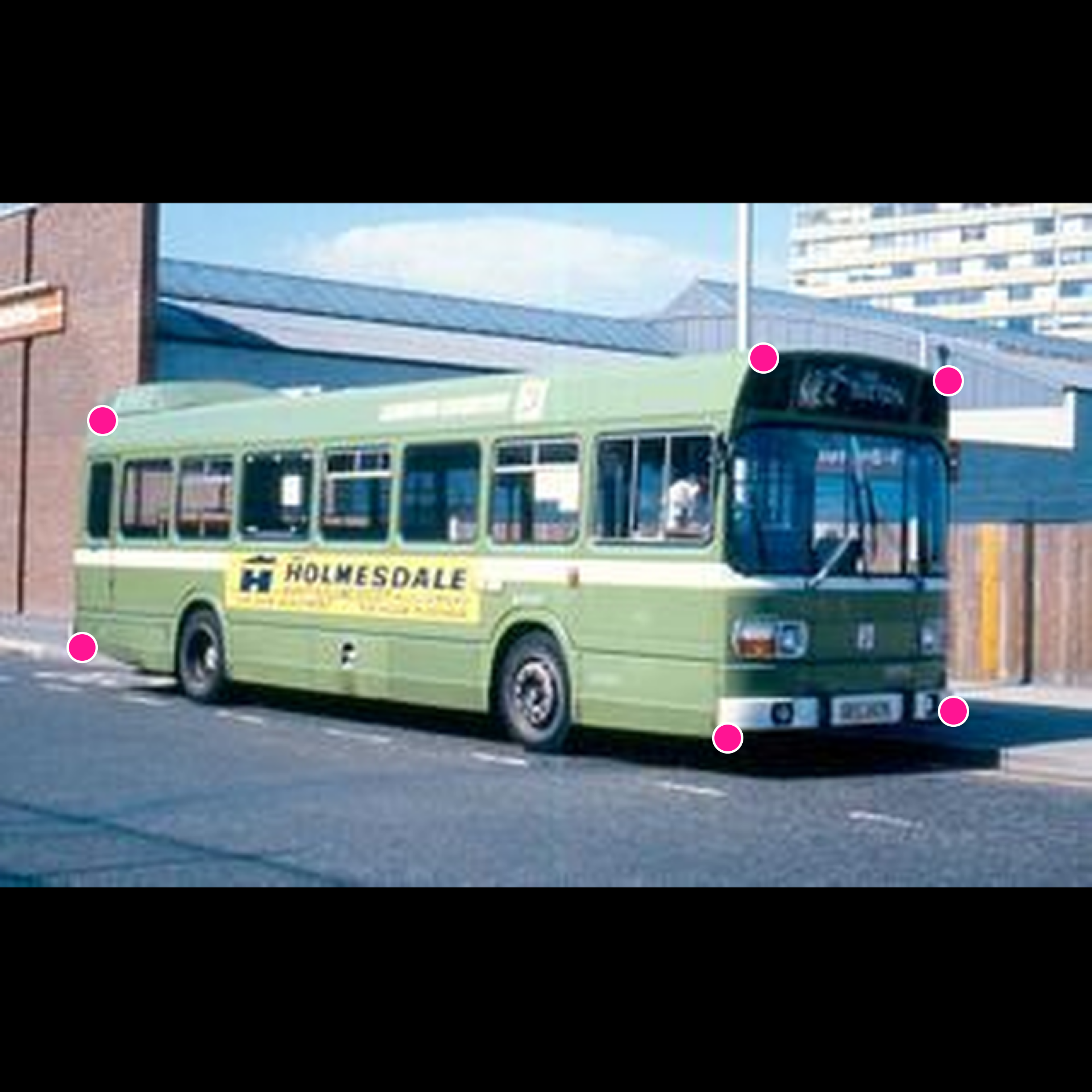}\par
      \vspace{2pt}
      \zoomwidecrop{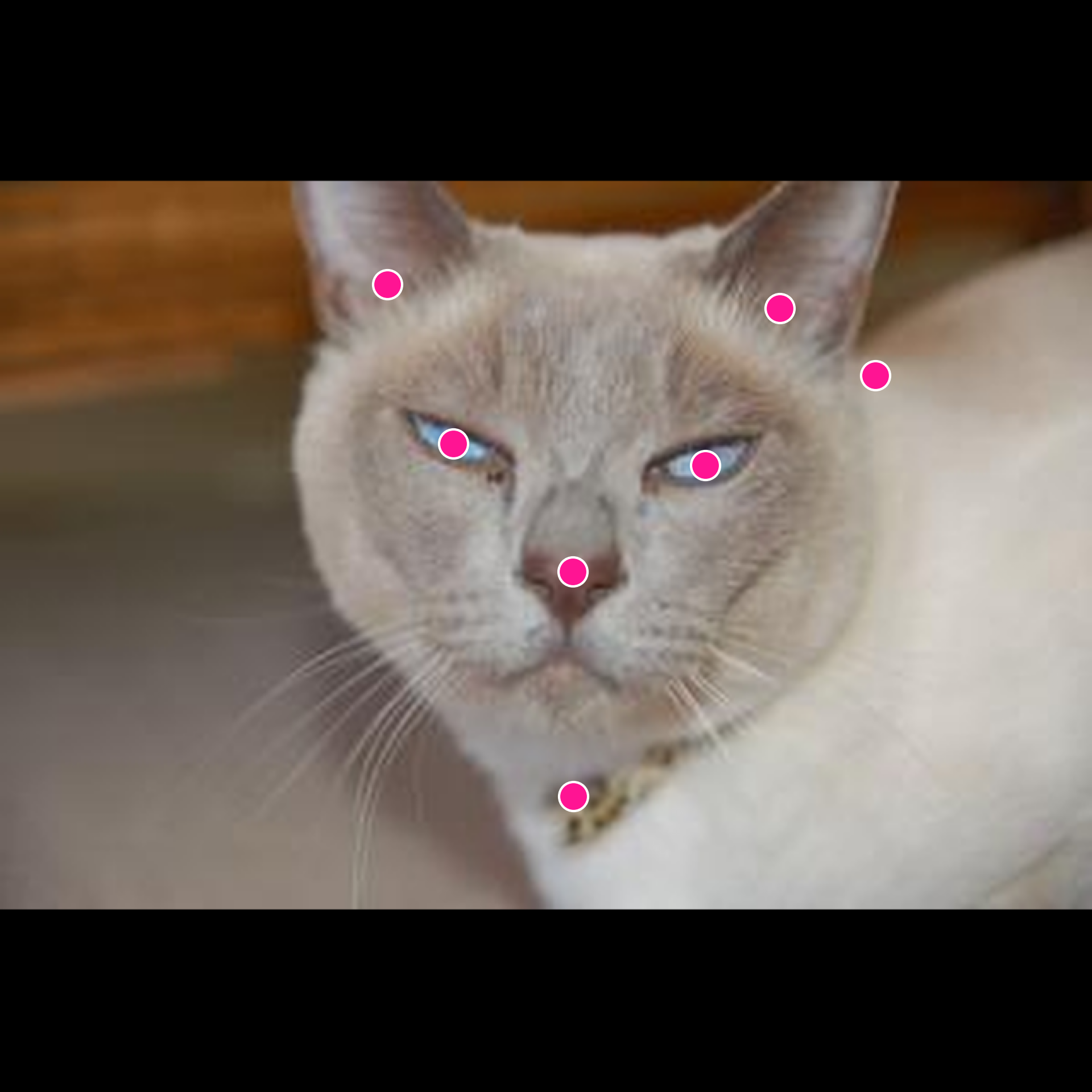}}
    \hspace{0.25pt}%
    \subcaptionbox{MEF}[0.322\linewidth]{%
      \zoomwidecrop{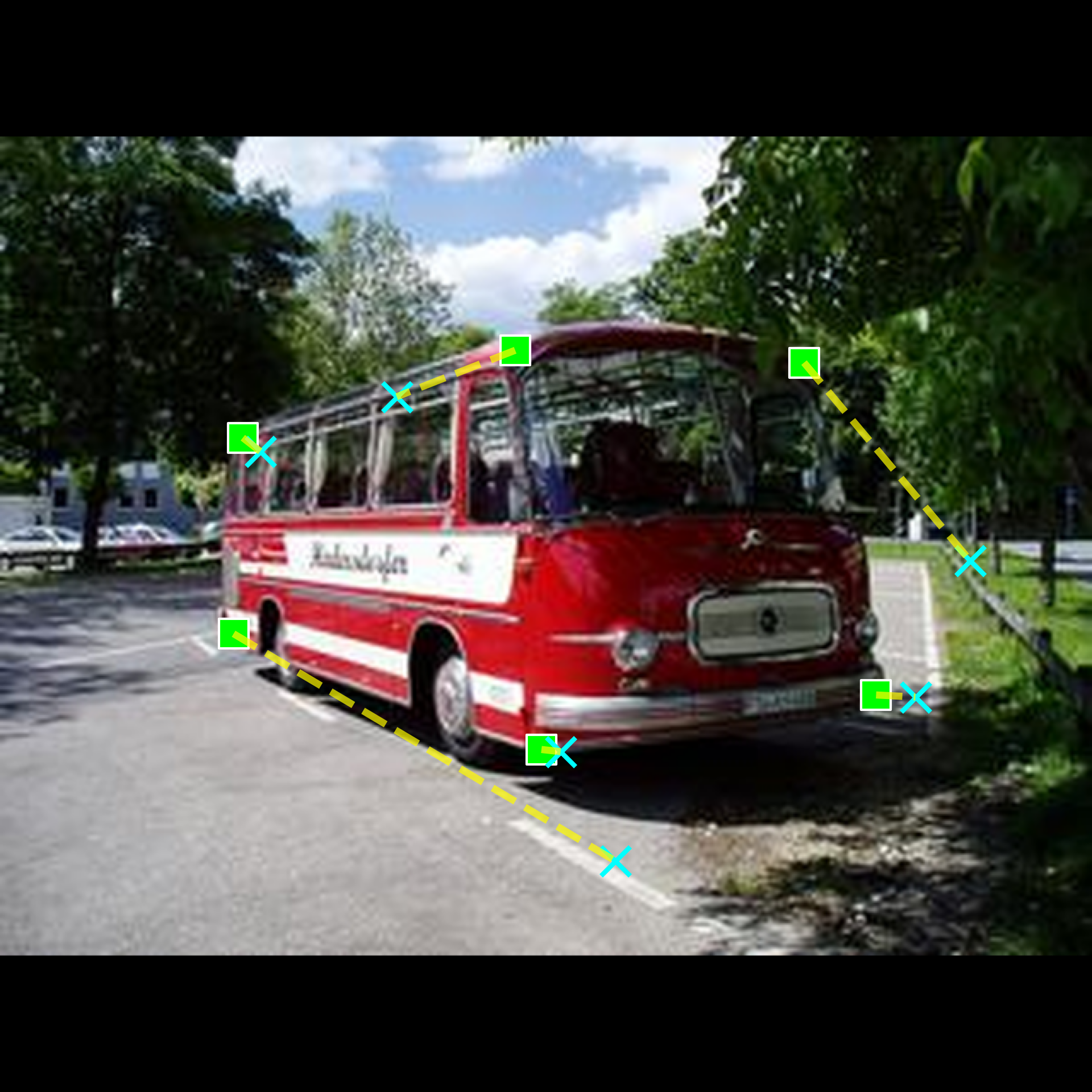}\par
      \vspace{2pt}
      \zoomwidecrop{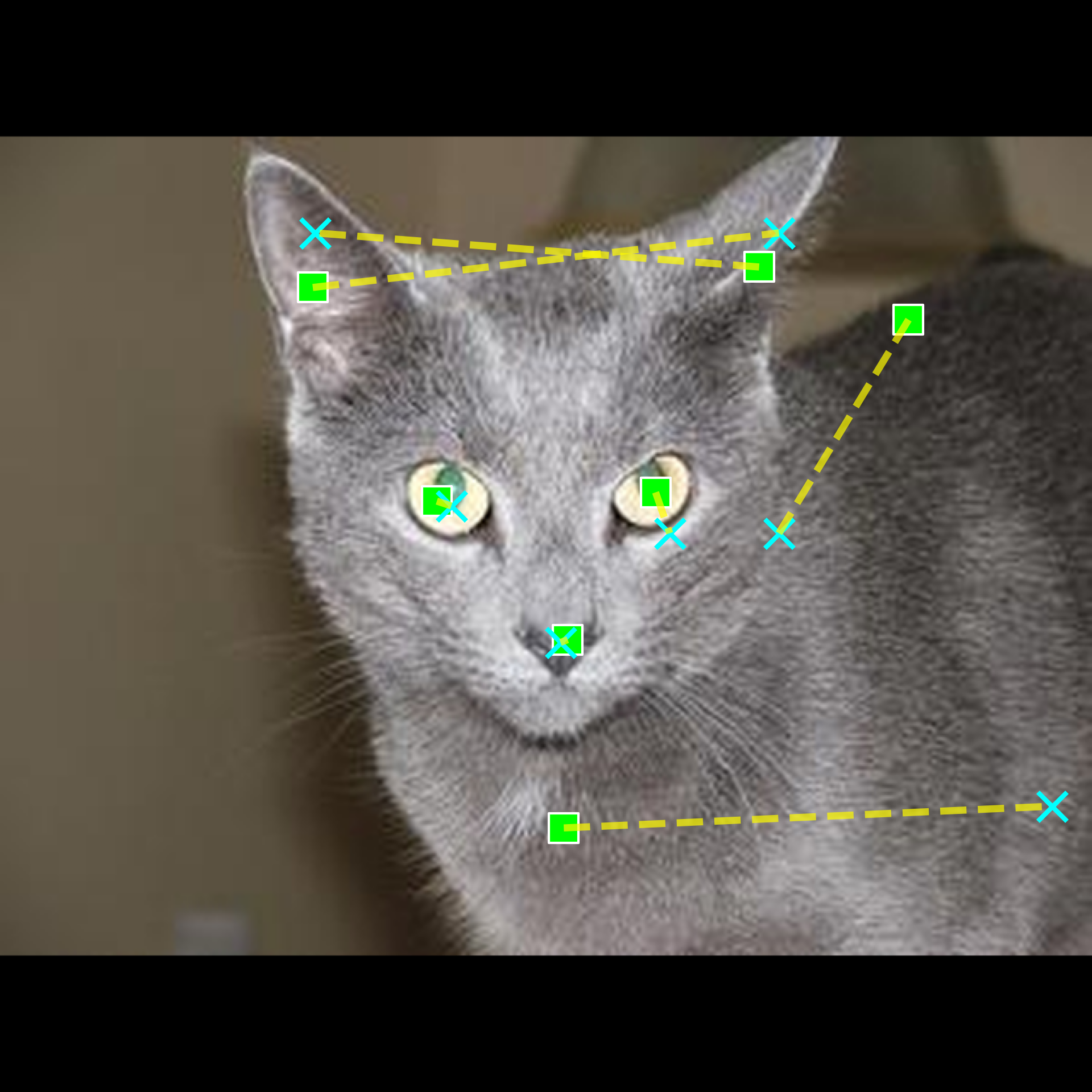}}
    \hspace{0.25pt}%
    \subcaptionbox{Ours}[0.322\linewidth]{%
      \zoomwidecrop{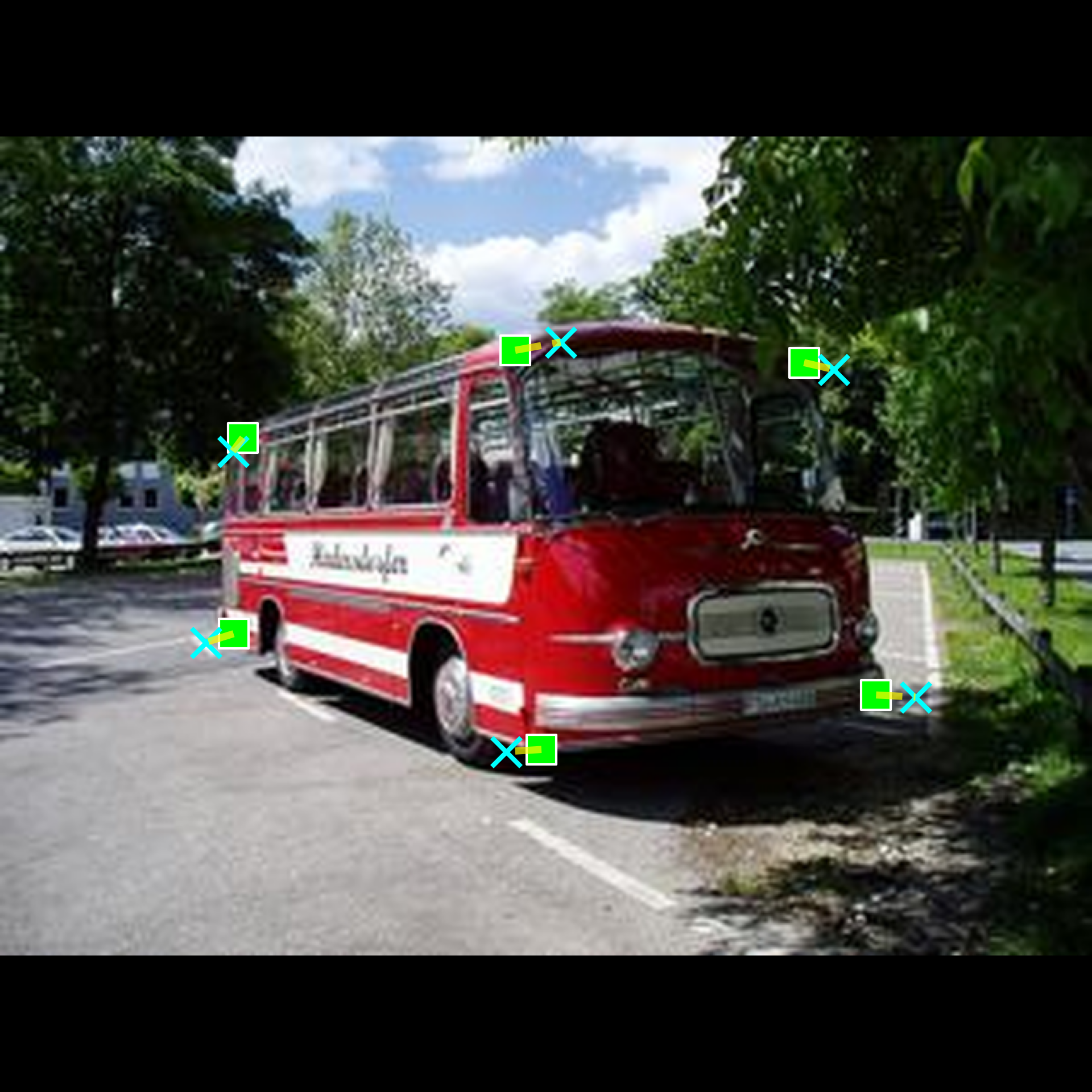}\par
      \vspace{2pt}
      \zoomwidecrop{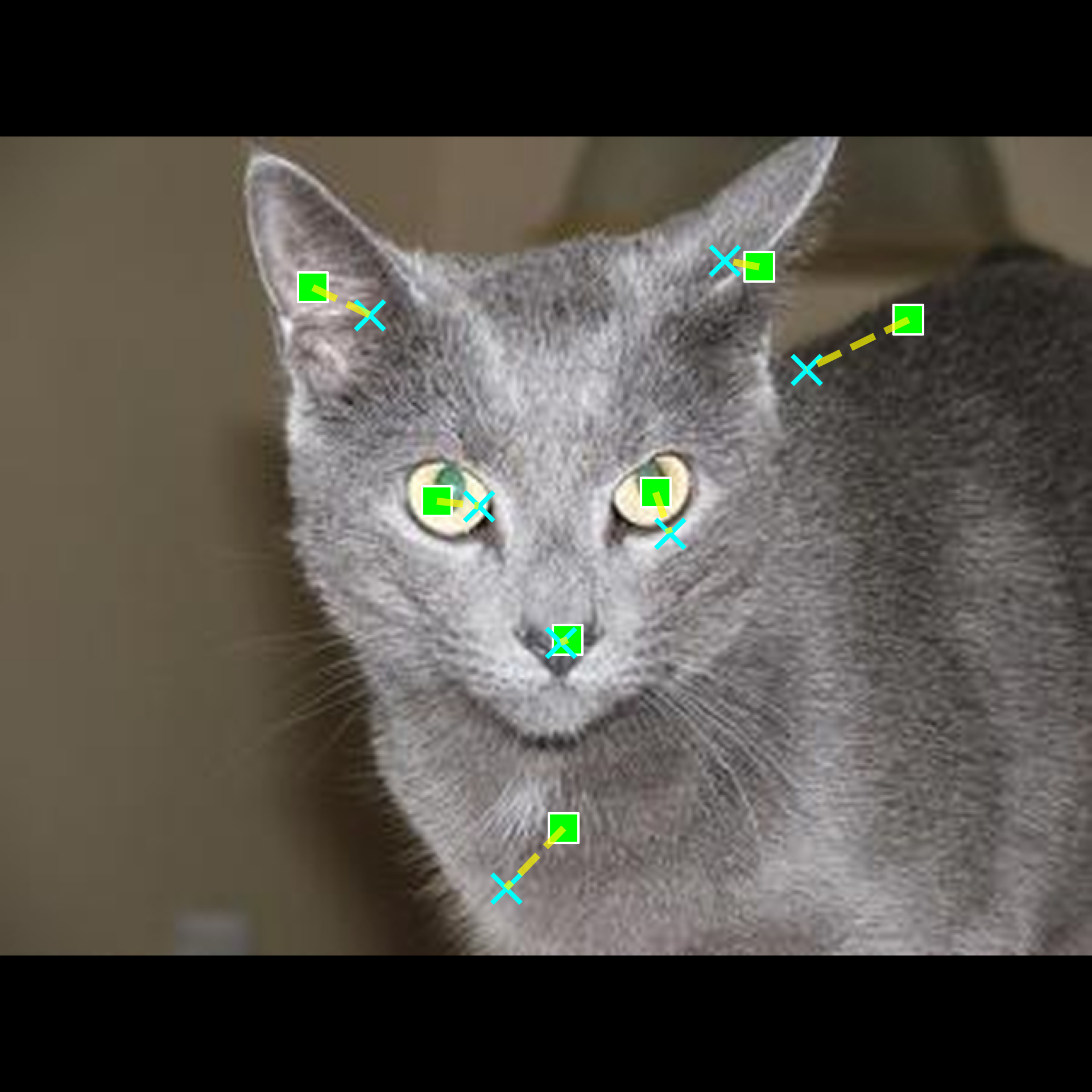}}
    \vspace{-0.6em}
    \caption{
        \textbf{Semantic Transfer.}
        (a) Source image with annotated keypoints. Transfer results using (b) MEF~\cite{you2024multiview_ME} and (c) our approach.
        Our method produces more accurate and spatially consistent transfers.
    }
    \label{fig:quali_semantic_transfer}
    \vspace{-1.5em}
  \end{minipage}
\end{figure}

\begin{table*}[t!]
  \centering
  \begin{small}
  \caption{Quantitative comparison with linear probing on depth estimation and semantic segmentation.}
  \vspace{-1.0em}
  \label{tab:main_linear_probing}
  \scalebox{0.85}{
  \begin{threeparttable}
  \begin{tabular}{@{}lccccccccc@{}}
    \toprule
    \multirow{2}{*}{\textbf{Method}} &
    \multirow{2}{*}{\textbf{Train Time} ($\downarrow$)} &
    \multicolumn{2}{c}{\textbf{ScanNet++}} &
    \multicolumn{2}{c}{\textbf{KITTI}} &
    \multicolumn{2}{c}{\textbf{ADE20K}} &
    \multicolumn{2}{c}{\textbf{VOC2012}} \\ 
    \cmidrule(lr){3-4} \cmidrule(lr){5-6} \cmidrule(lr){7-8} \cmidrule(lr){9-10}
      & & RMSE ($\downarrow$) & Rel.\ ($\downarrow$) &
          RMSE ($\downarrow$) & Rel.\ ($\downarrow$) &
          mIoU ($\uparrow$) & mAcc ($\uparrow$) &
          mIoU ($\uparrow$) & mAcc ($\uparrow$) \\
    \midrule
    (Vanilla) CLIP                & --               & 0.432 & 0.317 & 3.946 & 0.150 & 40.11 & 55.75 & 76.44 & 89.42 \\
    FiT3D~\cite{yue2024improving_FiT3D} & $\sim$3\,d        & \underline{0.394} & \underline{0.278} & \underline{3.542} & \underline{0.125} & \textbf{42.53} & \underline{56.61} & \textbf{79.21} & \underline{90.25} \\
    MEF~\cite{you2024multiview_ME}      & \textbf{$\sim$1\,h} & 0.429 & 0.312 & 3.891 & 0.145 & 40.16 & 55.93 & 76.47 & 89.46 \\
    Ours                               & \underline{$\sim$1\,h 20 m} & \textbf{0.367} & \textbf{0.260} & \textbf{3.529} & \textbf{0.117} & \underline{41.86} & \textbf{57.01} & \underline{78.74} & \textbf{90.41} \\
    \vspace{-4pt} &  & \gainp{-0.065} & \gainp{-0.057} &
                     \gainp{-0.417} & \gainp{-0.033} &
                     \gainp{+1.75} & \gainp{+1.26} & 
                     \gainp{+2.30} & \gainp{+0.99} \\
    \bottomrule
  \end{tabular}
  \end{threeparttable}}
  \end{small}

  \vspace{-0.5em}
\end{table*}

\begin{table*}[t!]
    \caption{
        Comparison of 3D vision-language reasoning on SQA3D and ScanQA.
    }
    \vspace{-1.0em}
    \label{tab:main_3d_vqa}
    \centering
    \begin{small}
    \scalebox{0.825}{
    \begin{threeparttable}
    \begin{tabular}{@{}lcccccccccc@{}}
        \toprule
        \multirow{2}{*}{\textbf{Method}} &
        \multicolumn{5}{c}{\textbf{SQA3D}} &
        \multicolumn{5}{c}{\textbf{ScanQA}} \\
        \cmidrule(l){2-6} \cmidrule(l){7-11}
          & EM-1 & BLEU-1 & METEOR & ROUGE & CIDEr & EM-1 & BLEU-1 & BLEU-4 & METEOR & ROUGE \\
        \midrule
        (Vanilla) CLIP                         & 48.1 & 47.3 & \underline{34.6} & 48.6 & 124.5 & \underline{19.6} & \underline{36.4} & \underline{10.7} & \underline{14.4} & \underline{36.0} \\
        MEF~\cite{you2024multiview_ME}         & \underline{48.2} & \underline{47.4} & \underline{34.6} & \underline{48.7} & \underline{124.7} & 19.0 & 36.1 & 10.4 & 14.3 & 35.1 \\
        Ours & \textbf{48.6} & \textbf{47.7} & \textbf{35.0} & \textbf{49.0} & \textbf{125.5} & \textbf{20.7} & \textbf{36.6} & \textbf{11.6} & \textbf{14.5} & \textbf{36.3} \\
        \vspace{-4pt} & \gainp{+0.5} & \gainp{+0.4} & \gainp{+0.4} & \gainp{+0.4} & \gainp{+1.0} & \gainp{+1.1} & \gainp{+0.2} & \gainp{+0.9} & \gainp{+0.1} & \gainp{+0.3} \\
        \bottomrule
    \end{tabular}
    \end{threeparttable}}
    \end{small}

    \vspace{-0.5em}
\end{table*}

\begin{table*}[t!]
  \centering
    \caption{Ablation study of loss components on 3D correspondence understanding after finetuning on Objaverse.}
    \vspace{-1.0em}
  \label{tab:main_ablation_components}
  \small
  \resizebox{\textwidth}{!}{
  \begin{threeparttable}
  \begin{tabular}{ccc cccccc cc ccc}
    \toprule
    \multicolumn{3}{c}{\textbf{Loss Components}} &
    \multicolumn{6}{c}{\textbf{Semantic Correspondence}} &
    \multicolumn{2}{c}{\textbf{Video Tracking}} &
    \multicolumn{3}{c}{\textbf{Pose Estimation}} \\

    \cmidrule(lr){1-3}\cmidrule(lr){4-9}\cmidrule(lr){10-11}\cmidrule(lr){12-14}
    \multirow{2}{*}{$\mathcal{L}_{\texttt{match}}$} &
    \multirow{2}{*}{$\mathcal{L}_{\texttt{depth}}$} &
    \multirow{2}{*}{$\mathcal{L}_{\texttt{cost}}$} &
      \multicolumn{3}{c}{Different Views} &
      \multicolumn{3}{c}{Same Views} &
      \multirow{2}{*}{Jaccard} & \multirow{2}{*}{Avg.\ Pts} &
      \multicolumn{3}{c}{Accuracy within Thresholds} \\
    \cmidrule(lr){4-6}\cmidrule(lr){7-9}\cmidrule(lr){12-14}
      & & & 0.05 & 0.10 & 0.15 & 0.05 & 0.10 & 0.15 & & & 
      1cm–1deg & 3cm–3deg & 5cm–5deg \\
    \midrule
    \yesmark & \nomark & \nomark & 21.18 & 33.54 & 43.58 & 25.94 & 43.33 & 53.87 & 34.61 & 50.58 &  6.32 & 32.00 & 48.33 \\
    \yesmark & \yesmark & \nomark & \underline{24.89} & \underline{38.32} & \underline{49.00} & \underline{31.92} & \underline{52.05} & \underline{62.88} & \underline{35.36} & \underline{53.43} &  \underline{8.38} & \textbf{42.01} & \textbf{60.26} \\
    \yesmark & \yesmark & \yesmark & \textbf{25.87} & \textbf{39.85} & \textbf{50.21} & \textbf{36.77} & \textbf{56.61} & \textbf{67.93} & \textbf{35.60} & \textbf{54.65} &  \textbf{8.50} & \underline{39.30} & \underline{57.68} \\
    \bottomrule
  \end{tabular}
  \end{threeparttable}}

  \vspace{-1.0em}
\end{table*}

\subsubsection{3D Visual Understanding}
\label{subsec:3d_visual_eval}

\textbf{3D Correspondence Understanding.} 
We evaluate how effectively our distilled 3D-aware VLM representations capture robust multi-view correspondences, following established protocols from~\citet{you2024multiview_ME}.
As summarized in~\Cref{tab:main_semantic_corr,tab:main_video_track_and_pose_est}, the baseline CLIP and FiT3D~\cite{yue2024improving_FiT3D} exhibit limited performance. Specifically, FiT3D slightly degrades the ability of semantics matching, corroborating findings by~\cite{you2024multiview_ME}. MEF~\cite{you2024multiview_ME} significantly improves performance as it leverages explicit 3D annotations. Nevertheless, our approach consistently outperforms MEF even without such annotations. On the Objaverse dataset, our geometric distillation yields notable improvements over the vanilla CLIP.
Moreover, training on the real-world ScanNet++ dataset results in further substantial gains of \textbf{+11.87\%} in PCK@0.05, \textbf{+12.36\%} in average Jaccard index, and \textbf{+8.46\%} accuracy at the 1cm-1deg threshold. This demonstrates the practical value and strong generalization power of our method. Unlike MEF, which indiscriminately uses 3D annotations, our distillation naturally selects semantically meaningful key regions, leading to more effective correspondence learning. These observations confirm that our approach effectively transfers strong geometric priors into VLM representations by improving cross-view consistency without explicit ground-truth 3D supervision. Further qualitative comparisons provided in~\Cref{fig:quali_semantic_transfer} support these quantitative results.

\noindent \textbf{Depth Estimation and Semantic Segmentation.} We demonstrate the transferability of our distilled VLM features via linear probing on monocular depth estimation and semantic segmentation tasks after fine-tuning on ScanNet++. Although traditionally 2D-oriented, performance on these tasks heavily relies on robust 3D geometric understanding~\cite{yue2024improving_FiT3D}. We measure depth prediction accuracy with RMSE and absolute relative error (Rel.), and semantic segmentation using mIoU and mAcc. As shown in~\Cref{tab:main_linear_probing}, FiT3D significantly improves both tasks but requires approximately three days of training on four NVIDIA A6000 GPUs due to costly 3D Gaussian optimization across training scenes. MEF shows marginal improvements over baseline CLIP, indicating limited effectiveness for dense predictions.
Our approach achieves the best depth estimation performance, reducing RMSE from \textbf{0.432} to \textbf{0.367} on ScanNet++, and obtains competitive semantic segmentation results while requiring up to 54 times less computation than FiT3D on a single GPU. Without explicit dense 3D optimization, our method effectively injects robust depth priors into VLMs, enhancing semantic scene understanding.

\subsubsection{3D Vision-Language Understanding}
\label{subsec:exp_3dvqa}

To evaluate whether our distilled VLM features effectively enhance 3D vision-language understanding, we conduct experiments on two representative 3D VQA benchmarks with fine-tuned CLIP features, following the evaluation protocol from Lexicon3D~\cite{man2024lexicon3d}. 
We measure performance using EM-1, BLEU, METEOR, ROUGE, and CIDEr. 
Among these metrics, EM-1 is particularly crucial as it directly measures the model's exact answer prediction accuracy. For fair comparisons, we fine-tune all baselines on the Objaverse dataset. As shown in~\Cref{tab:main_3d_vqa}, MEF does not show significant improvements over the vanilla CLIP on SQA3D and even lower performance on ScanQA. In contrast, our method consistently outperforms both CLIP and MEF across all metrics and datasets. Specifically, our approach increases EM-1 on SQA3D \textbf{to 48.6\%}, and notably improves EM-1 on ScanQA \textbf{from 19.6\% to 20.7\%}. These results demonstrate that our fine-tuning approach provides better 3D visual understanding which effectively leads to improvement of 3D spatial knowledge for vision-language reasoning.

\subsection{Ablation Study}
\label{subsec:ablation_study}


We conduct an ablation study to analyze the effectiveness of each loss component for 3D correspondence understanding as in~\Cref{subsec:3d_visual_eval} after fine-tuning on Objaverse. 
Compared to fine-tuning solely with $\mathcal{L}_{\texttt{match}}$ equivalent to MEF, adding $\mathcal{L}_{\texttt{depth}}$ consistently improves performance across all metrics.
Incorporating $\mathcal{L}_{\texttt{cost}}$ further boosts PCK@0.05 by \textbf{+4.69\%} and video tracking position accuracy by \textbf{+4.07\%}. Although pose estimation accuracy slightly decreases at some thresholds, it maintains improved performance with a gain of \textbf{+2.18\%} at the challenging 1cm-1deg threshold. These results demonstrate that $\mathcal{L}_{\texttt{depth}}$ significantly enhances semantic matching and precise localization, while cost $\mathcal{L}_{\texttt{cost}}$ further strengthens cross-view feature consistency. Additional ablation analyses are provided in the appendix.

\section{Conclusion}
\label{sec:conclusion}

We present Geometric Distillation, a lightweight and annotation-free framework that enhances 3D spatial awareness and reasoning in VLMs.
By distilling rich geometric signals such as multi-view correspondences, relative depth relations, and dense cost volumes from high-capacity 3D foundation models like MASt3R and VGGT, our method equips pretrained 2D VLMs with robust 3D perception.
Without requiring architectural modifications or explicit 3D annotations, our approach improves state-of-the-art results across diverse spatial reasoning tasks, including semantic correspondence, depth estimation, and 3D visual question answering.
Extensive experiments demonstrate that our method consistently outperforms prior approaches while offering greater scalability and generalization to real-world scenes.
Our work highlights an effective pathway to bridge the gap between 2D vision-language understanding and 3D perception.


\section{Limitations \& Future Work}
\label{sec:Limitations}


While our approach achieves notable improvements in 3D spatial reasoning for vision-language models without requiring explicit annotations or architectural changes, several limitations remain.
First, the method assumes access to multi-view imagery during training, which may not always be feasible in practical applications.
Second, the reliance on 3D foundation models as supervision sources introduces potential biases and limits the controllability over the distilled geometric signals.
Additionally, our framework does not directly generalize to other 3D modalities such as point clouds or meshes.

Future work will focus on extending geometric distillation to monocular settings and exploring self-supervised alternatives to reduce dependence on external teacher models.


\section*{Acknowledgements}
\label{sec:Acknowledgements}


\noindent
This research was supported by the Basic Science Research Program through the National Research Foundation of Korea (NRF), funded by the MSIP (RS-2025-00520207, RS-2023-00219019), KEIT grant funded by the Korean government (MOTIE) (No. 2022-0-00680, No. 2022-0-01045), Artificial Intelligence Graduate School Program (KAIST) (RS-2019-II190075), and SAMSUNG Research, Samsung Electronics Co., Ltd.


\bibliography{references}

\appendix


\appendix
\clearpage

\section*{Appendix Contents}


{
\begin{itemize}
    \item \textbf{A. Potential Risks}
    \item \textbf{B. Use or Create Scientific Artifacts}
    \begin{itemize}
        \item B.1 Discuss The License For Artifacts
        \item B.2 Documentation of Artifacts
        \item B.3 Statistics for Dataset
    \end{itemize}
    \item \textbf{C. Computational Experiments}
    \begin{itemize}
        \item C.1 Model Size and Budget
        \item C.2 Experimental Setup and Hyperparameters
        \item C.3 Descriptive Statistics
        \item C.4 Parameters for Packages
    \end{itemize}
    \item \textbf{D. Use of AI Assistants}
    \begin{itemize}
        \item Information About Use Of AI Assistants
    \end{itemize}
    \item \textbf{E. Additional Quantitative Evaluation}
    \begin{itemize}
        \item Feature Visualization
        \item More Qualitative Results
        \item Example Result of 3D VQA
    \end{itemize}
    \item \textbf{F. Additional Ablation Study}
    \begin{itemize}
        \item Comparison of Absolute and Relative Depth Understanding
        \item Ablation on Loss Components with Different Training Dataset
        \item Comparison of MASt3R and VGGT as a Teacher Model
    \end{itemize}
    \item \textbf{G. Failure Cases}
\end{itemize}
}

\section{Potential Risks}

Our proposed method, Geometric Distillation, enhances vision-language models (VLMs) with 3D spatial understanding by leveraging supervision signals from pretrained 3D foundation models.
While our approach is annotation-free and lightweight, there are potential risks associated with its deployment.
First, since the 3D models used as teachers may contain biases learned from their own training data, such biases could be inadvertently transferred to the VLMs.
Second, because our method relies on pseudo-supervision (e.g., depth maps and correspondences), inaccuracies in the geometric signals could result in incorrect spatial reasoning or degraded model performance.
Finally, although our work is intended for academic and constructive use, enhanced spatial reasoning capabilities could potentially be misused in surveillance, military applications, or other ethically sensitive scenarios.

\section{Use or Create Scientific Artifacts}

Our study builds entirely on existing resources, including publicly available pretrained models and benchmark datasets.
In the following, we briefly describe the licensing status of the artifacts used and provide key statistics for the datasets involved in our experiments.

\subsection{Discuss The License for Artifacts}

In this work, we do not introduce new datasets, but instead make use of publicly available pretrained models and benchmarks.
Specifically, we use MASt3R~\cite{leroy2024mast3r} and VGGT~\cite{wang2025_vggt} as geometric teacher models, which are distributed under research-friendly licenses: VGGT is released under the CC BY-NC 4.0 license, and MASt3R, DUSt3R is licensed under the CC BY-NC-SA 4.0 license. 
Additionally, we evaluate our method using several publicly available datasets: TAP-Vid-DAVIS~\cite{doersch2022tap} (Apache 2.0), OnePose-LowTexture~\cite{he2022onepose++} (Apache 2.0), ADE20K~\cite{zhou2019semantic_ade20k} (CC BSD 3), and Objaverse~\cite{deitke2023objaverse} (Apache 2.0).
All datasets are used strictly for non-commercial research purposes in accordance with their respective licenses or terms.


\subsection{Documentation of Artifacts}

All code, pretrained model checkpoints, and evaluation scripts used in this study will be publicly released upon publication.
These artifacts will be hosted on a GitHub repository, accompanied by detailed documentation including installation instructions, dataset preparation scripts, and usage examples.
A complete README file will be provided to ensure the reproducibility of our results.
For datasets that cannot be redistributed due to licensing constraints, we include scripts and links to download them from their original sources.
Our release is intended to support both reproduction and future research based on our approach.


\subsection{Statistics for Dataset}

\begin{table*}[t]
  \caption{Dataset statistics and split details for each downstream task.}
  \label{tab:dataset_stats}
  \centering
  \small
  \begin{tabularx}{\linewidth}{@{}l X@{}}
    \toprule
    \textbf{Task / Dataset} & \textbf{Split information} \\
    \midrule
    \multicolumn{2}{l}{\textbf{3D Correspondence Understanding}} \\[-0.25em]
    \addlinespace[0.4em]
    \cline{1-2}
    \addlinespace[0.4em]
        PF-PASCAL            & 20 object classes; 308 image pairs; pairs randomly shuffled (in different viewpoint settings) \\ 
        TAP-Vid (DAVIS)      & 30 object-centric videos; 34–104 frames per video \\ 
        OnePose-LowTexture   & 40 objects with two videos per object; evaluation every 10th frame \\ 
    \midrule
    \multicolumn{2}{l}{\textbf{Dense Scene Understanding}} \\[-0.25em]
    \addlinespace[0.4em]
    \cline{1-2}
    \addlinespace[0.4em]
        ScanNet++            & Validation split — 50 scenes, 30,638 images \\ 
        KITTI                & Test split — 28 scenes, 697 images \\ 
        ADE20K               & Validation split — 2,000 images \\ 
        VOC2012              & Validation split — 1,449 images \\ 
    \midrule
    \multicolumn{2}{l}{\textbf{3D Vision–Language Understanding}} \\[-0.25em]
    \addlinespace[0.4em]
    \cline{1-2}
    \addlinespace[0.4em]
        SQA3D                & over 33K question–answer pairs \\ 
        ScanQA               & over 41K question–answer pairs \\ 
    \bottomrule
  \end{tabularx}
\end{table*}

We summarize the dataset statistics used in our experiments across different tasks in~\Cref{tab:dataset_stats}.

\noindent \textbf{3D Correspondence Understanding.} We evaluate on three benchmarks following the protocols from MEF~\cite{you2024multiview_ME}.
For semantic correspondence, we use PF-PASCAL that consists of 308 image pairs from 20 object classes, randomly shuffled in different viewpoint settings.
For video tracking, we follow the protocols of~\cite{doersch2022tap,tumanyan2024dino} and use TAP-Vid-DAVIS, which contains 30 object-centric videos with 34–104 frames per video.
For object pose estimation, we follow~\citet{he2022onepose++} and evaluate on the OnePose-LowTexture dataset which comprises 40 objects, each with two videos, performing evaluations on every 10th frame.

\noindent \textbf{Dense Scene Understanding.} Following FiT3D~\cite{yue2024improving_FiT3D}, we perform linear probing evaluations to estimate monocular depth and semantic segmentation.
For depth estimation, we use ScanNet++~\cite{yeshwanth2023scannet++}, specifically utilizing its validation split of 50 scenes with 30,638 images. We also use KITTI~\cite{geiger2013vision_kitti} to evaluate generalization performance on KITTI's test split consisting of 28 scenes and 697 images.
For semantic segmentation, we follow standard protocols and evaluate on ADE20K~\cite{zhou2019semantic_ade20k}'s validation split with 2,000 images and VOC2012~\cite{everingham2015pascal_voc2012}'s validation split with 1,449 images.

\noindent \textbf{3D Vision-Language Understanding.}
We evaluate 3D visual question-answering capabilities on SQA3D~\cite{ma2022sqa3d} and ScanQA~\cite{azuma2022scanqa}, following Lexicon3D~\cite{man2024lexicon3d}. Both datasets contain diverse QA pairs designed to probe 3D spatial and semantic reasoning. Specifically, SQA3D comprises over 33K synthetic question–answer pairs, while ScanQA contains over 41K real-world question–answer pairs generated from ScanNet scenes.





\section{Computational Experiments}
\label{sec:a_Computational_Experiments}

We conduct a series of computational experiments to evaluate the effectiveness and efficiency of our proposed method. This section outlines the scale and computational cost of our models, the training setup and hyperparameter choices, a summary of the reported evaluation metrics, and the software packages used for implementation and evaluation. Through careful design and efficient training strategies, we ensure that our method achieves strong performance while maintaining high computational efficiency.






\newcommand{\zoomcropslight}[1]{%
  \includegraphics[width=0.495\linewidth,
                   trim=0pt 60pt 0pt 24pt,
                   clip]{#1}}

\newcommand{\zoomcrop}[1]{%
  \includegraphics[width=0.495\linewidth,
                   trim=120pt 240pt 120pt 240pt,
                   clip]{#1}}

\begin{figure*}[t!]
  \centering
  \begin{subfigure}[t]{0.19\textwidth}
    \centering
    

    \zoomcrop{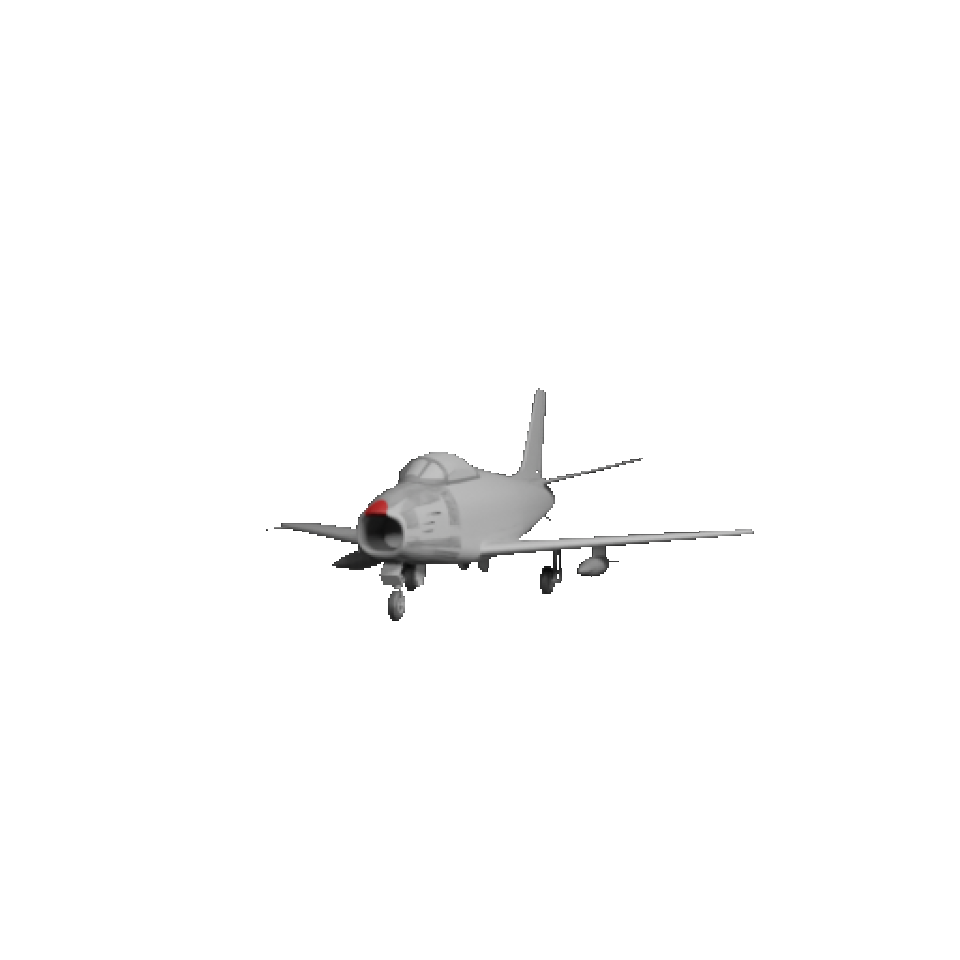}
    
    \zoomcrop{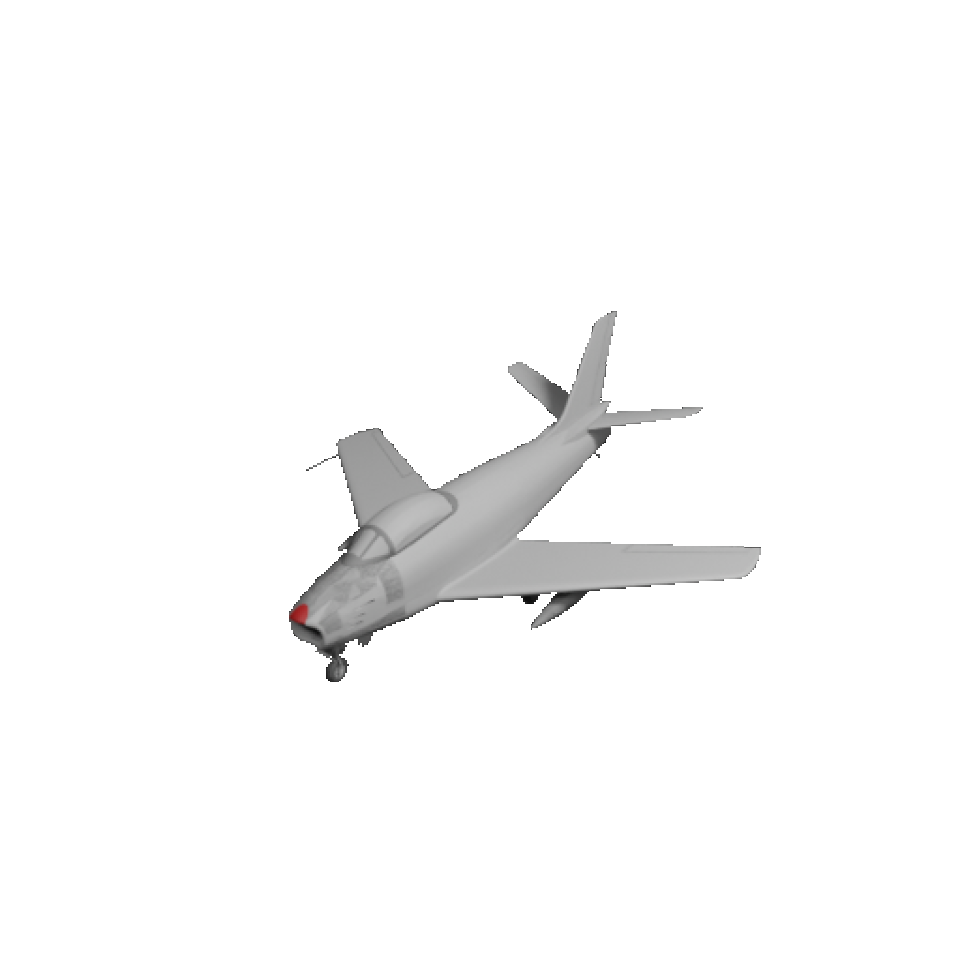}\hspace{0.5pt}%
    \zoomcrop{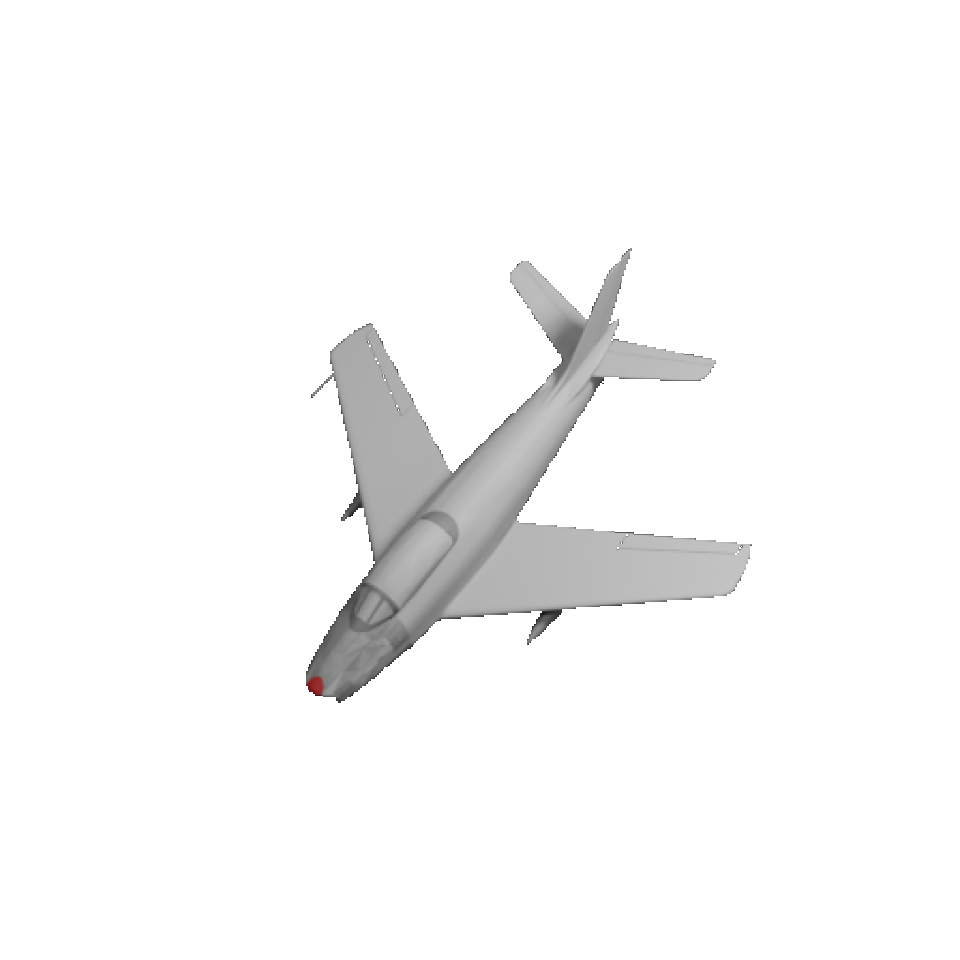}

    \vspace{4pt}

    \zoomcrop{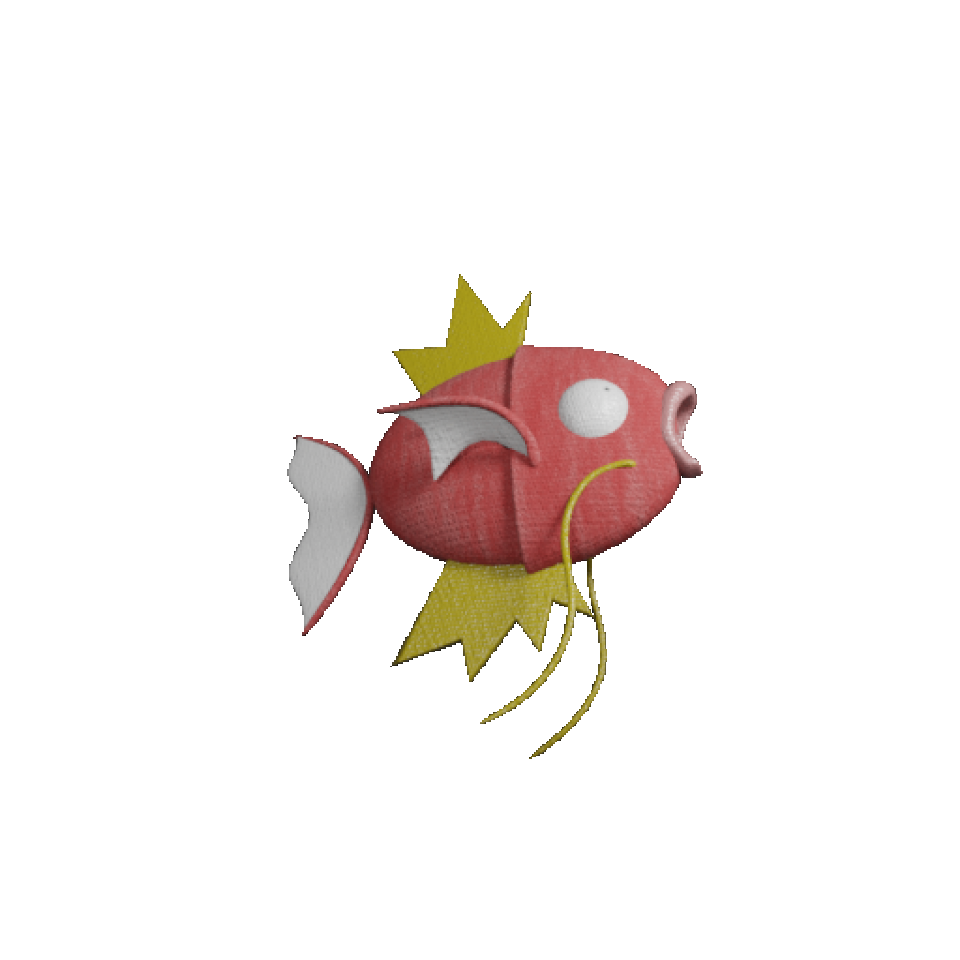}
    
    \zoomcrop{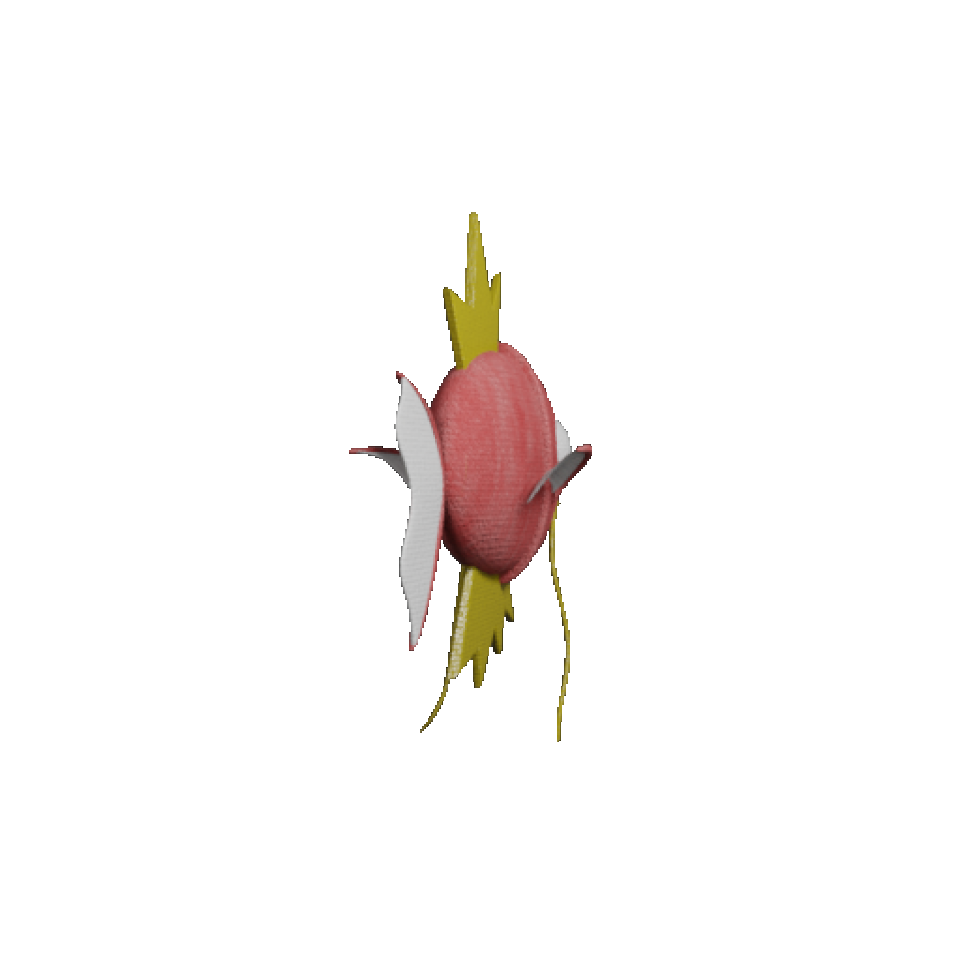}\hspace{0.5pt}%
    \zoomcrop{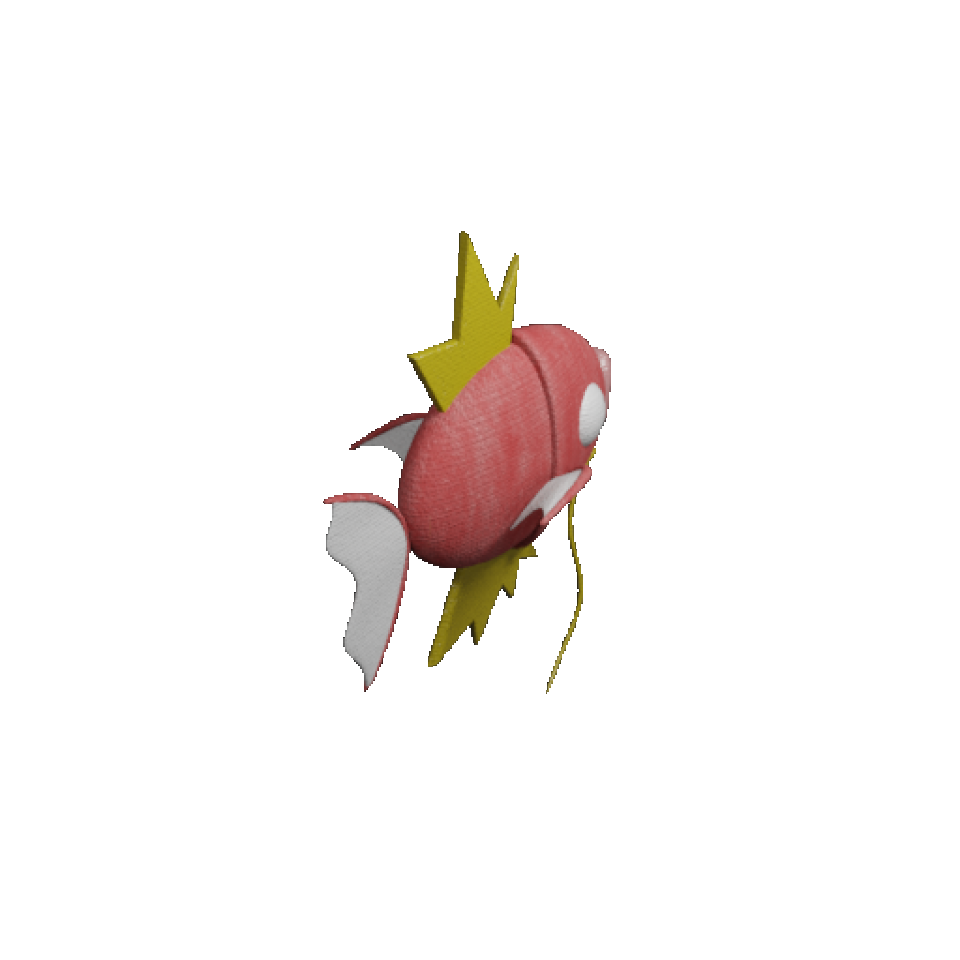}

    \vspace{4pt}
    
    \zoomcrop{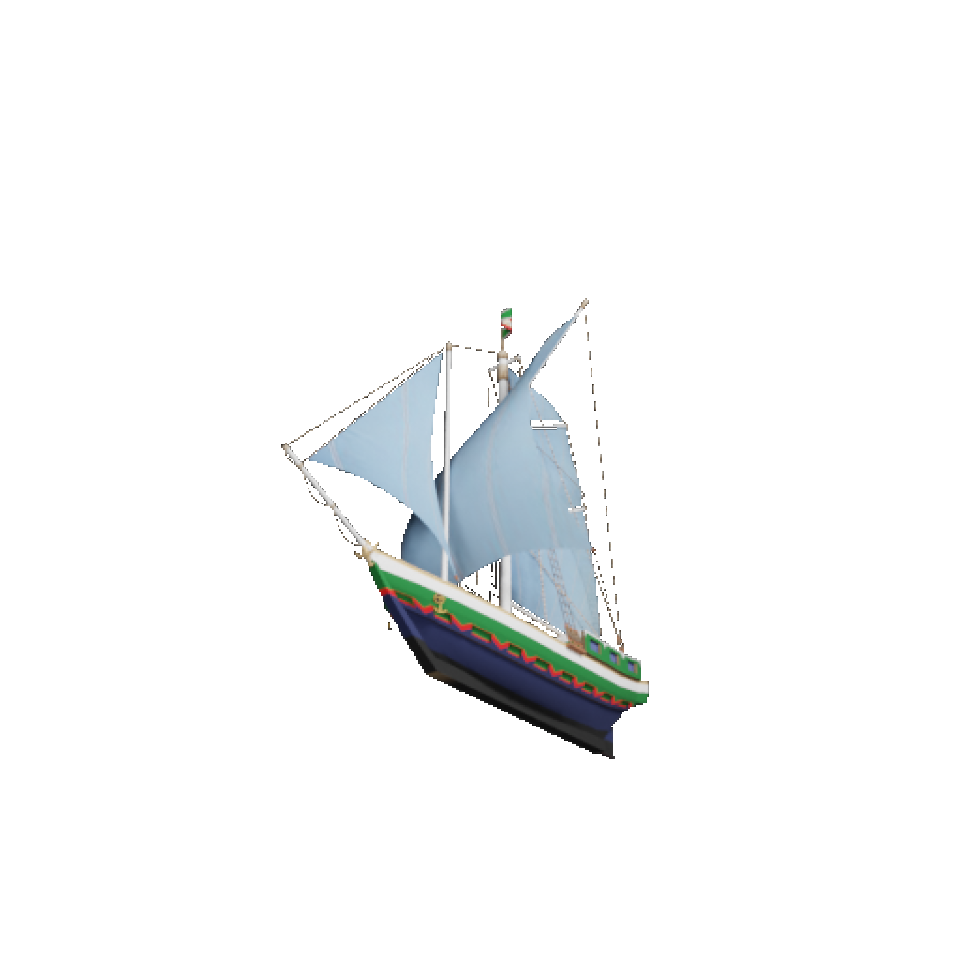}
    
    \zoomcrop{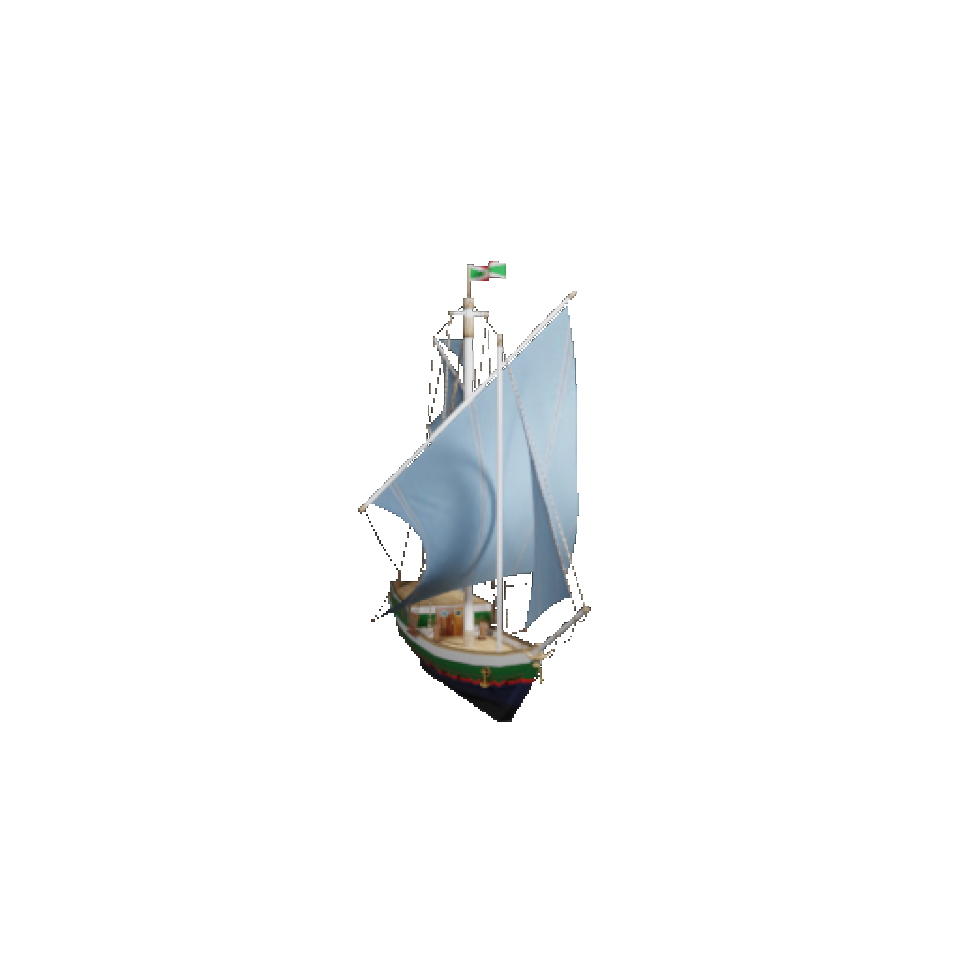}\hspace{0.5pt}%
    \zoomcrop{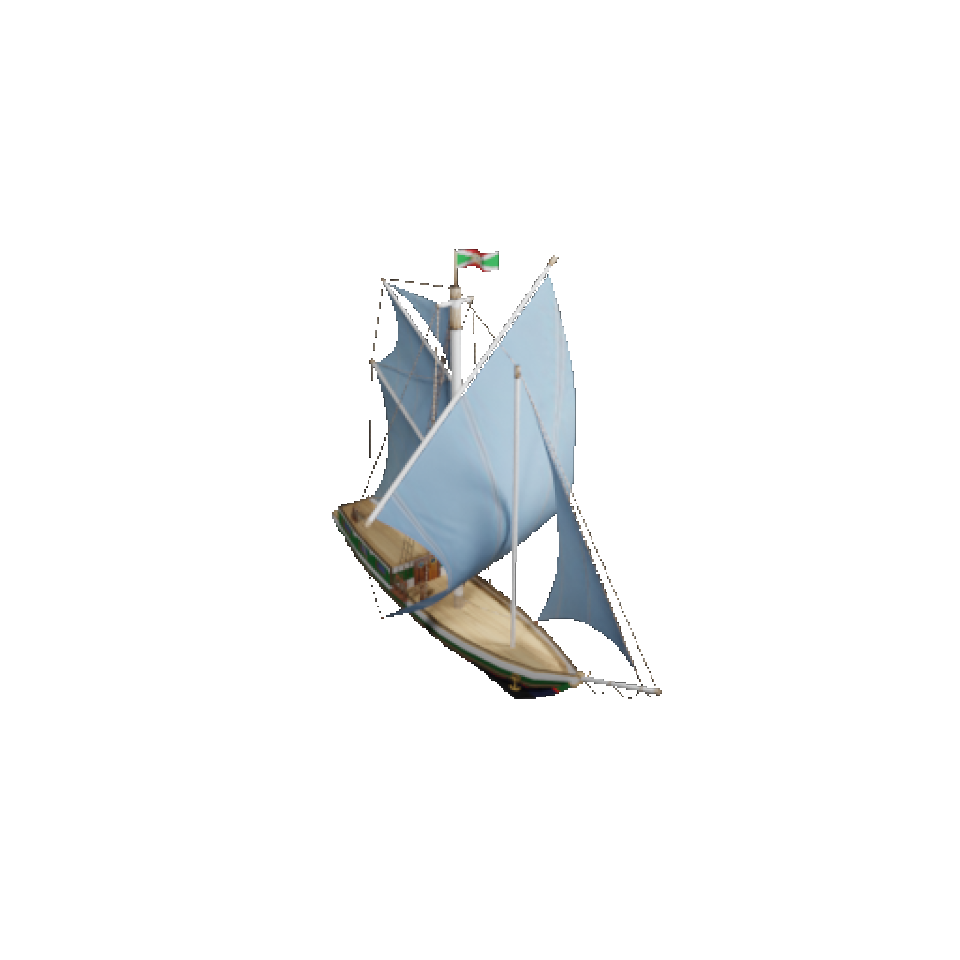}

    \vspace{4pt}
    
    \zoomcrop{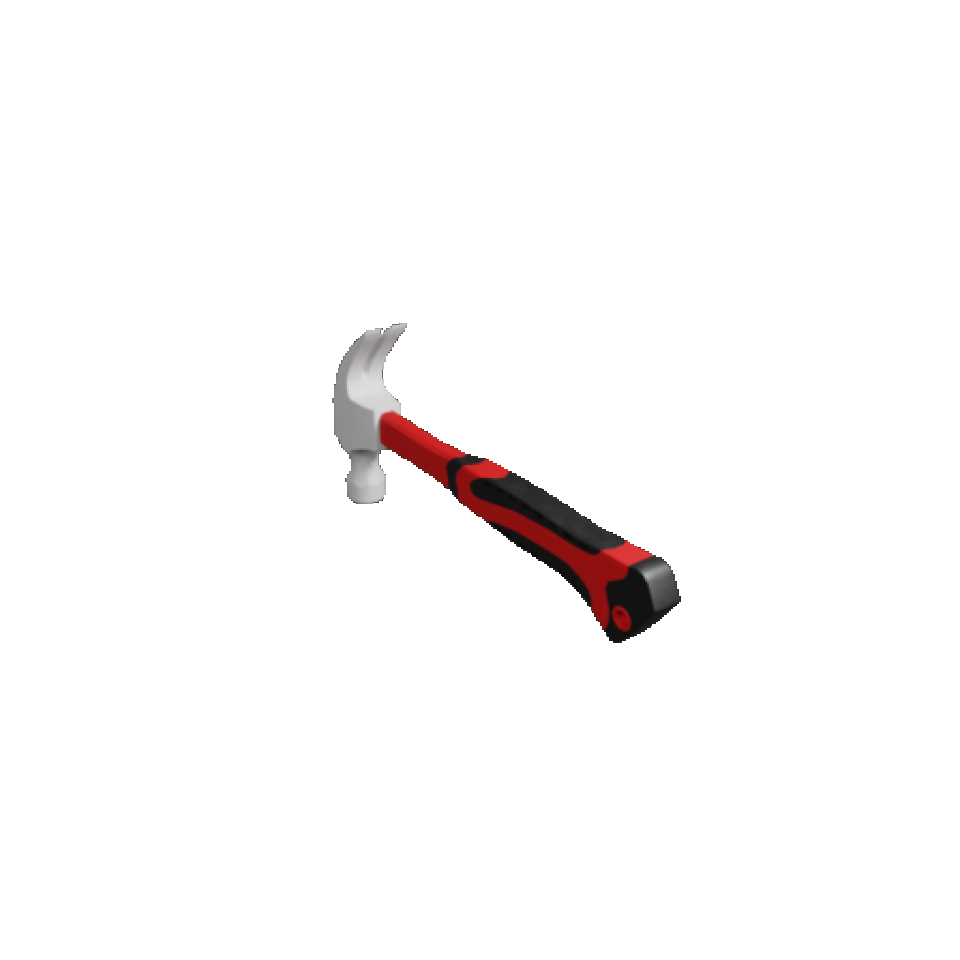}
    
    \zoomcrop{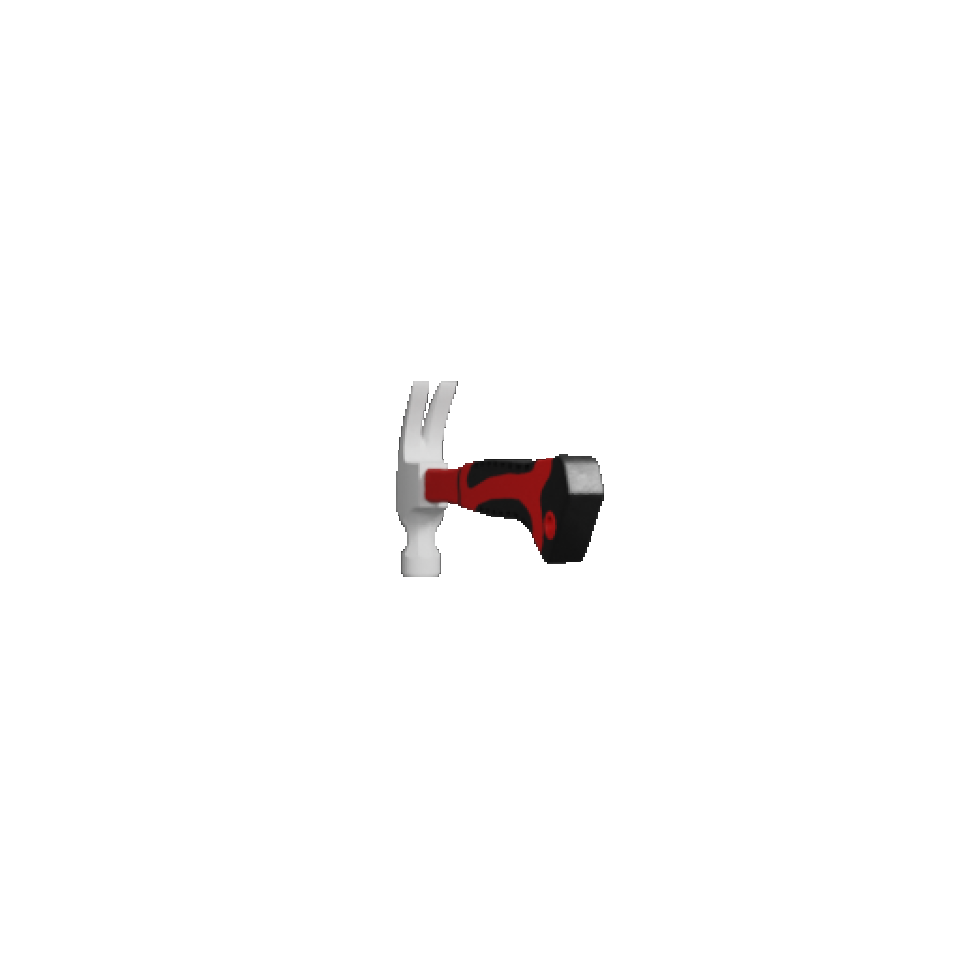}\hspace{0.5pt}%
    \zoomcrop{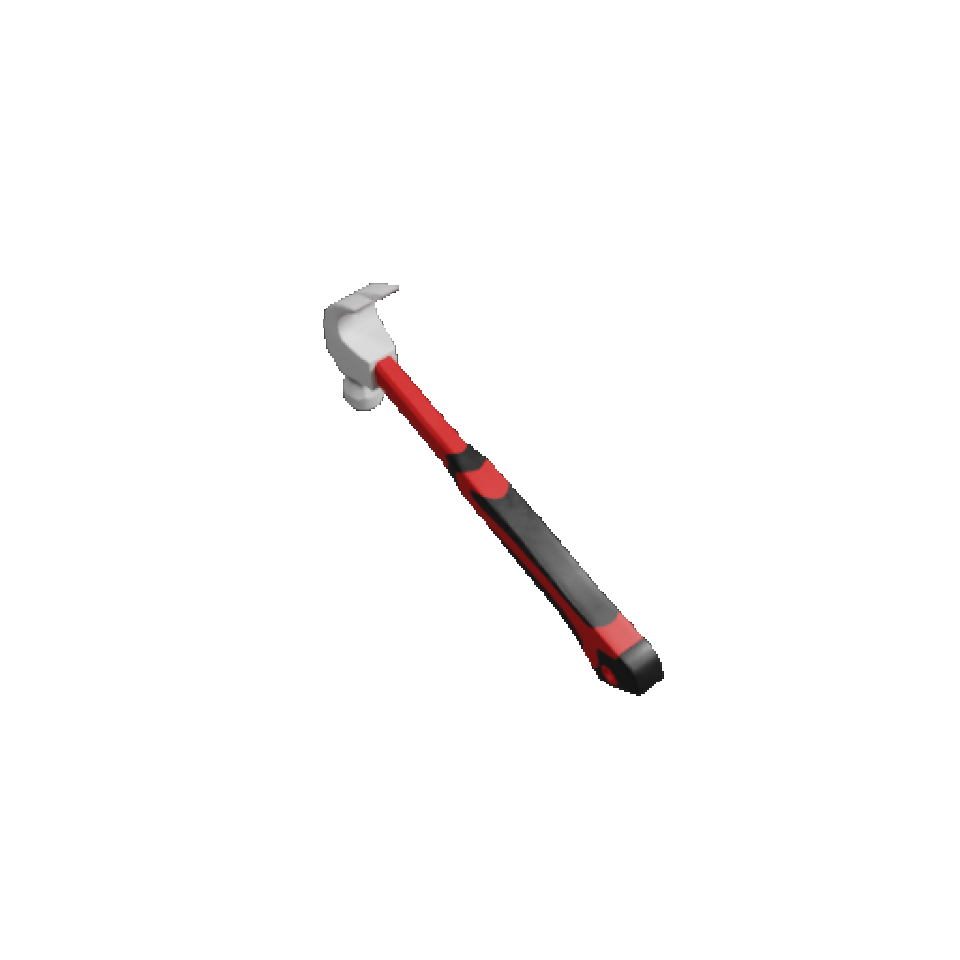}
    \captionsetup{skip=4pt}
    \caption{Images}
  \end{subfigure}
  \begin{subfigure}[t]{0.19\textwidth}
    \centering


    \zoomcrop{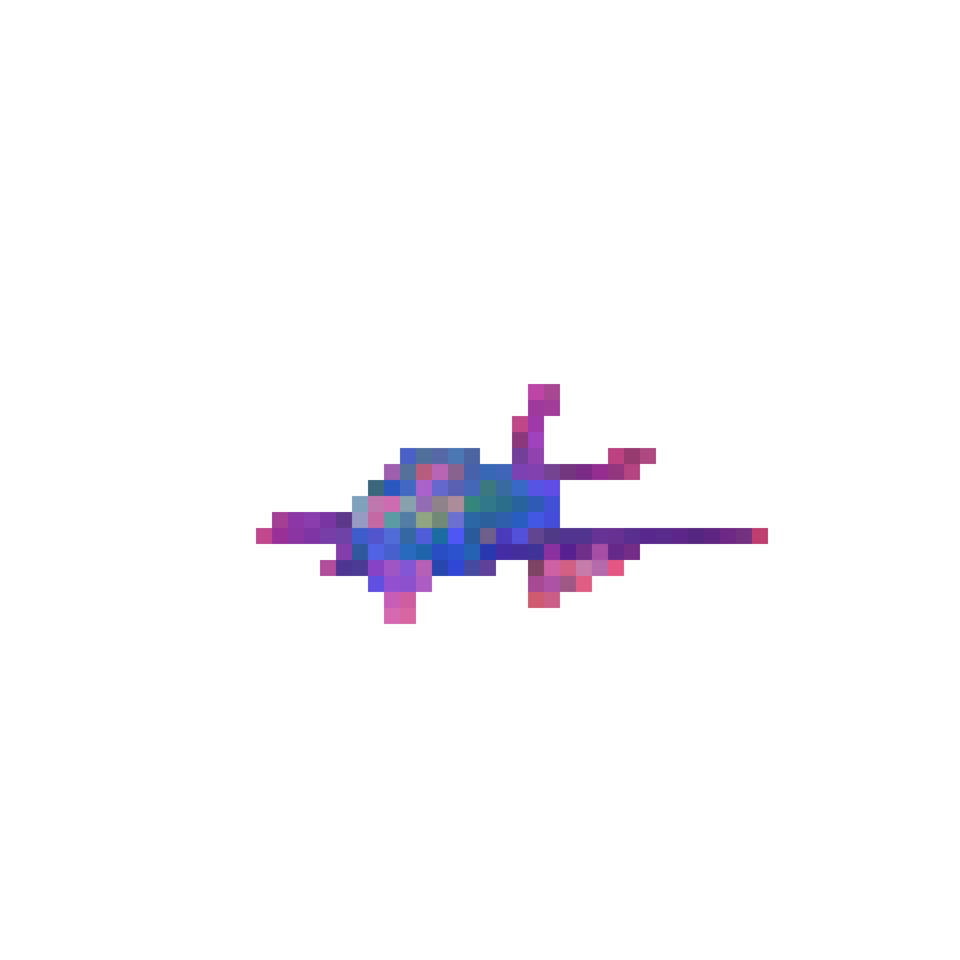}
    
    \zoomcrop{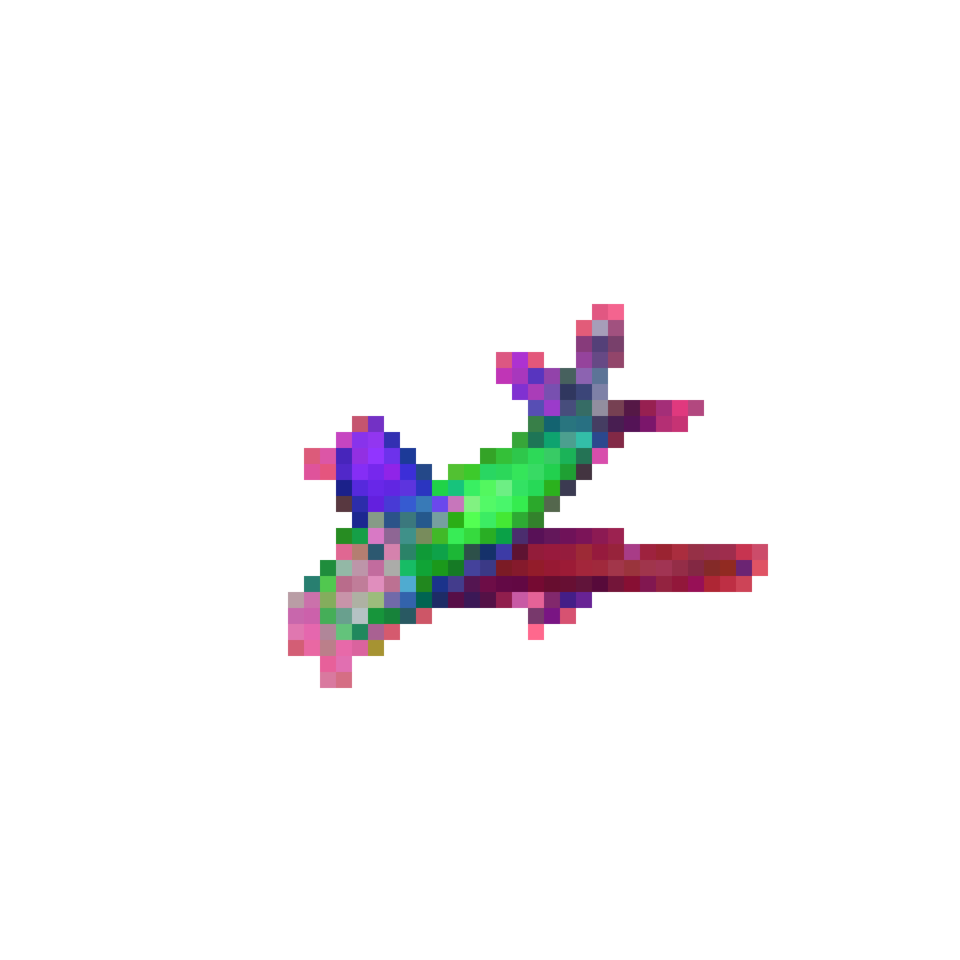}\hspace{0.5pt}%
    \zoomcrop{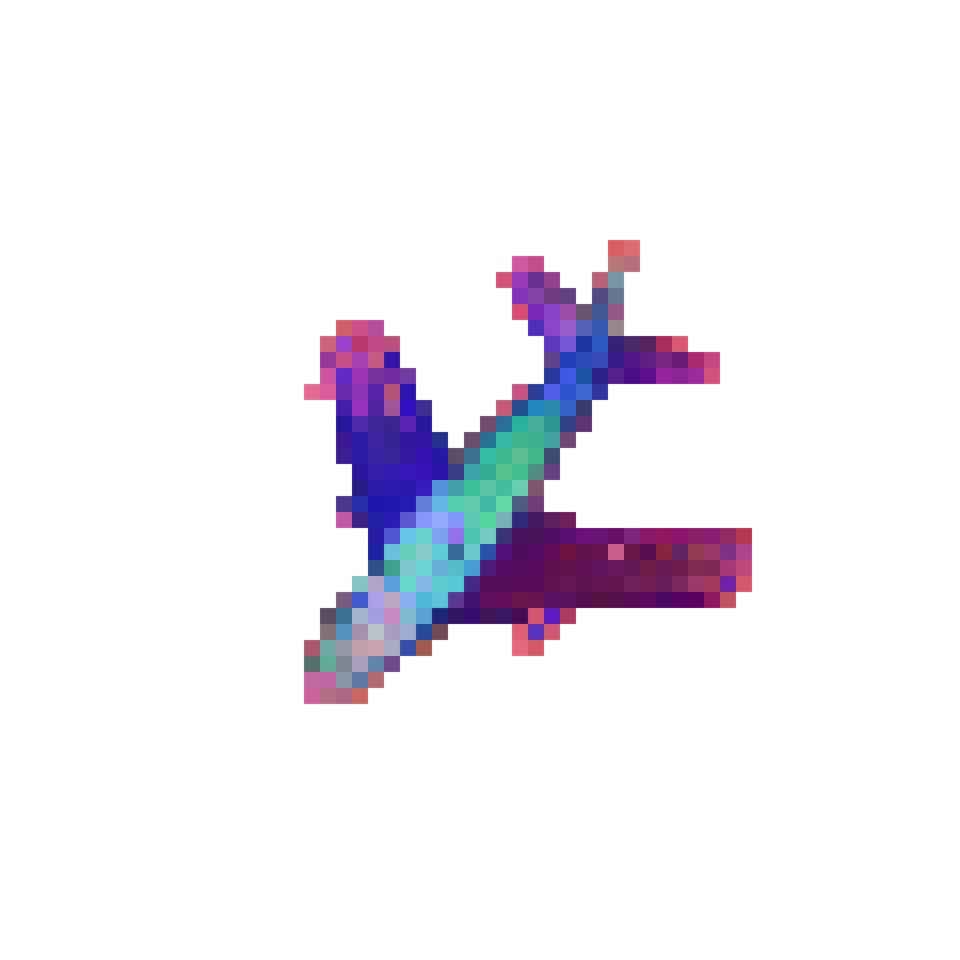}

    \vspace{4pt}

    \zoomcrop{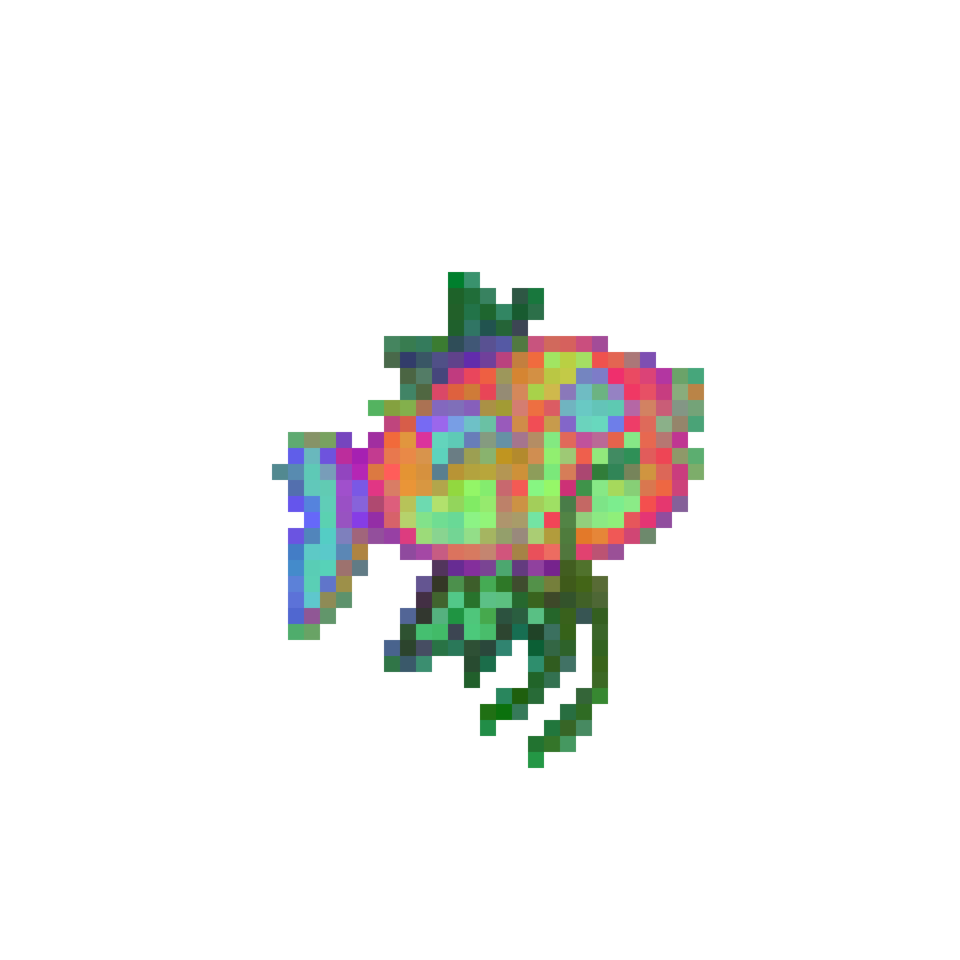}
    
    \zoomcrop{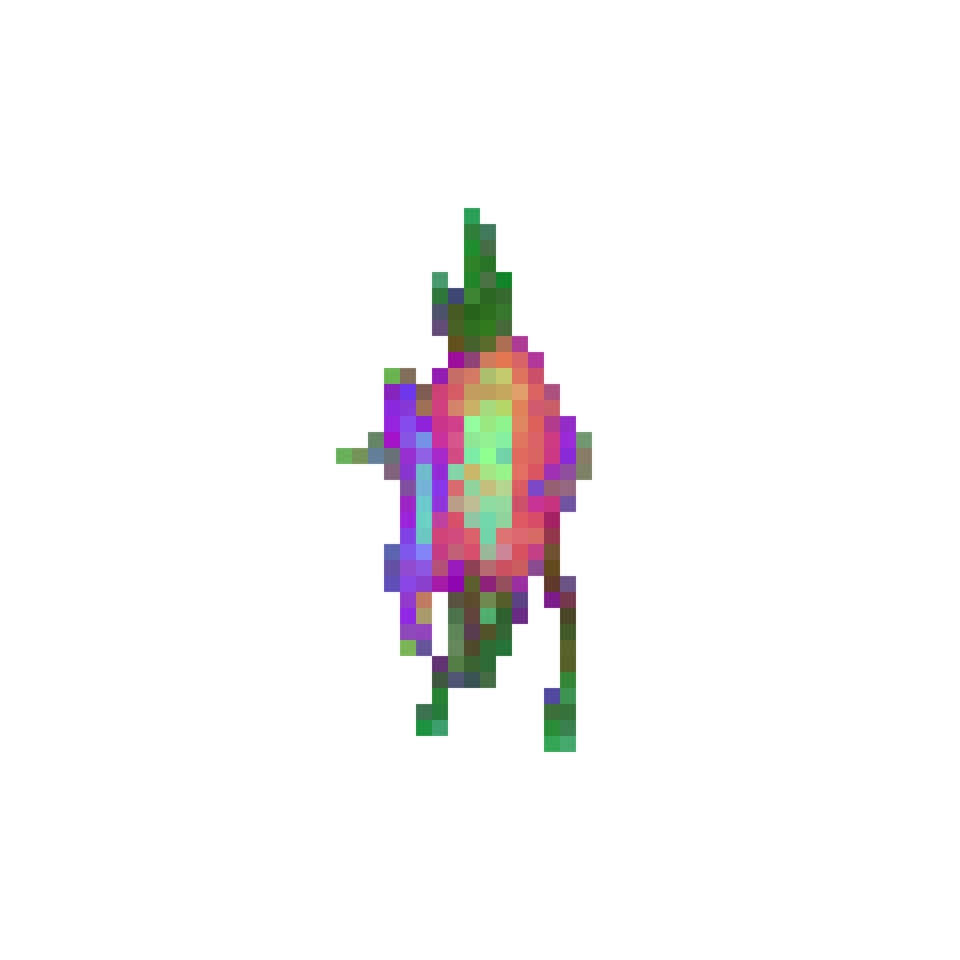}\hspace{0.5pt}%
    \zoomcrop{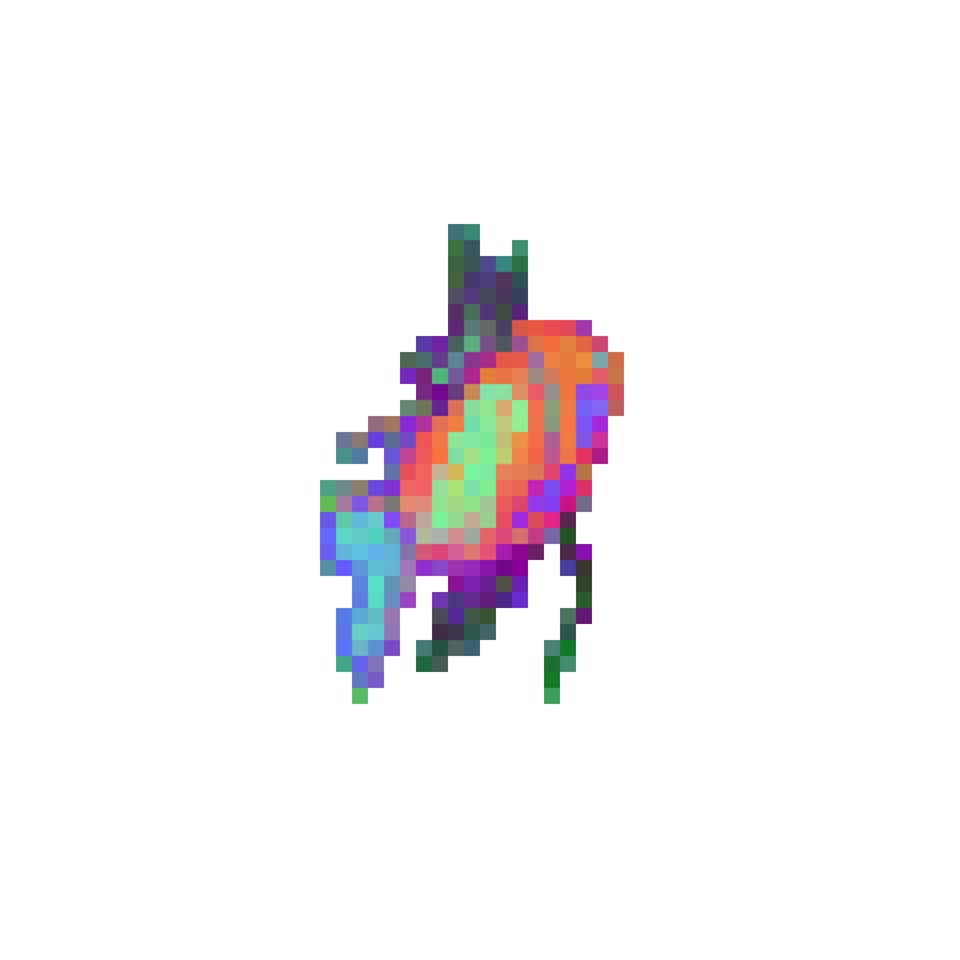}

    \vspace{4pt}

    \zoomcrop{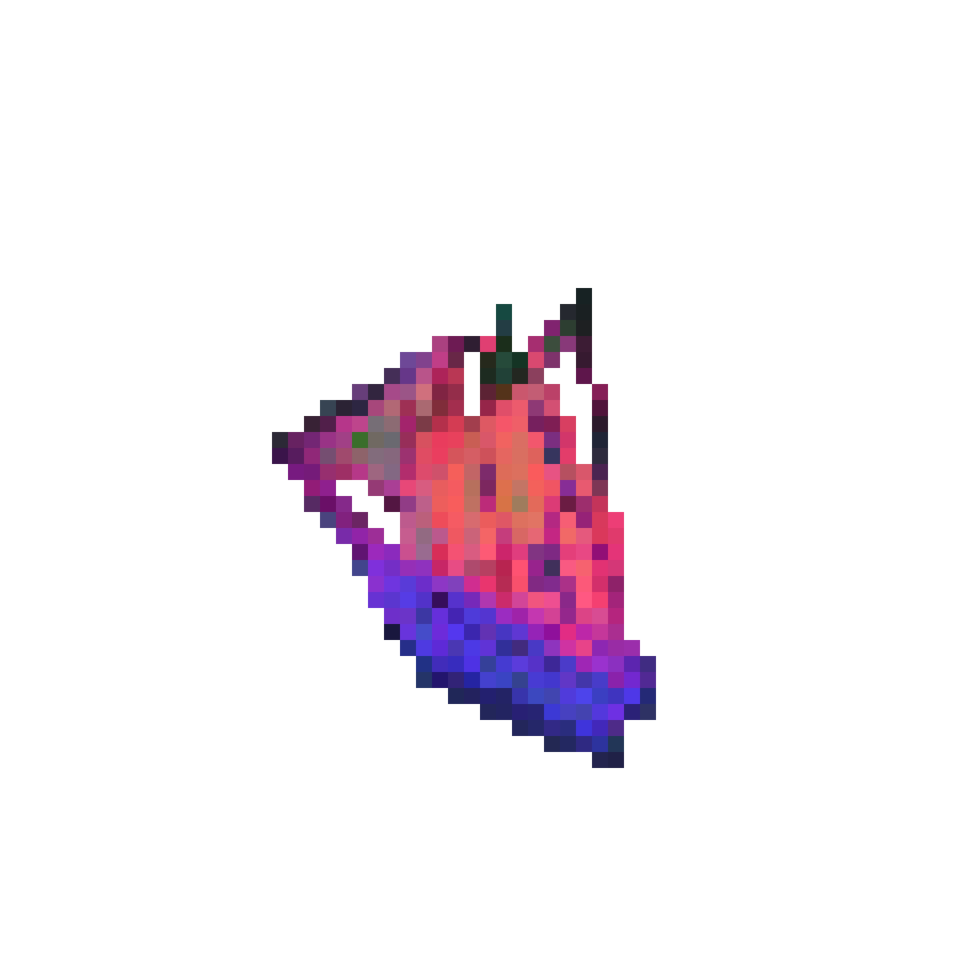}
    
    \zoomcrop{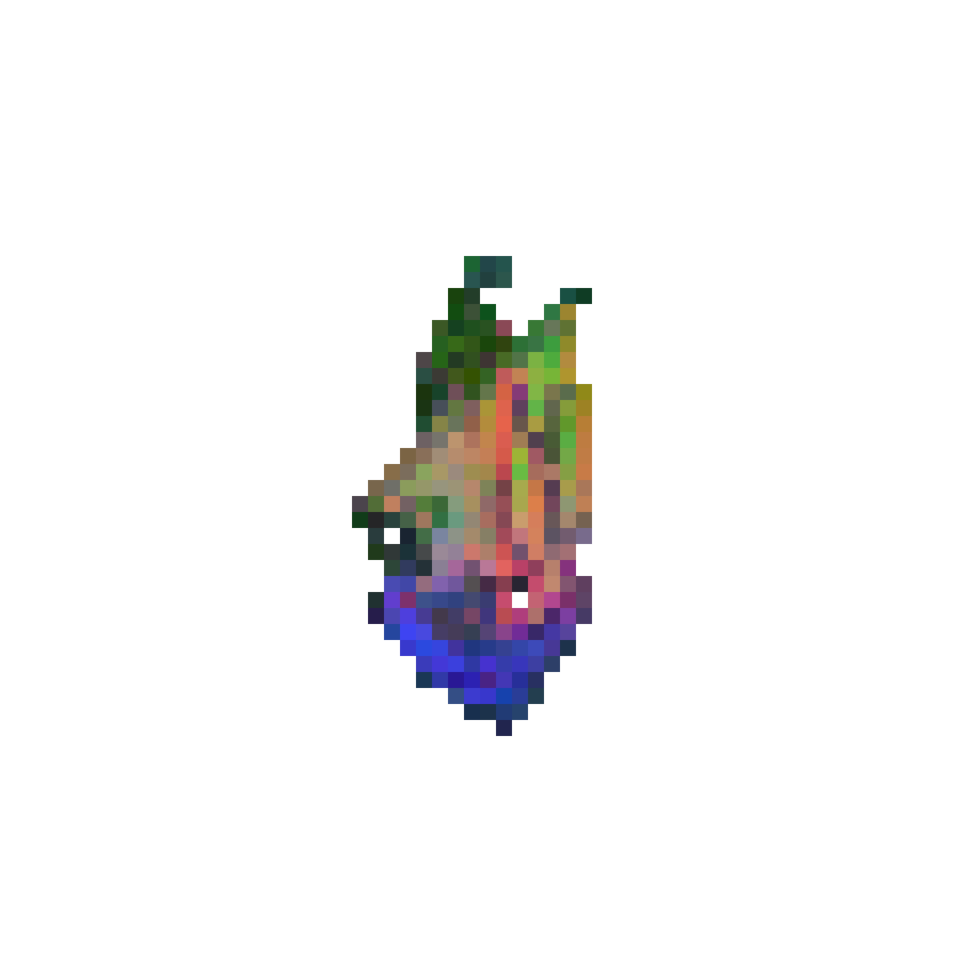}\hspace{0.5pt}%
    \zoomcrop{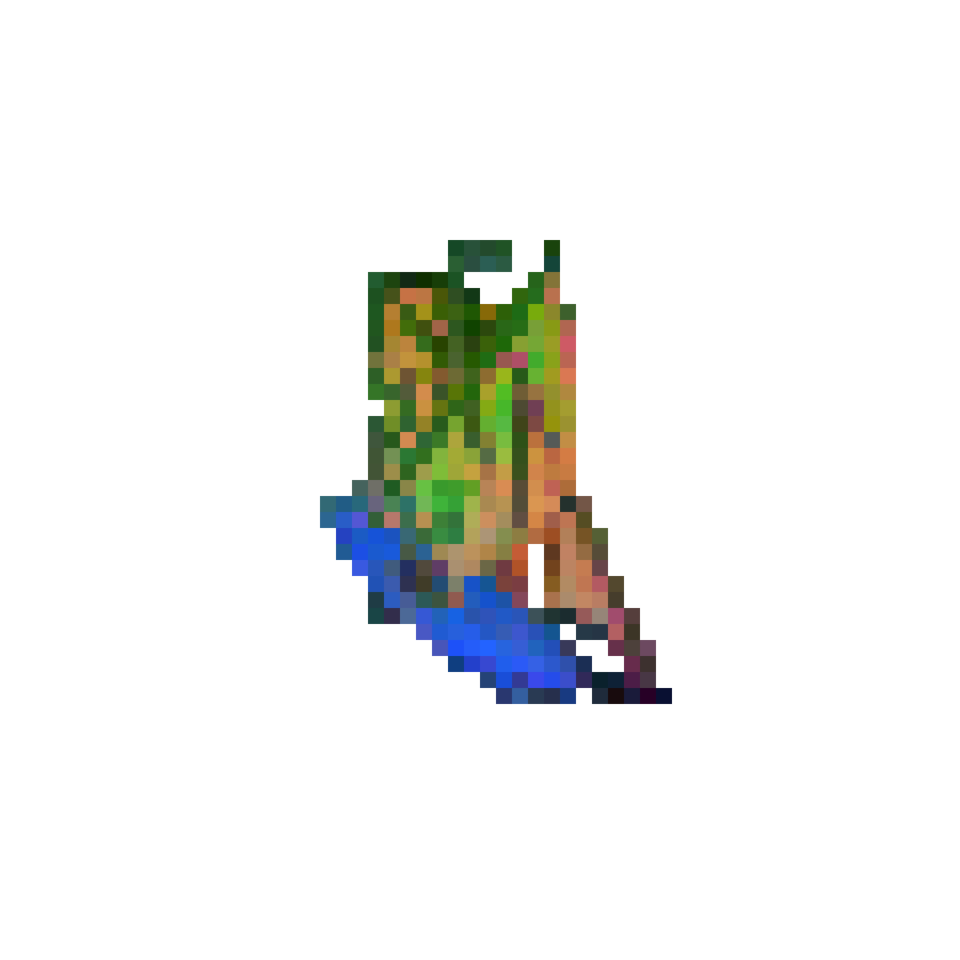}

    \vspace{4pt}

    \zoomcrop{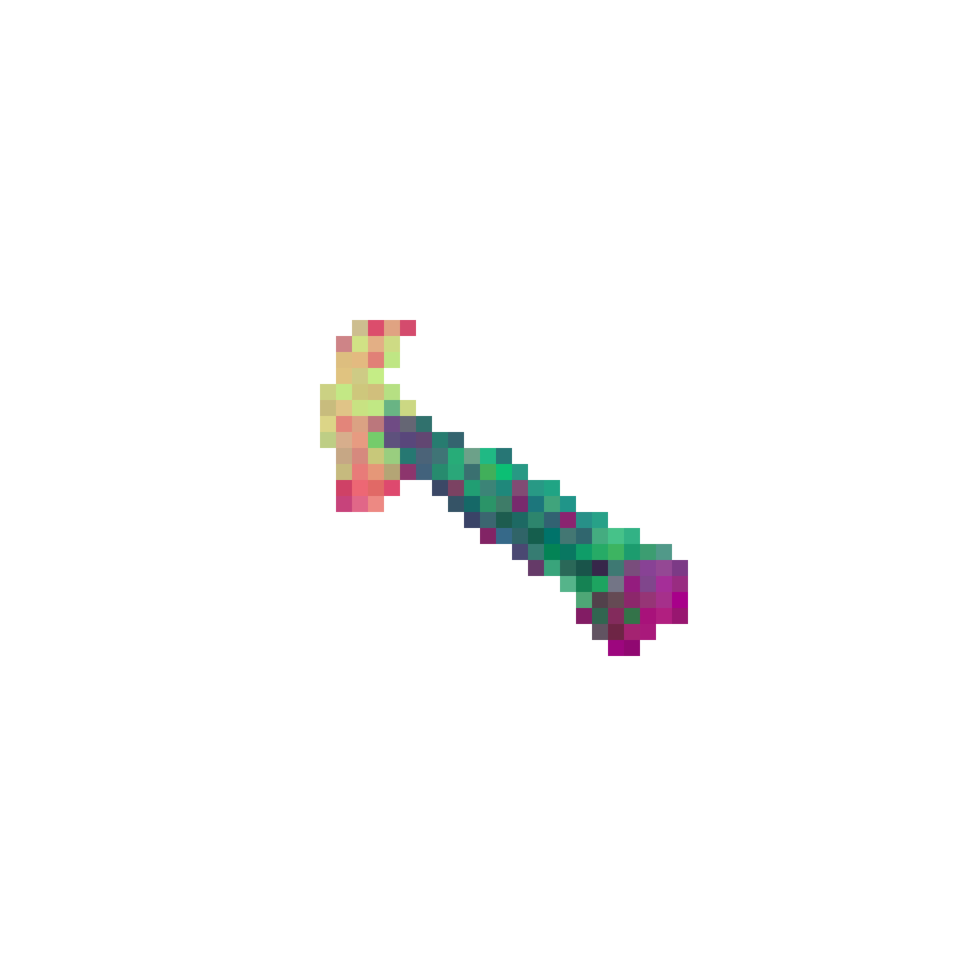}
    
    \zoomcrop{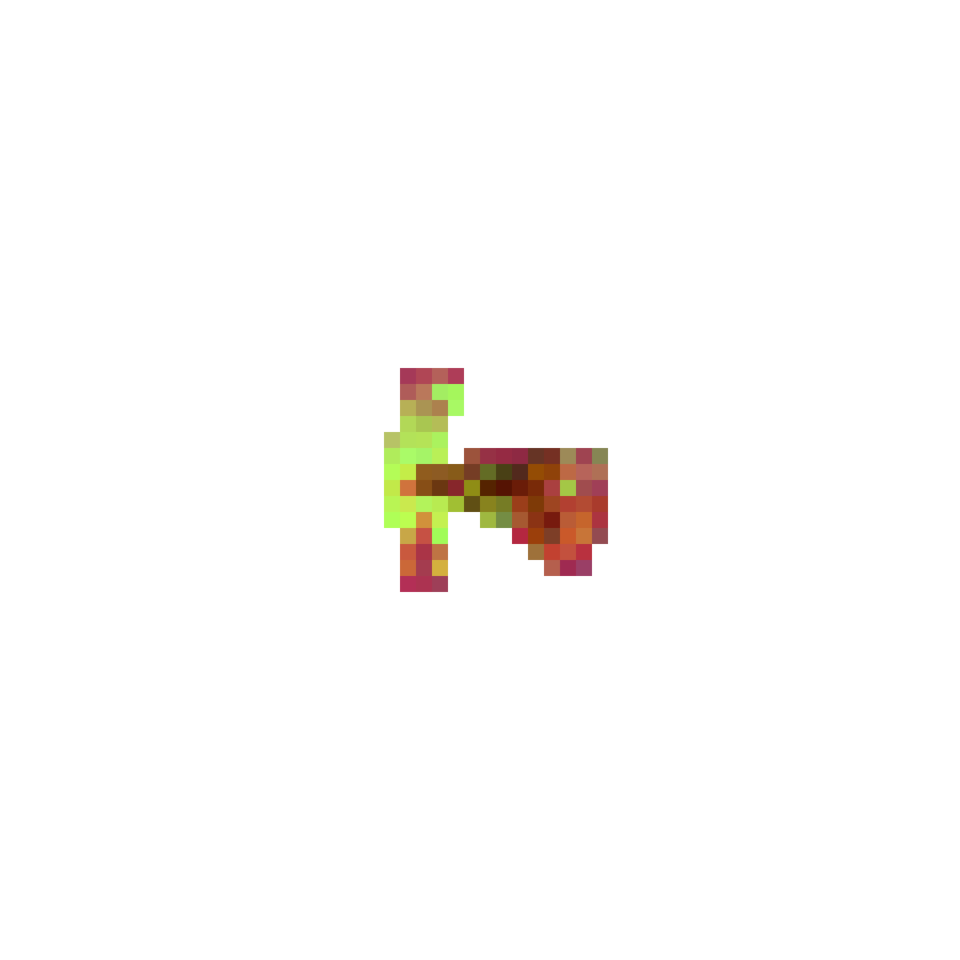}\hspace{0.5pt}%
    \zoomcrop{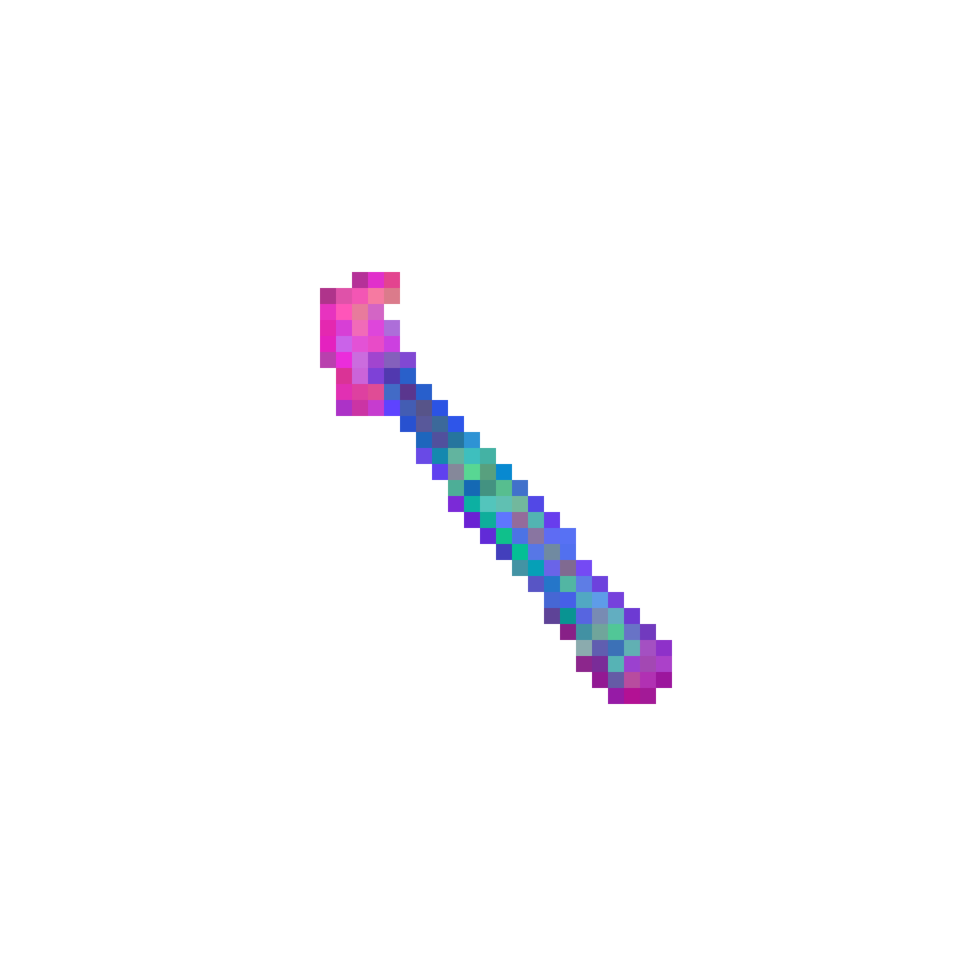}
    \captionsetup{skip=4pt}
    \caption{CLIP}
  \end{subfigure}
  \begin{subfigure}[t]{0.19\textwidth}
    \centering


    \zoomcrop{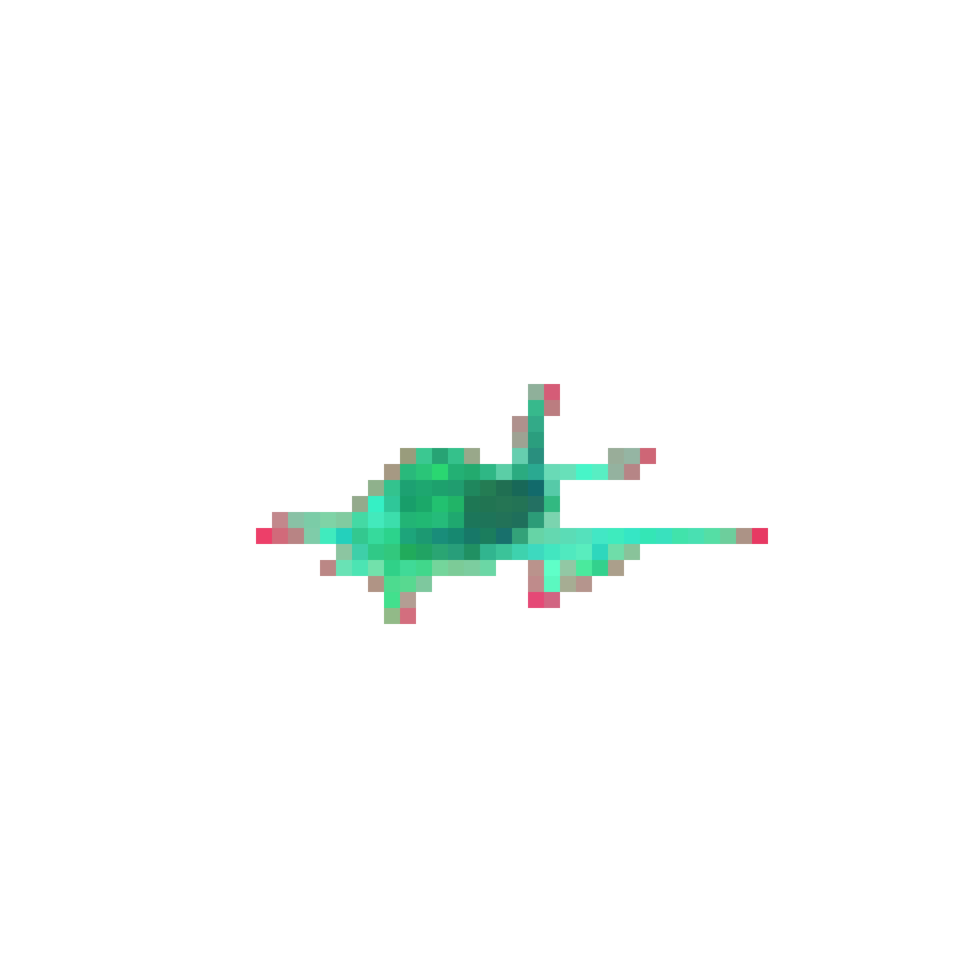}
    
    \zoomcrop{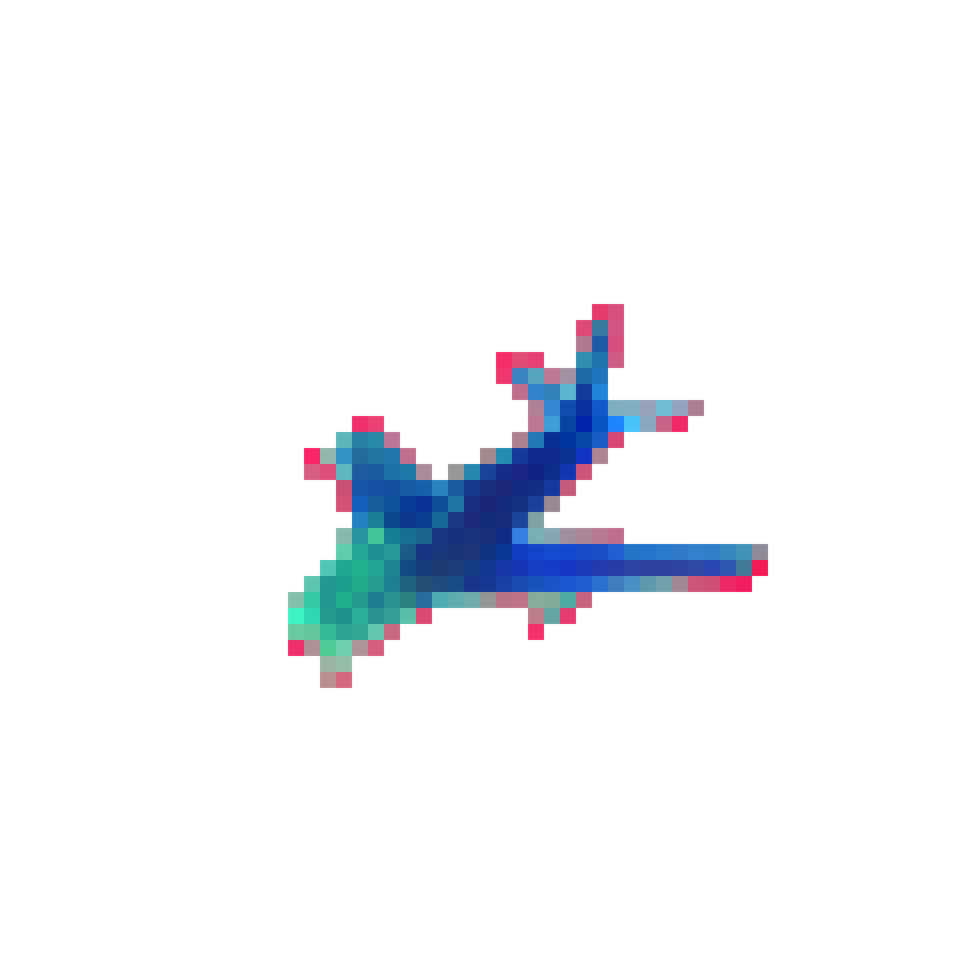}\hspace{0.5pt}%
    \zoomcrop{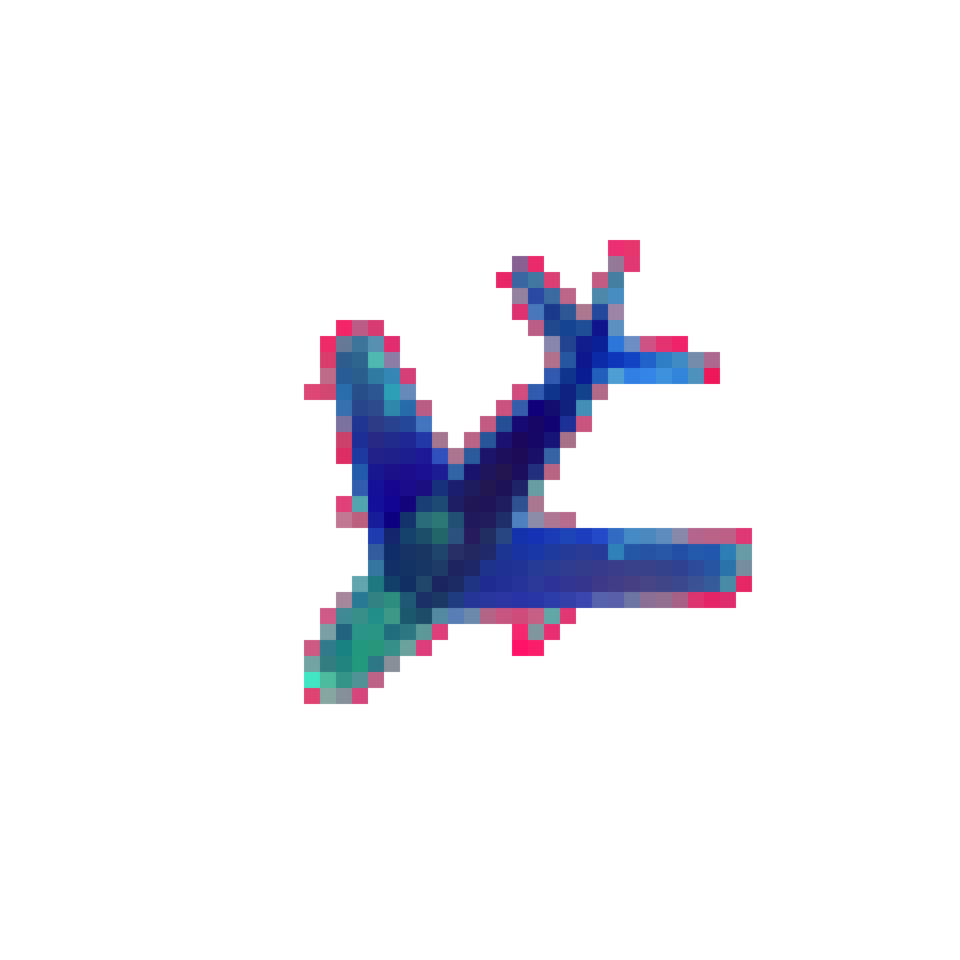}

    \vspace{4pt}

    \zoomcrop{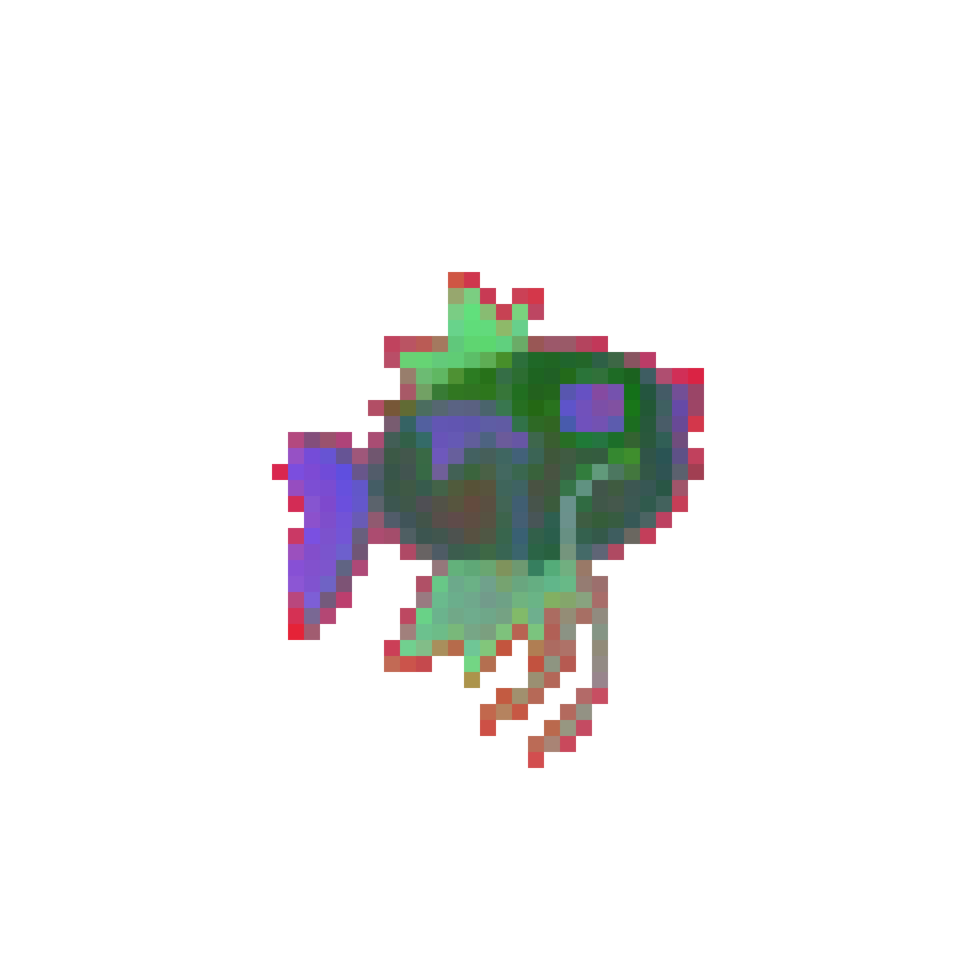}
    
    \zoomcrop{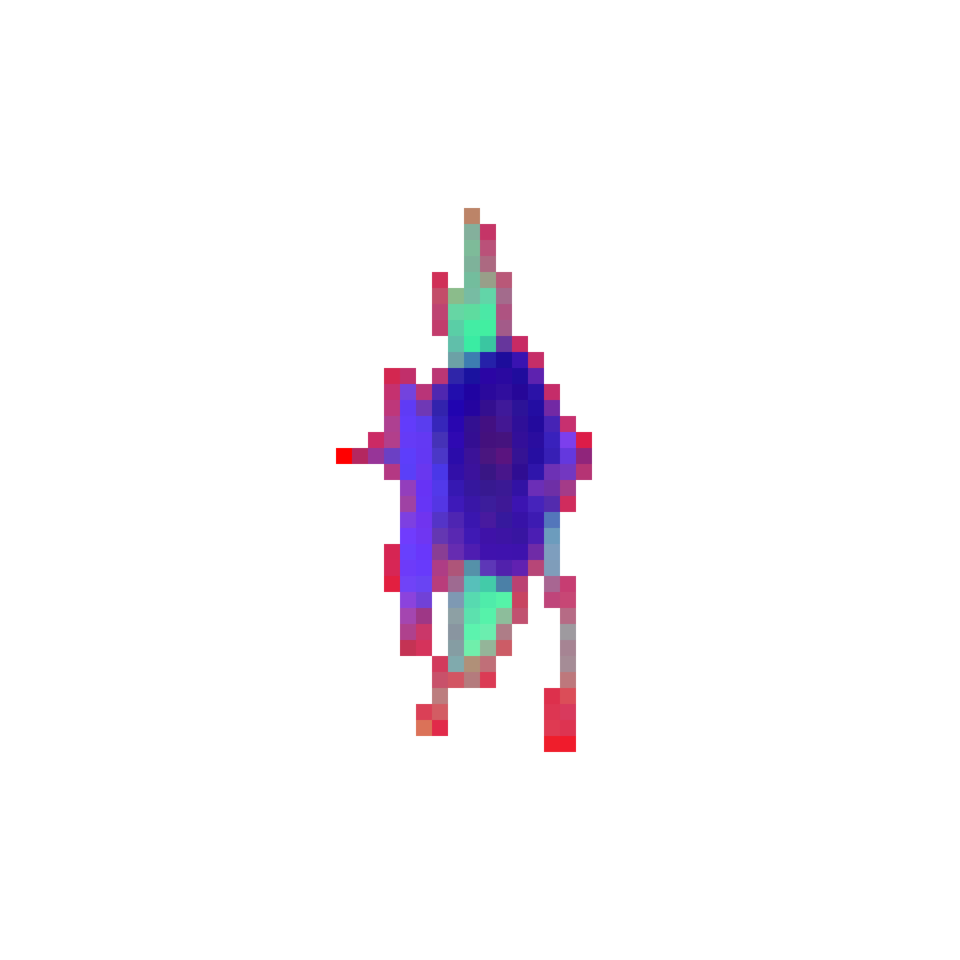}\hspace{0.5pt}%
    \zoomcrop{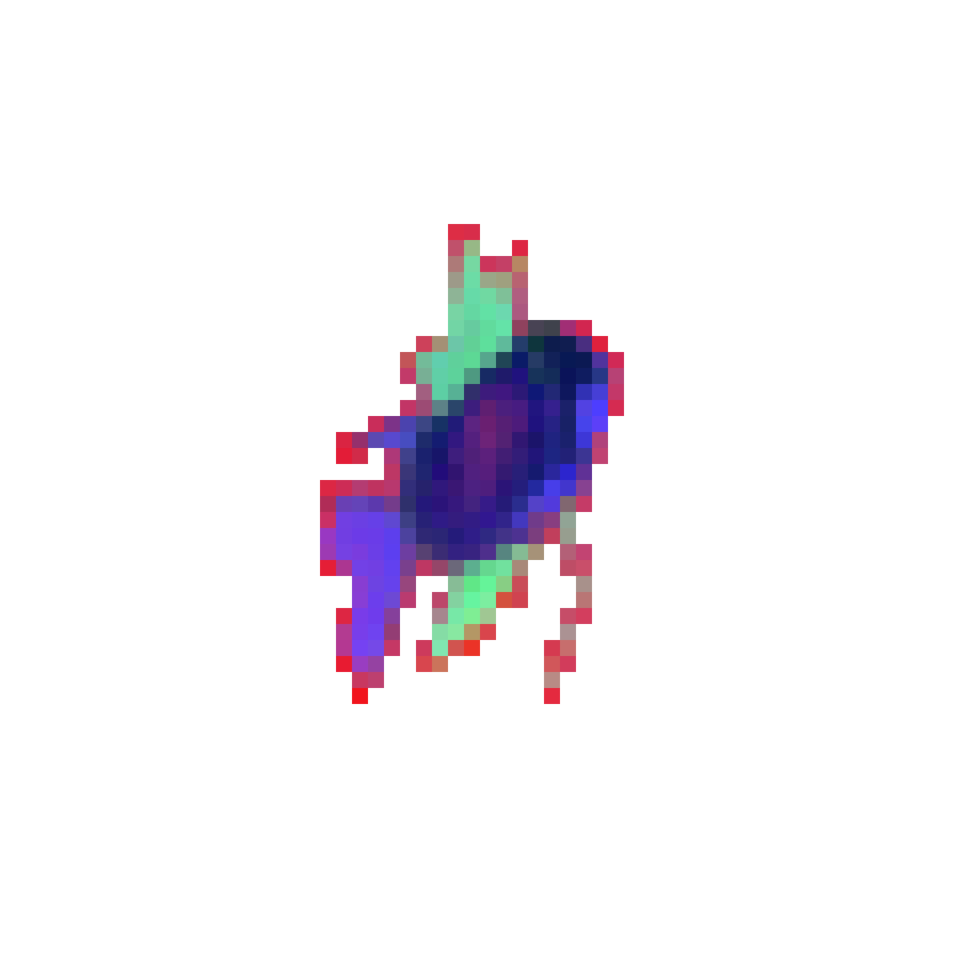}

    \vspace{4pt}

    \zoomcrop{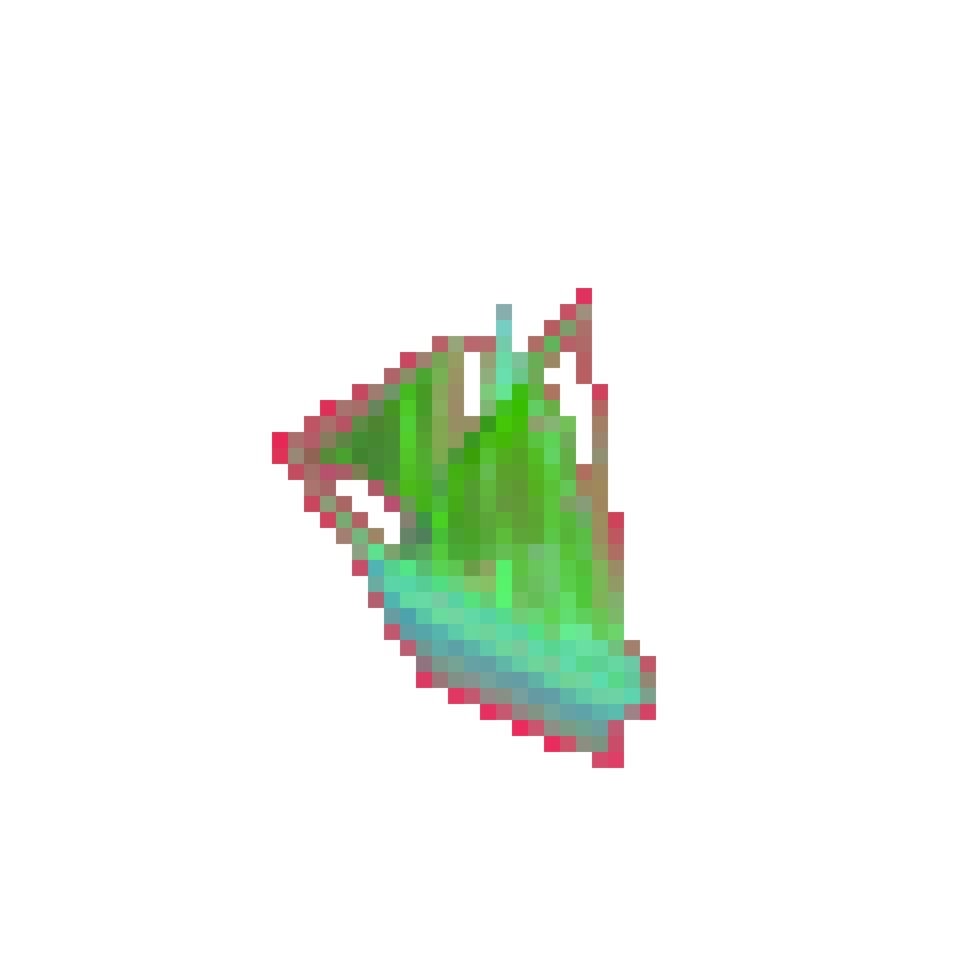}
    
    \zoomcrop{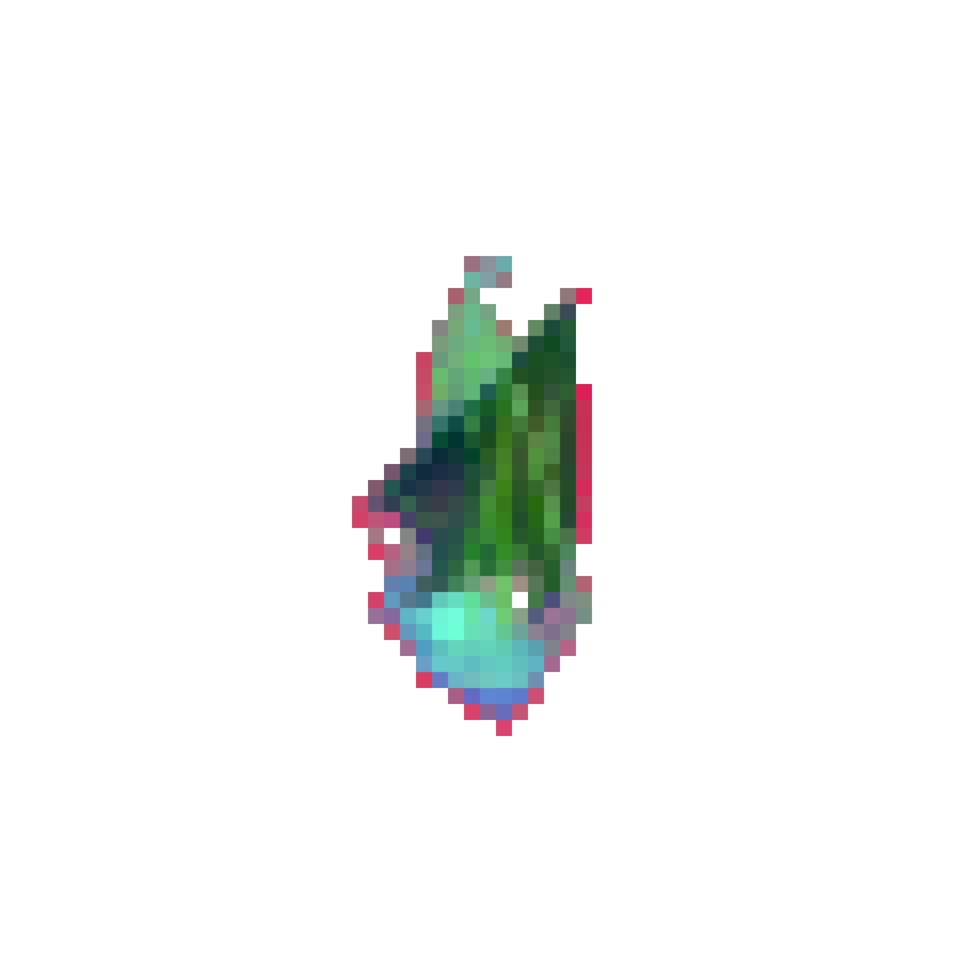}\hspace{0.5pt}%
    \zoomcrop{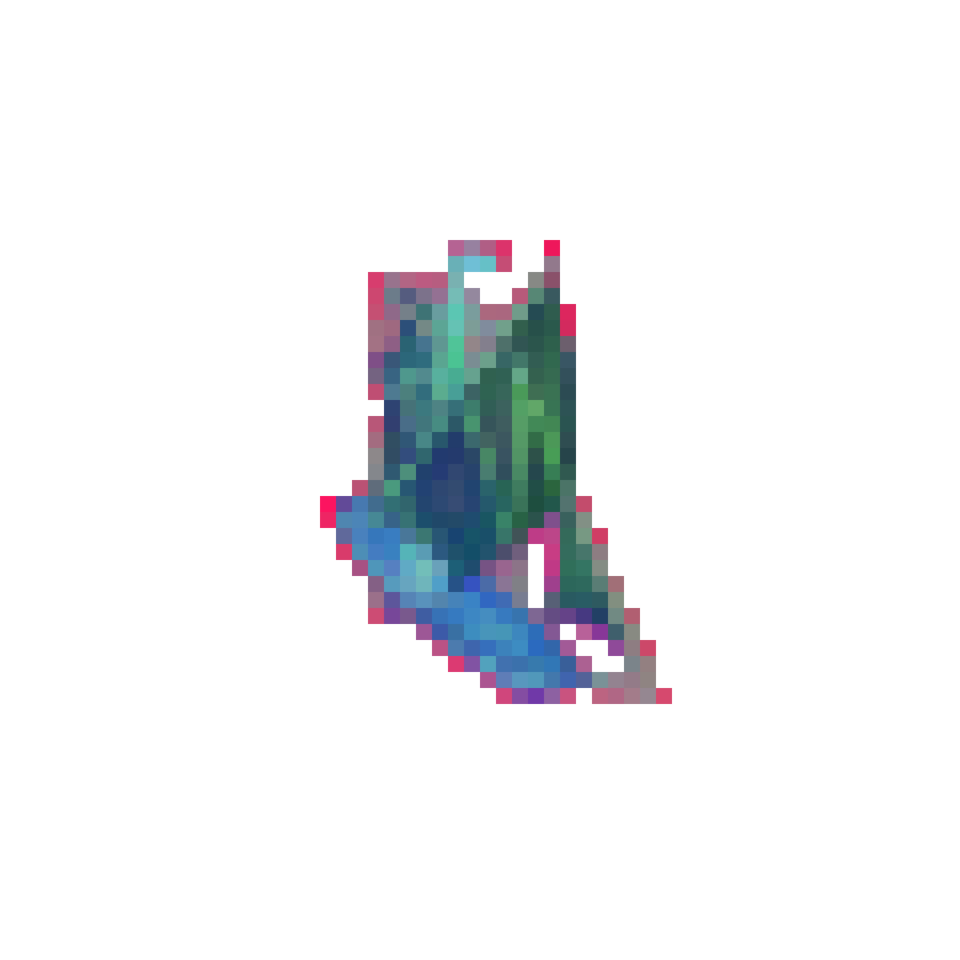}

    \vspace{4pt}

    \zoomcrop{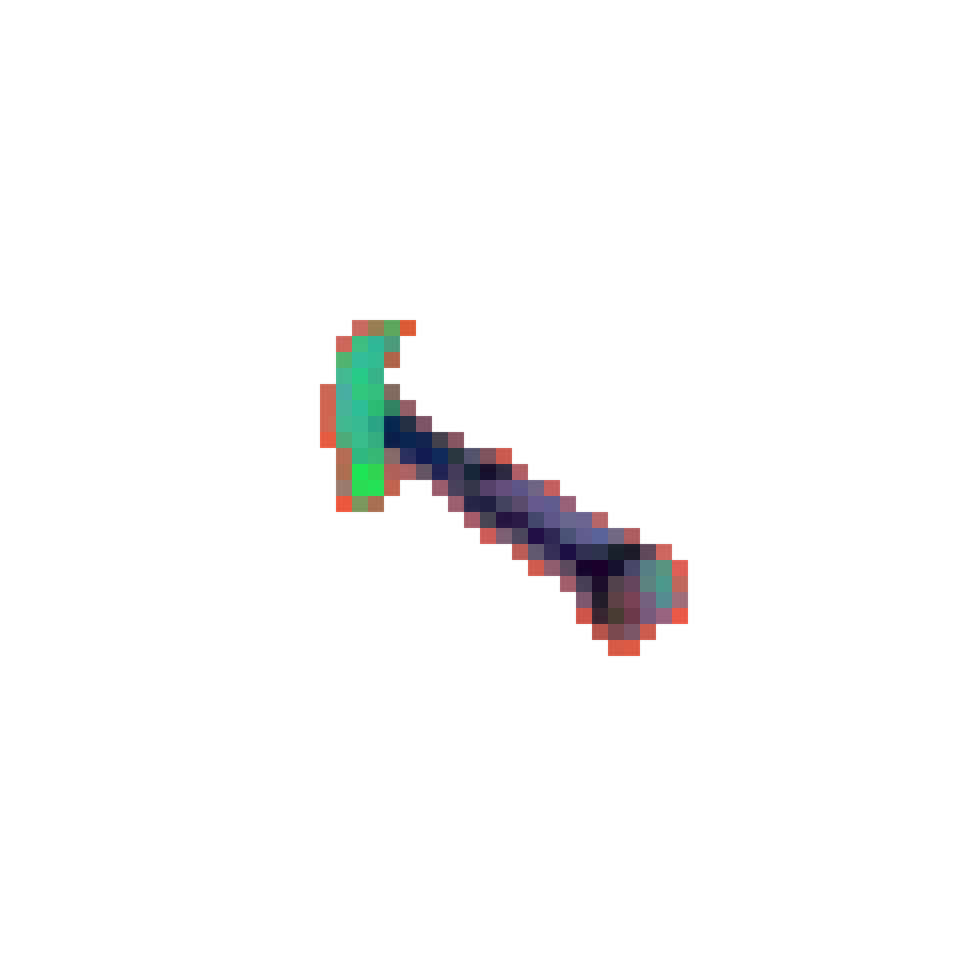}

    \zoomcrop{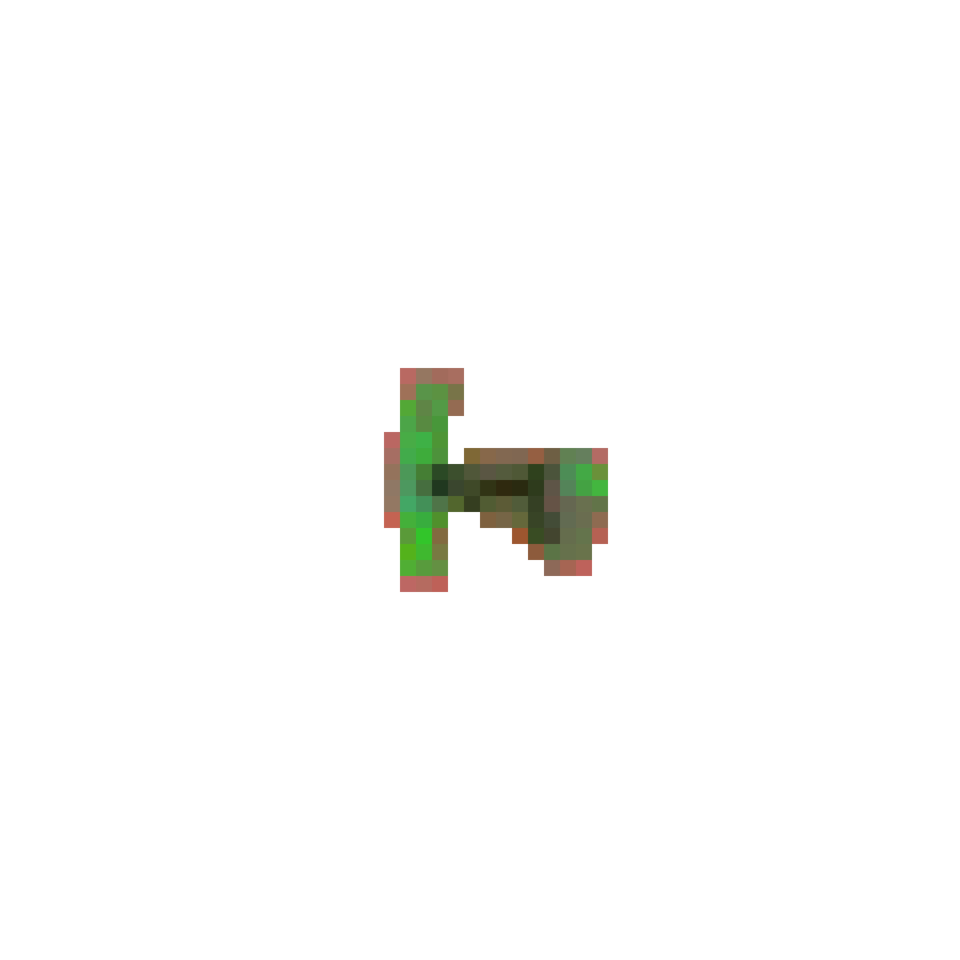}\hspace{0.5pt}%
    \zoomcrop{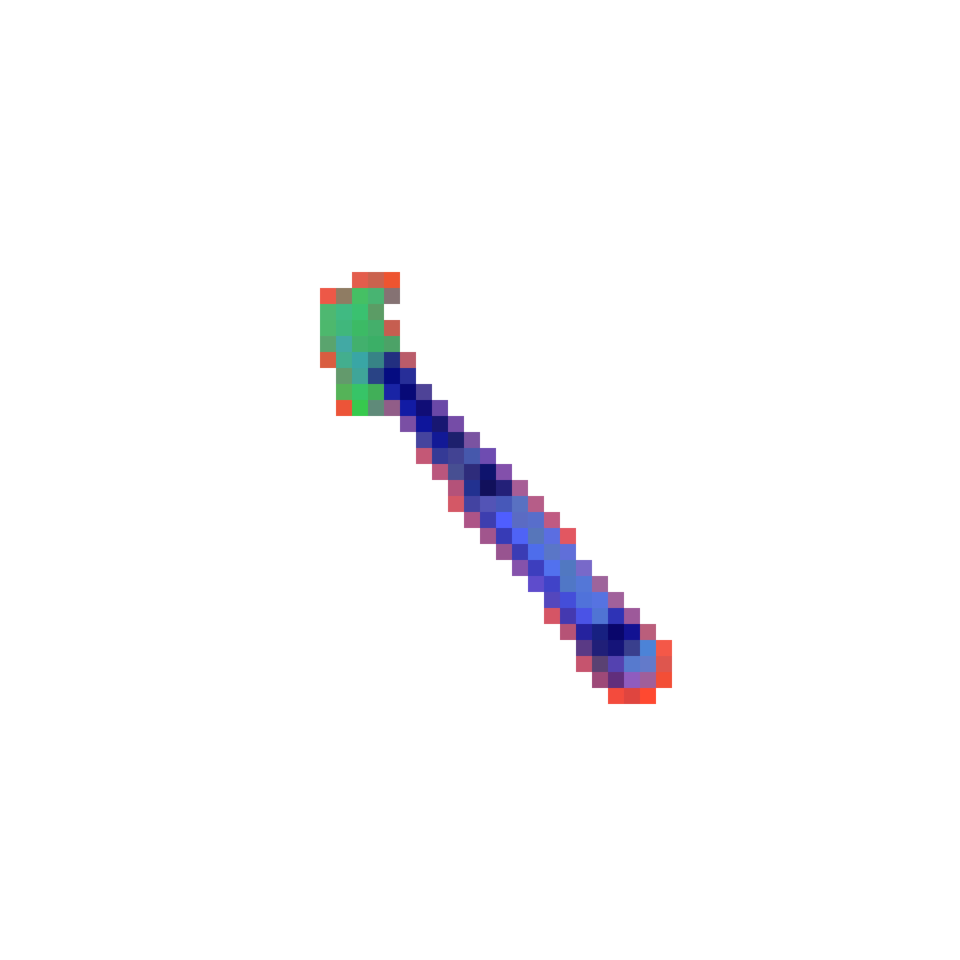}
    \captionsetup{skip=4pt}
    \caption{FiT3D}
  \end{subfigure}
  \begin{subfigure}[t]{0.19\textwidth}
    \centering



    \zoomcrop{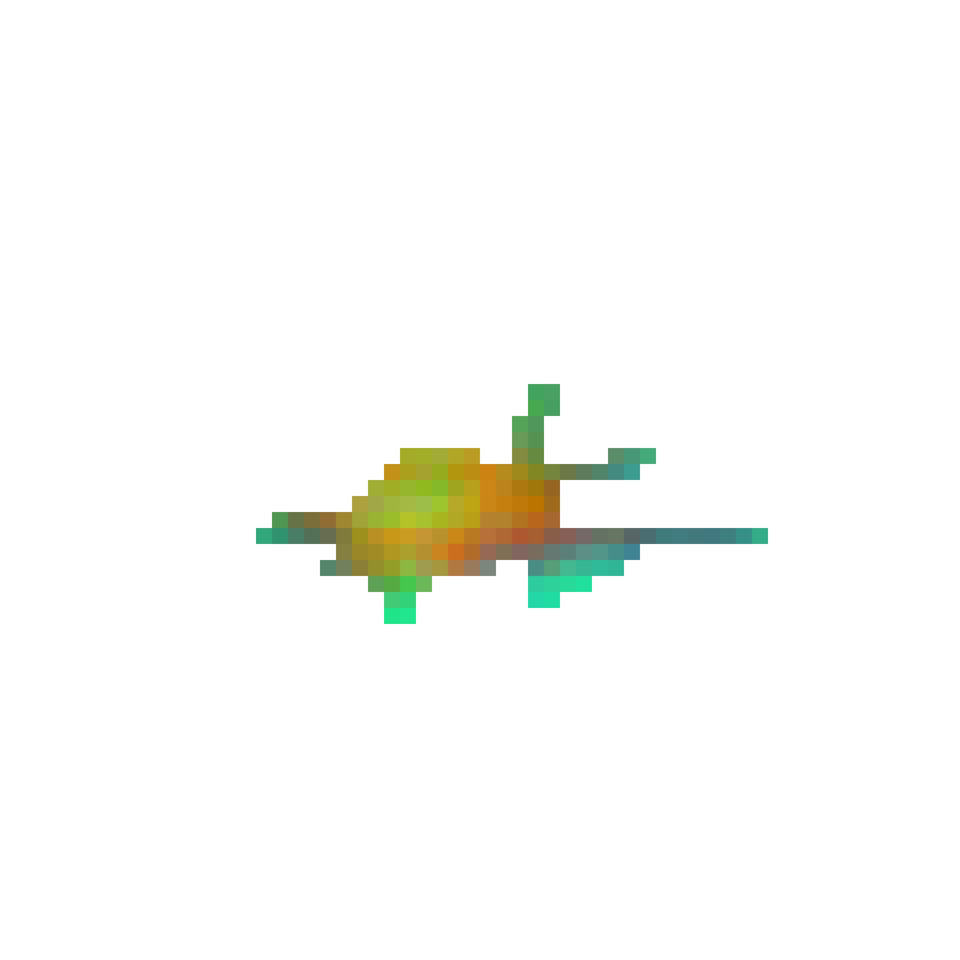}
    
    \zoomcrop{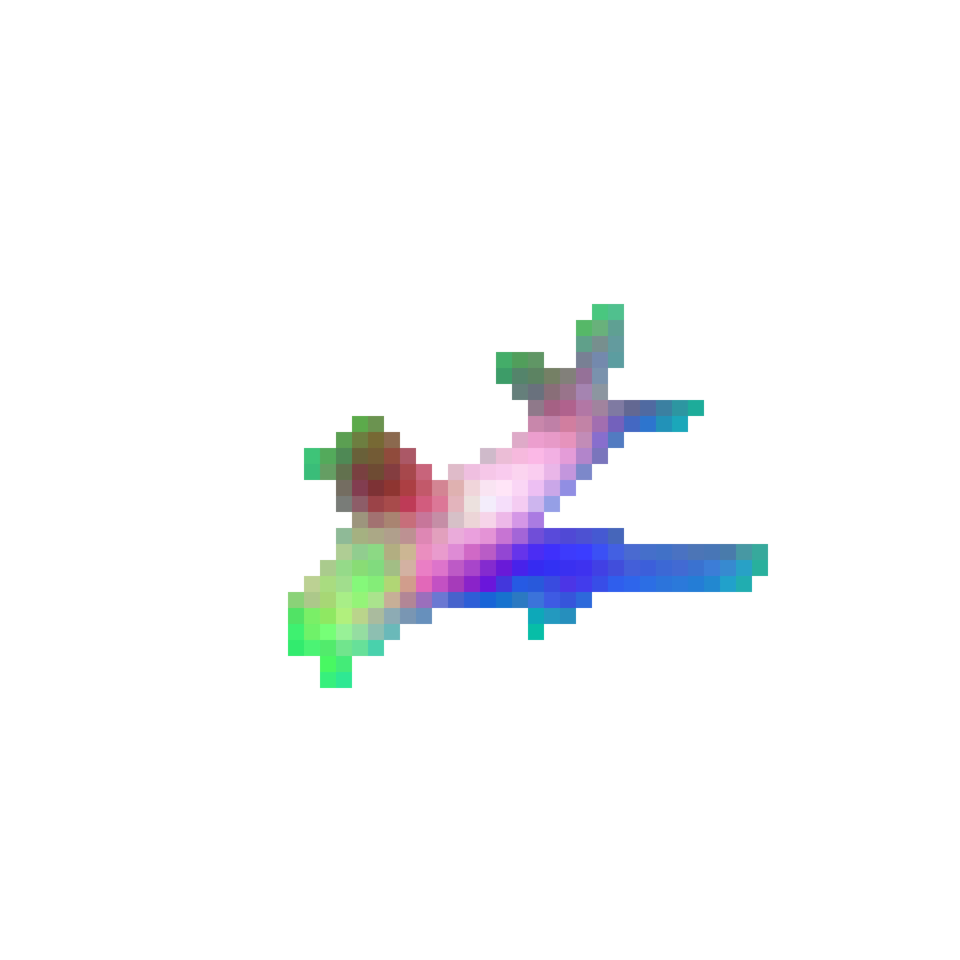}\hspace{0.5pt}%
    \zoomcrop{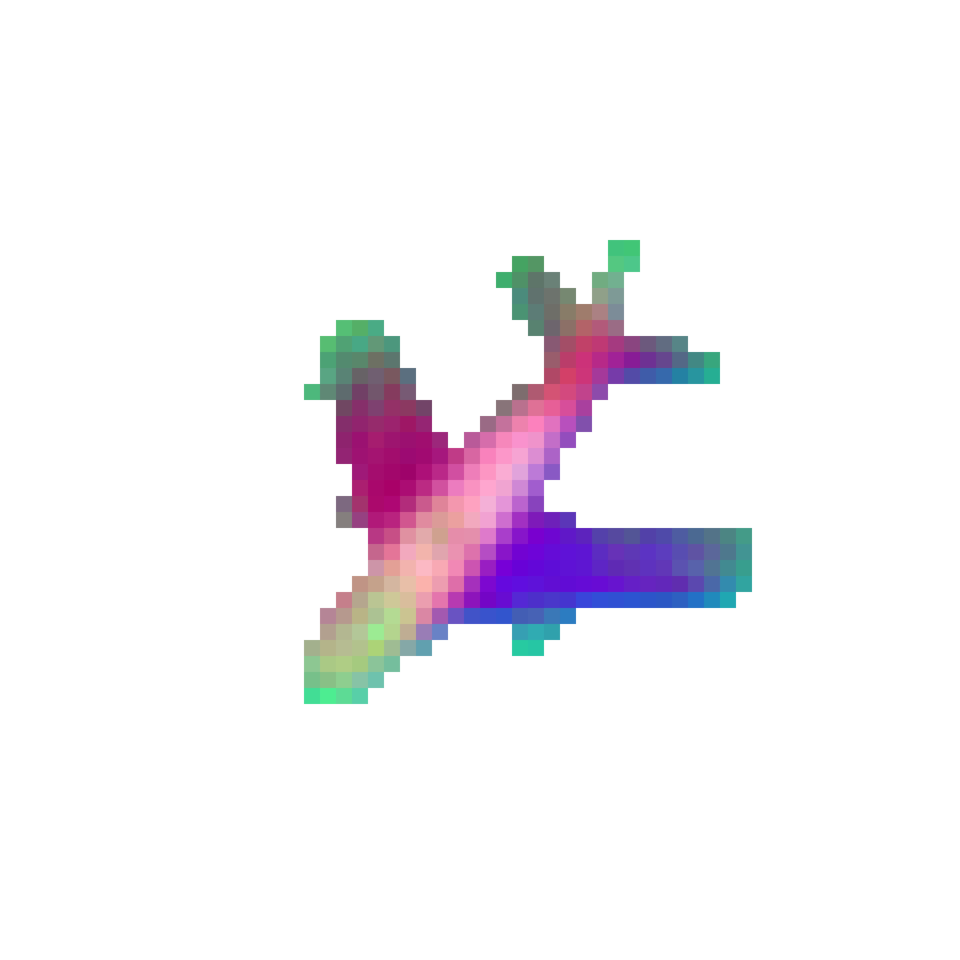}
    
    \vspace{4pt}

    \zoomcrop{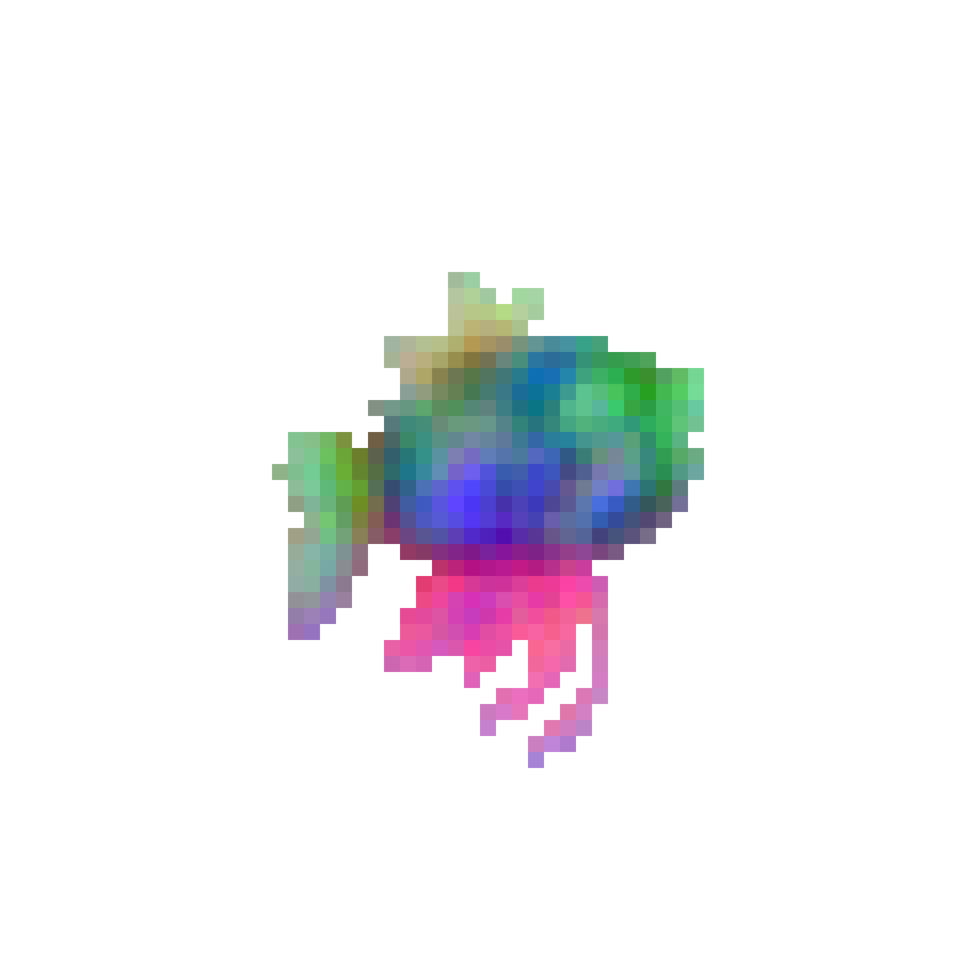}
    
    \zoomcrop{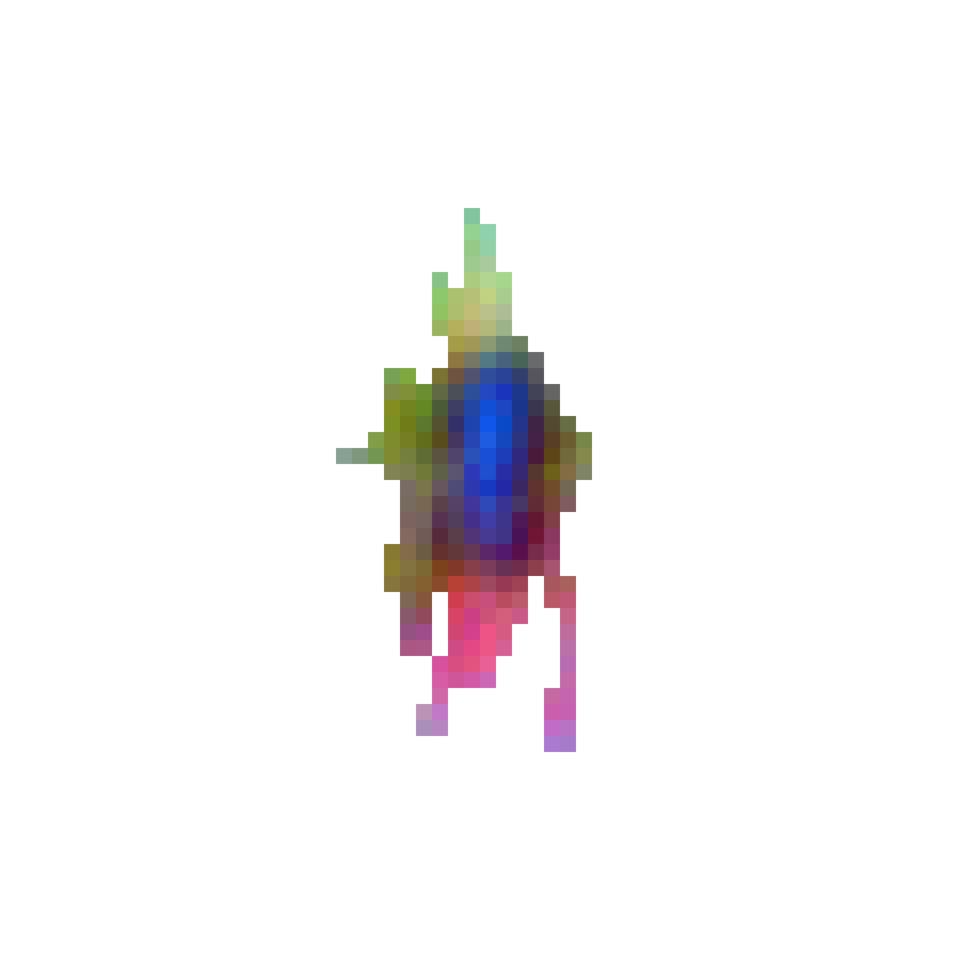}\hspace{0.5pt}%
    \zoomcrop{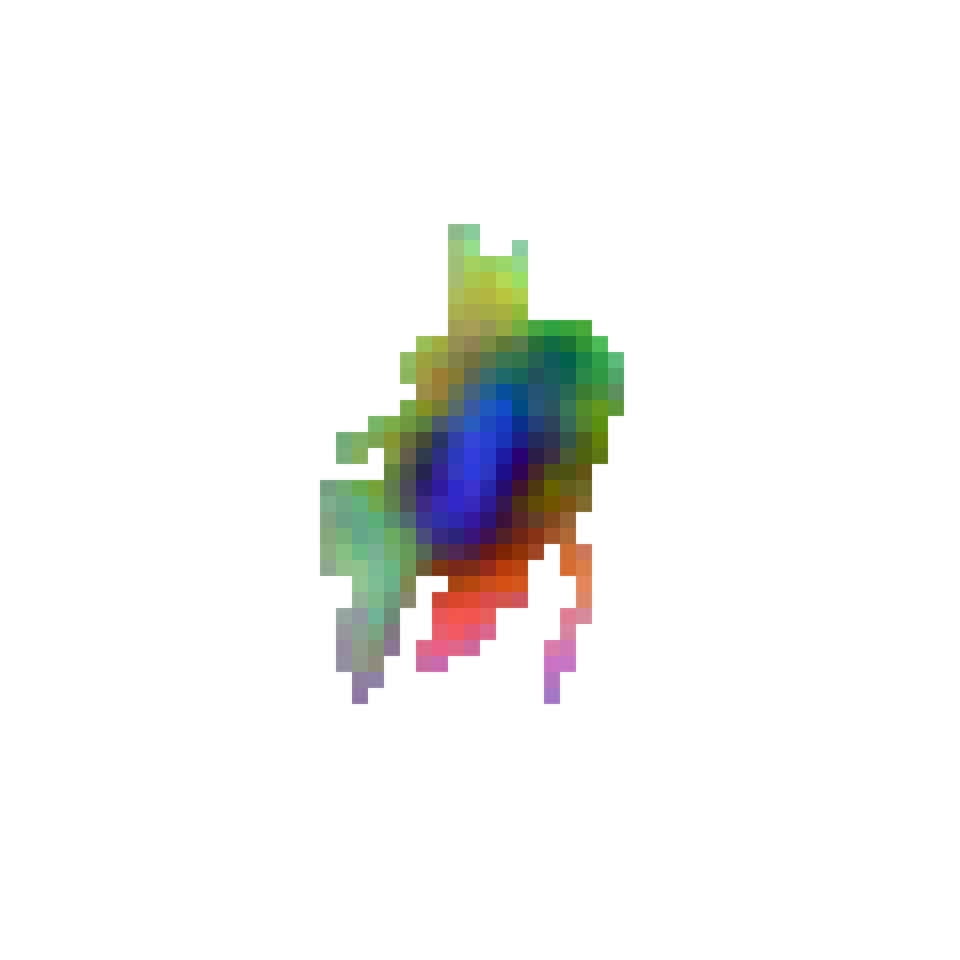}
    
    \vspace{4pt}

    \zoomcrop{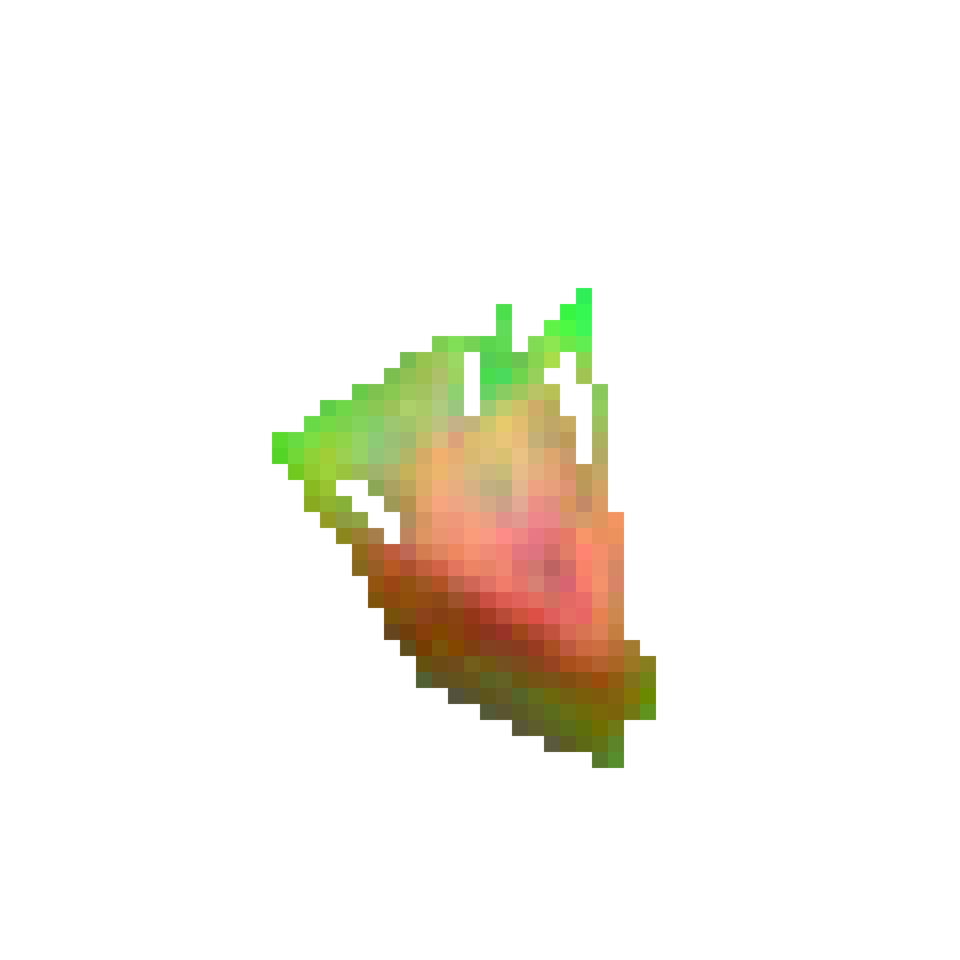}
    
    \zoomcrop{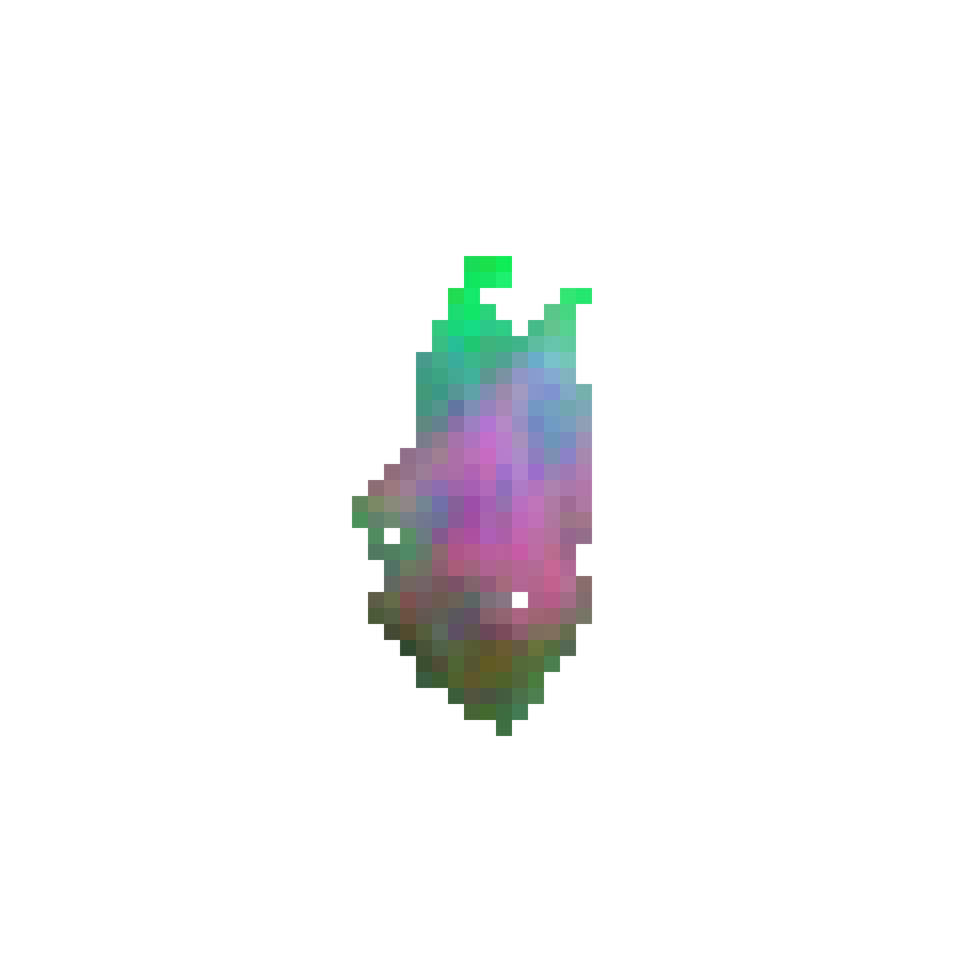}\hspace{0.5pt}%
    \zoomcrop{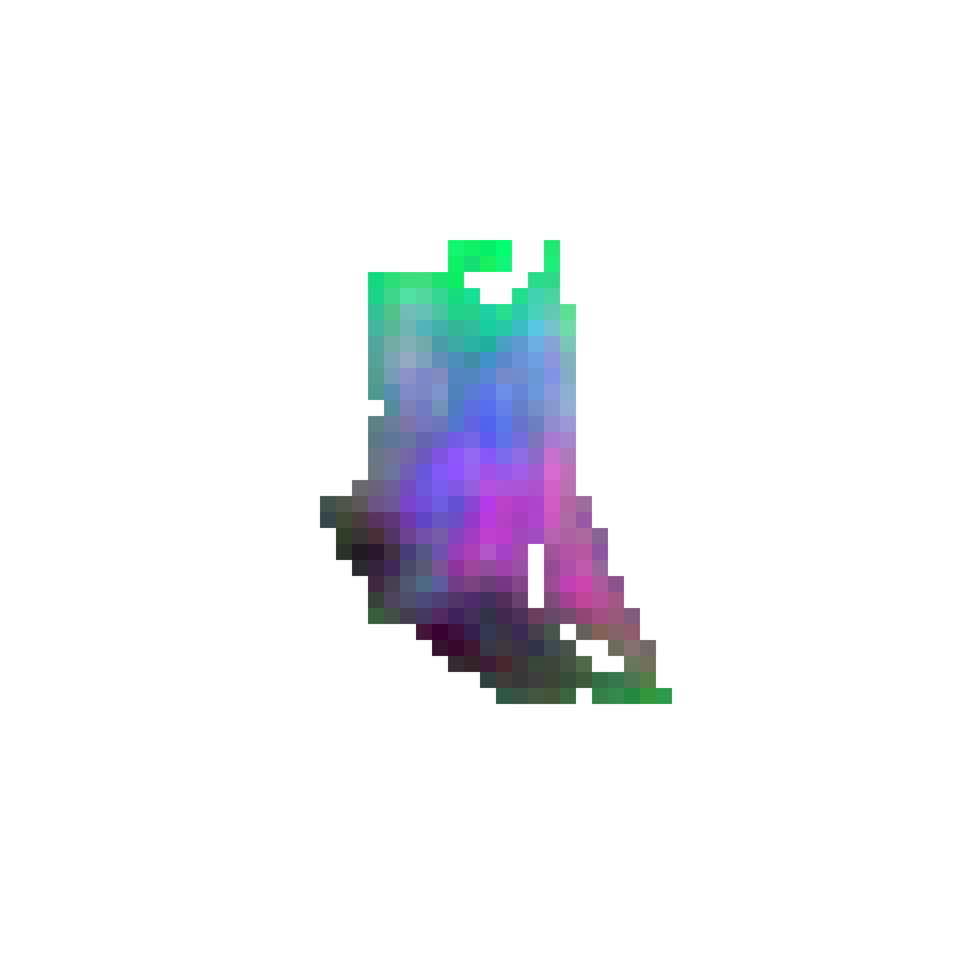}

    \vspace{4pt}
    
    \zoomcrop{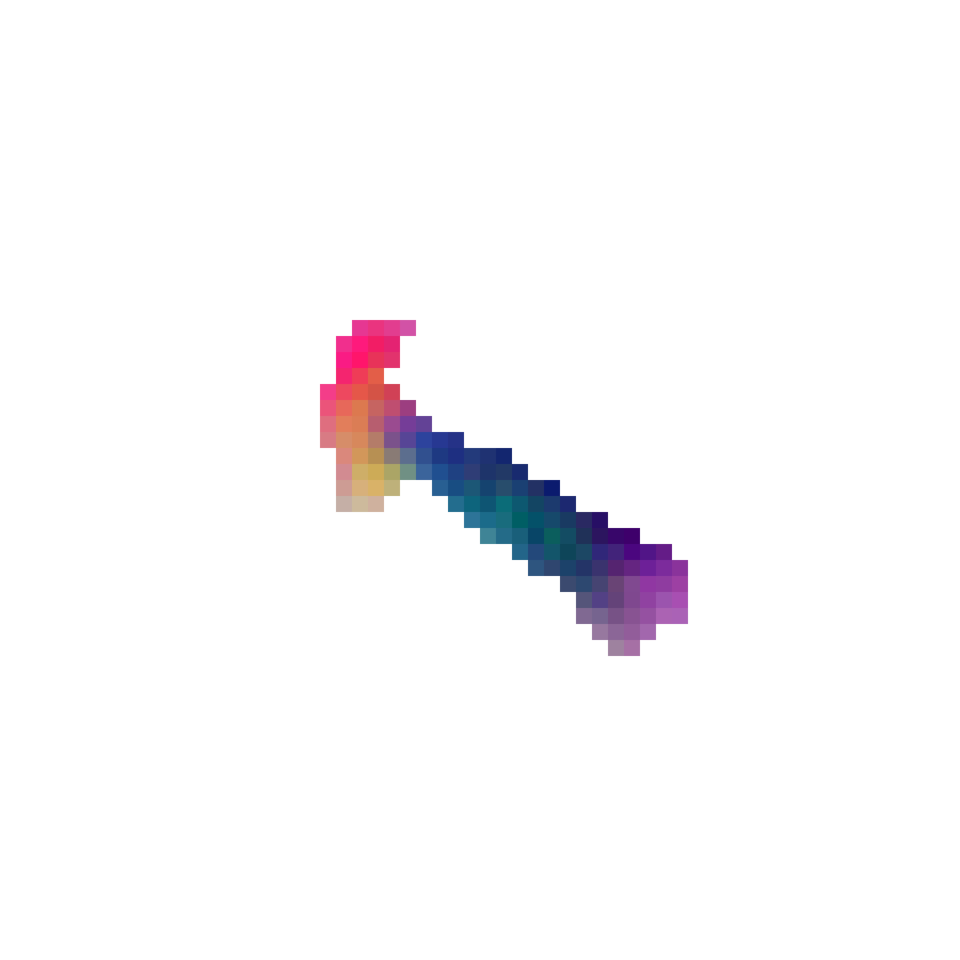}
    
    \zoomcrop{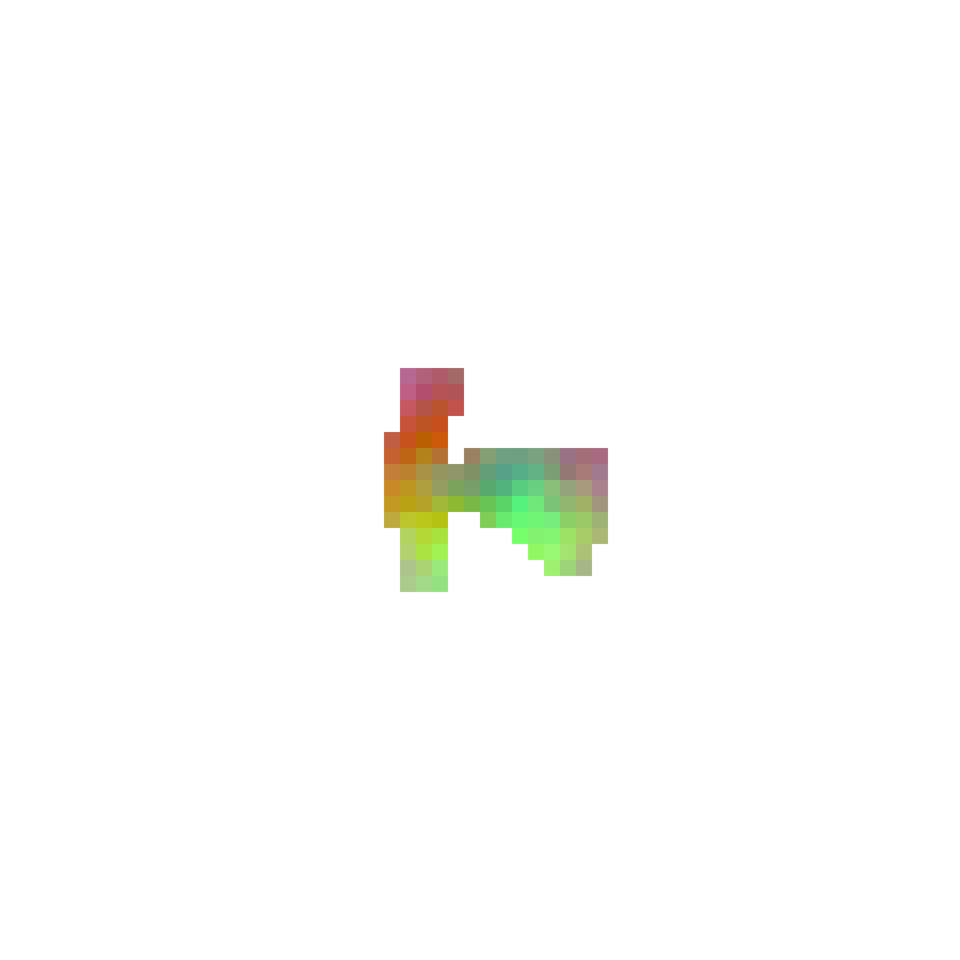}\hspace{0.5pt}%
    \zoomcrop{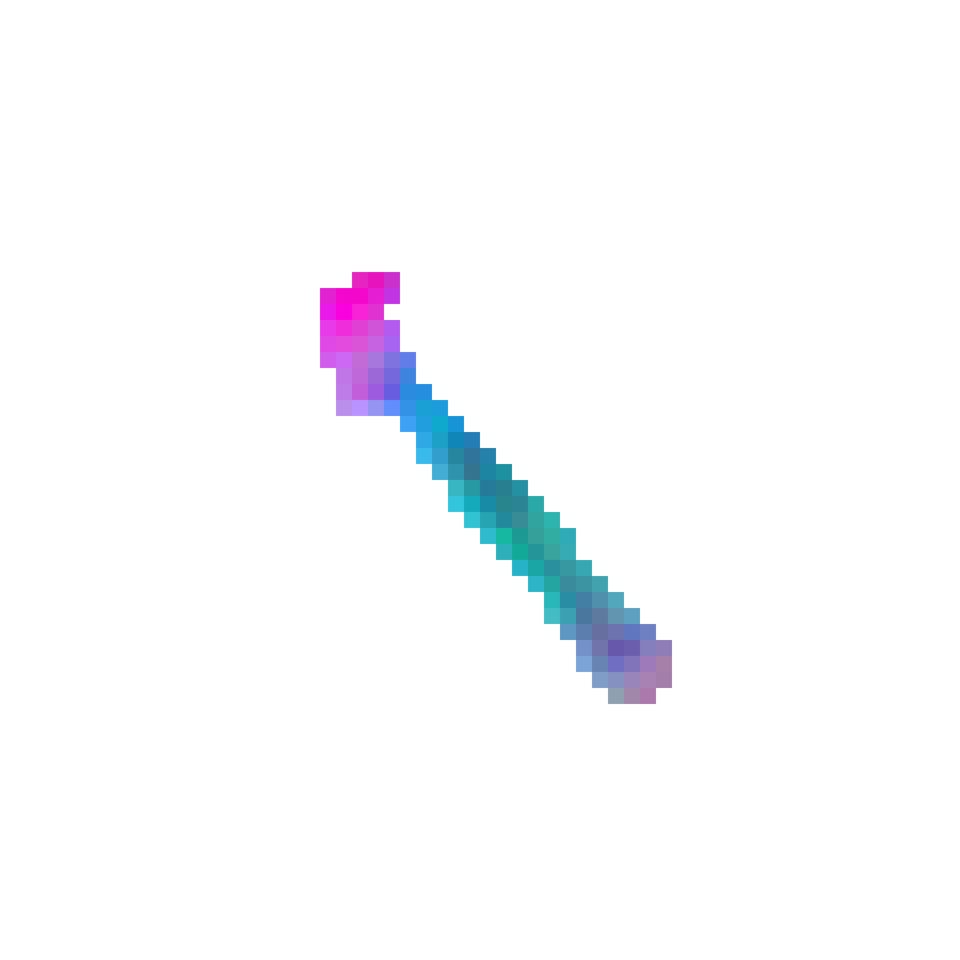}
    \captionsetup{skip=4pt}
    \caption{MEF}
  \end{subfigure}
  \begin{subfigure}[t]{0.19\textwidth}
    \centering



    \zoomcrop{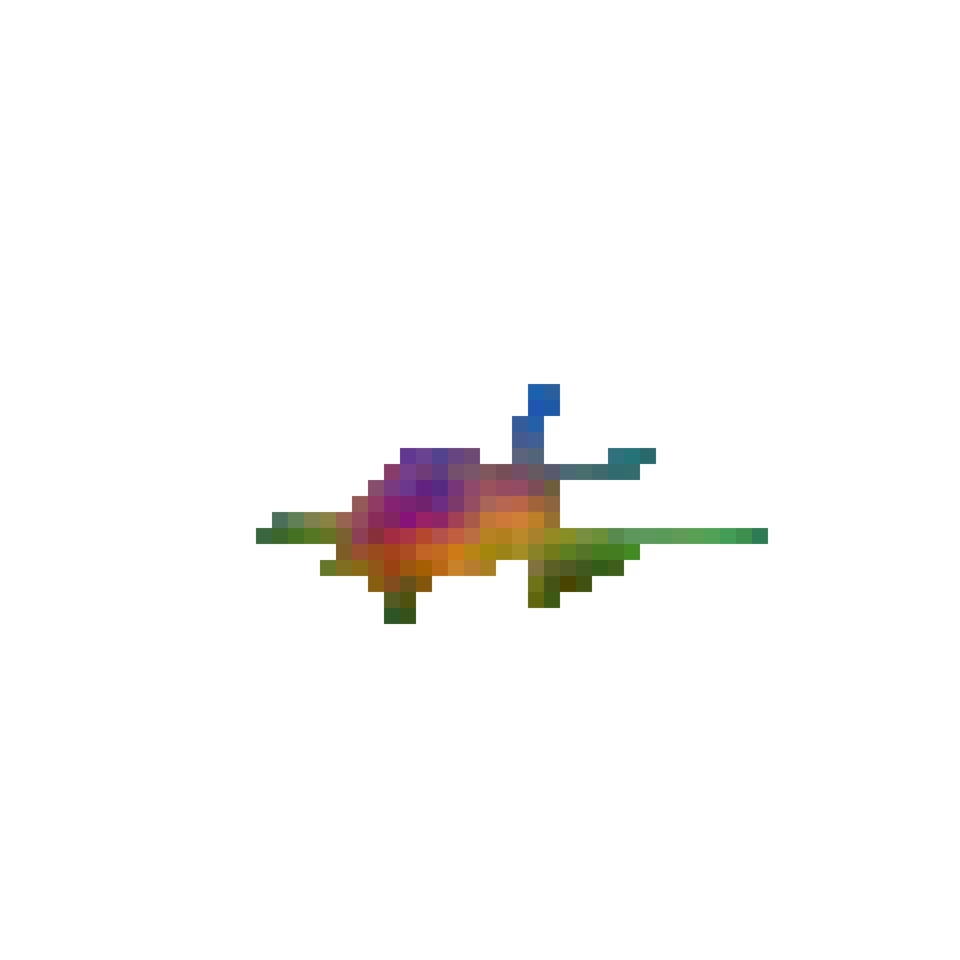}
    
    \zoomcrop{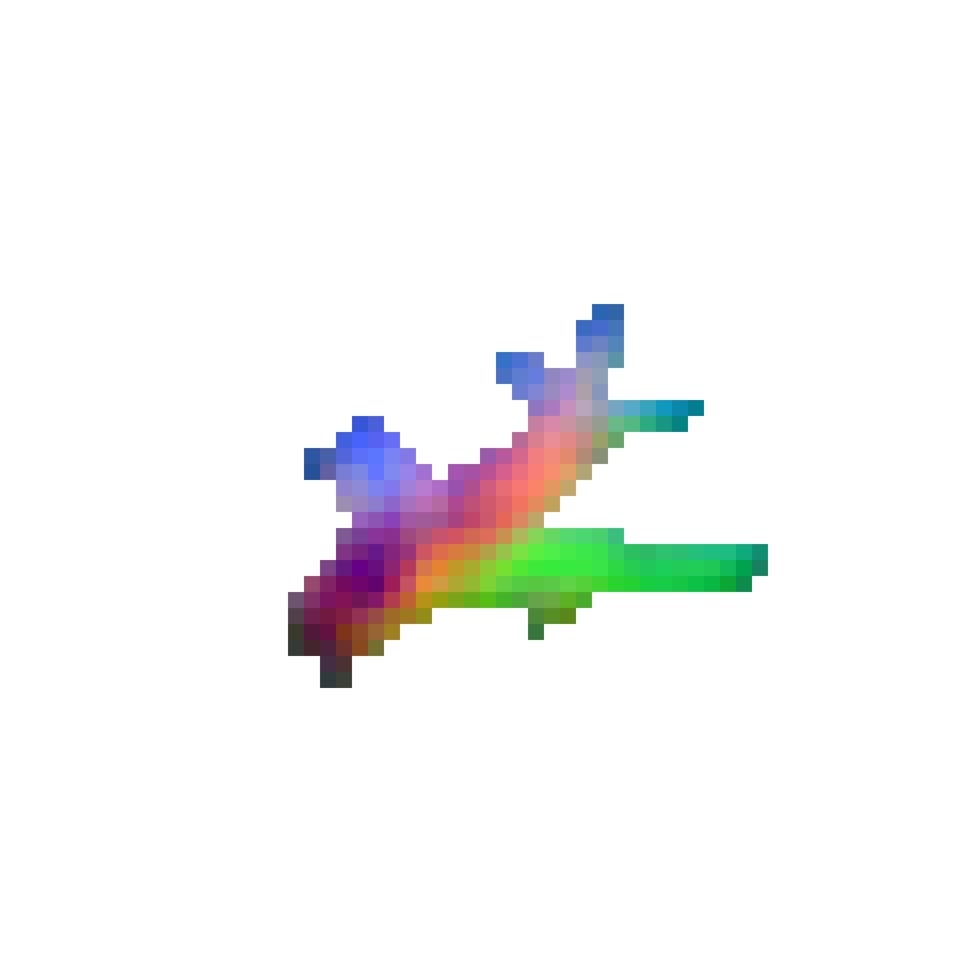}\hspace{0.5pt}%
    \zoomcrop{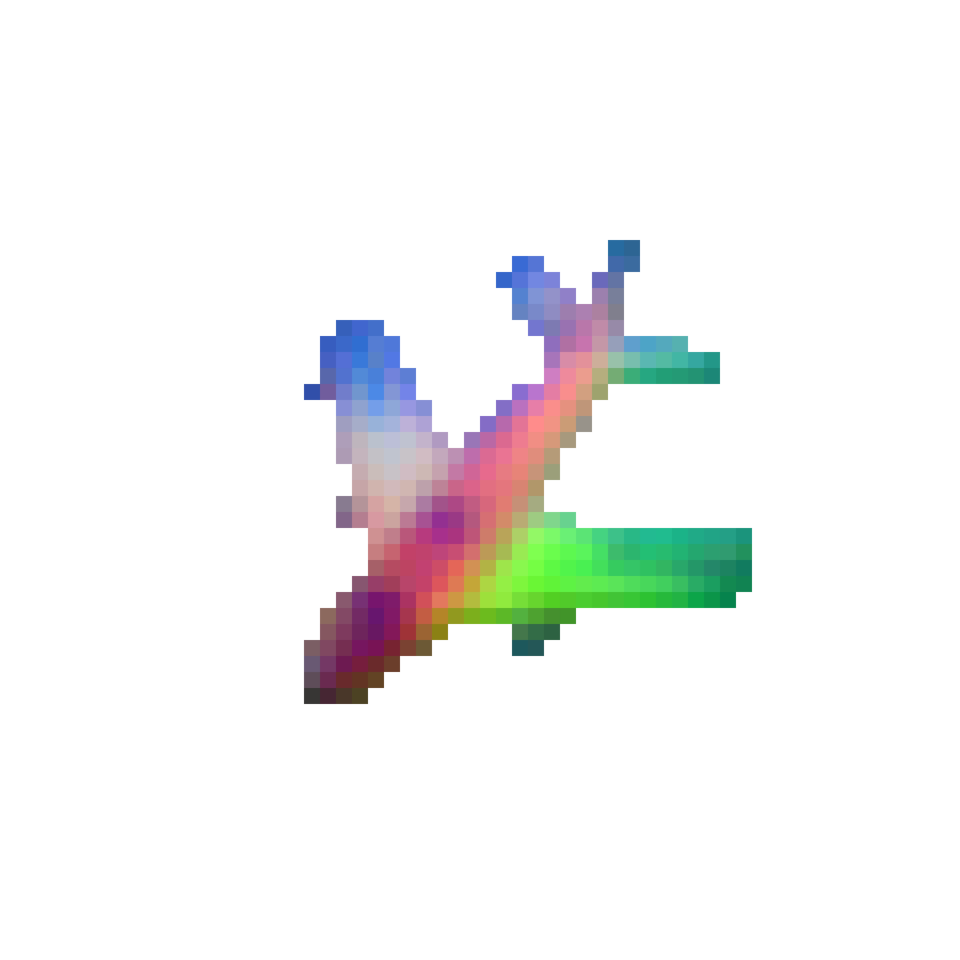}

    \vspace{4pt}

    \zoomcrop{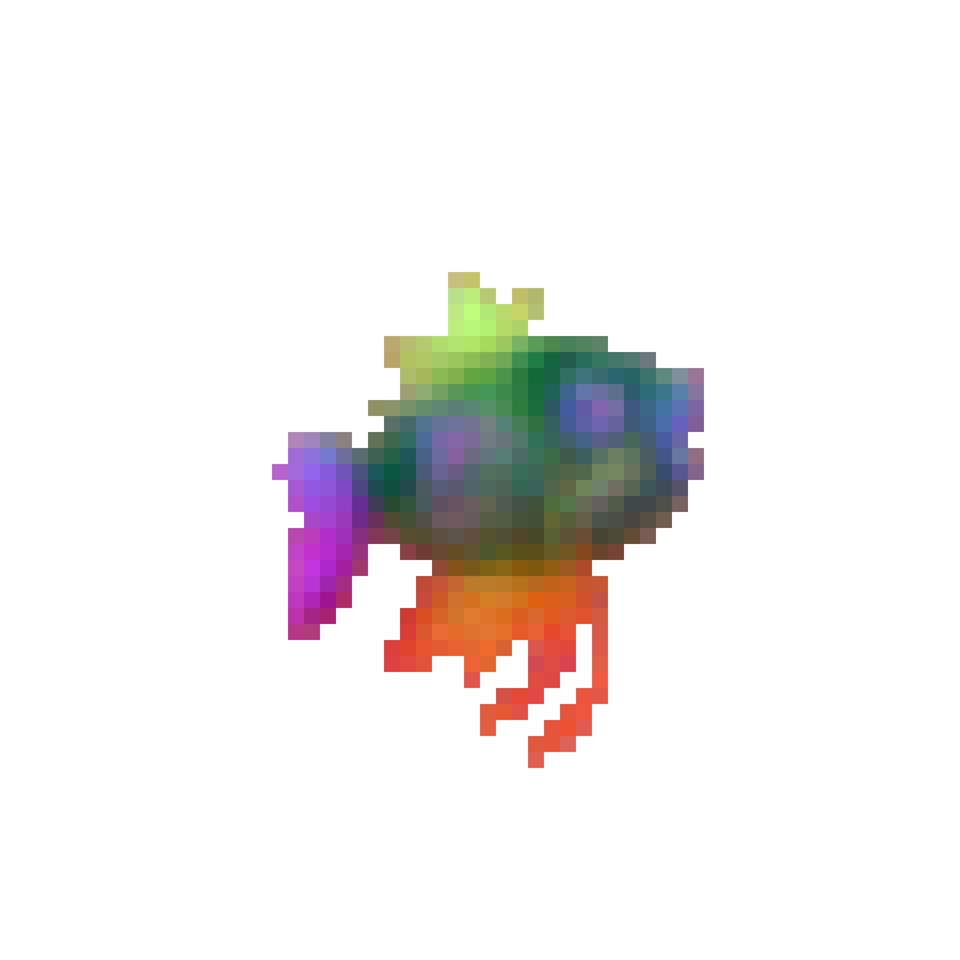}
    
    \zoomcrop{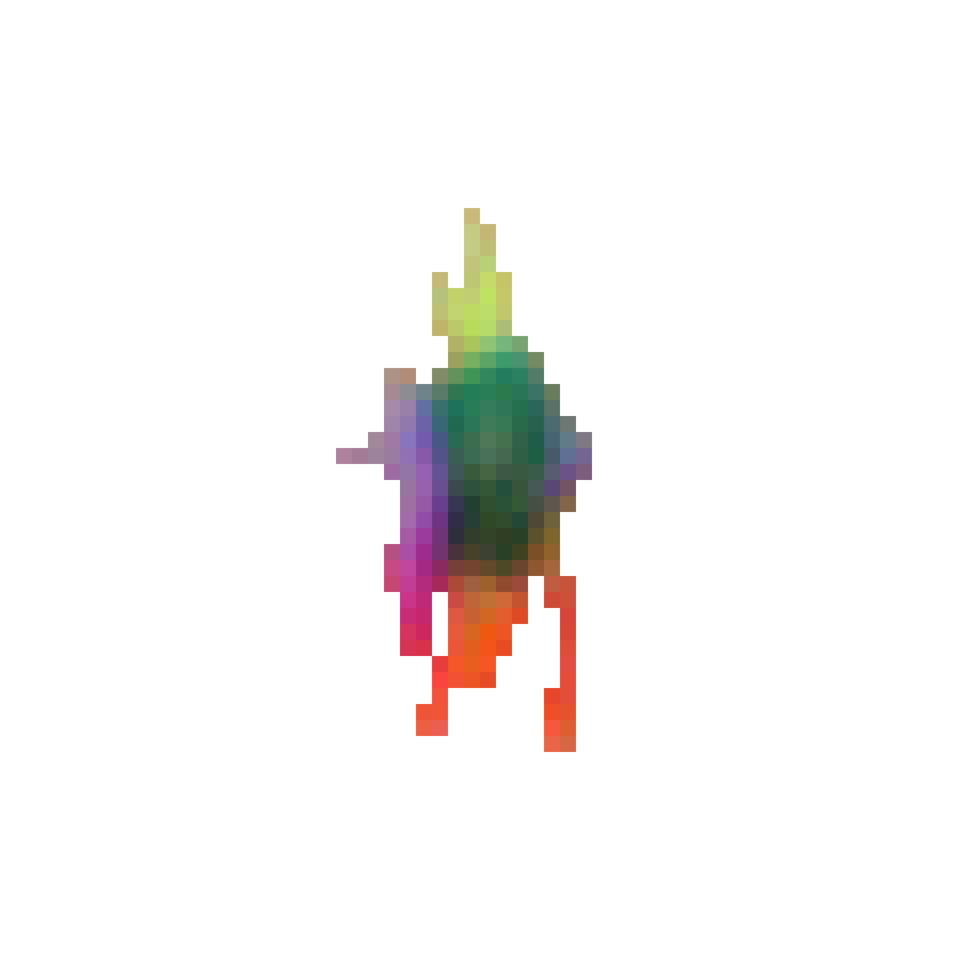}\hspace{0.5pt}%
    \zoomcrop{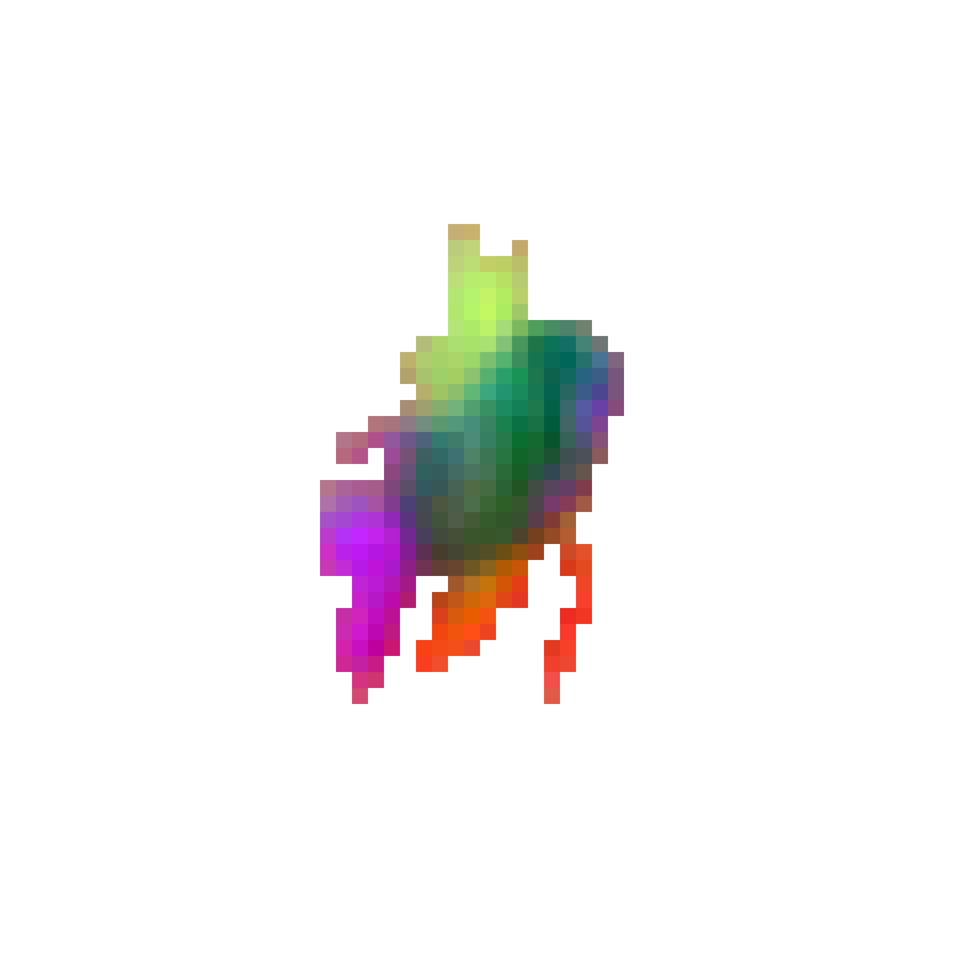}

    \vspace{4pt}

    \zoomcrop{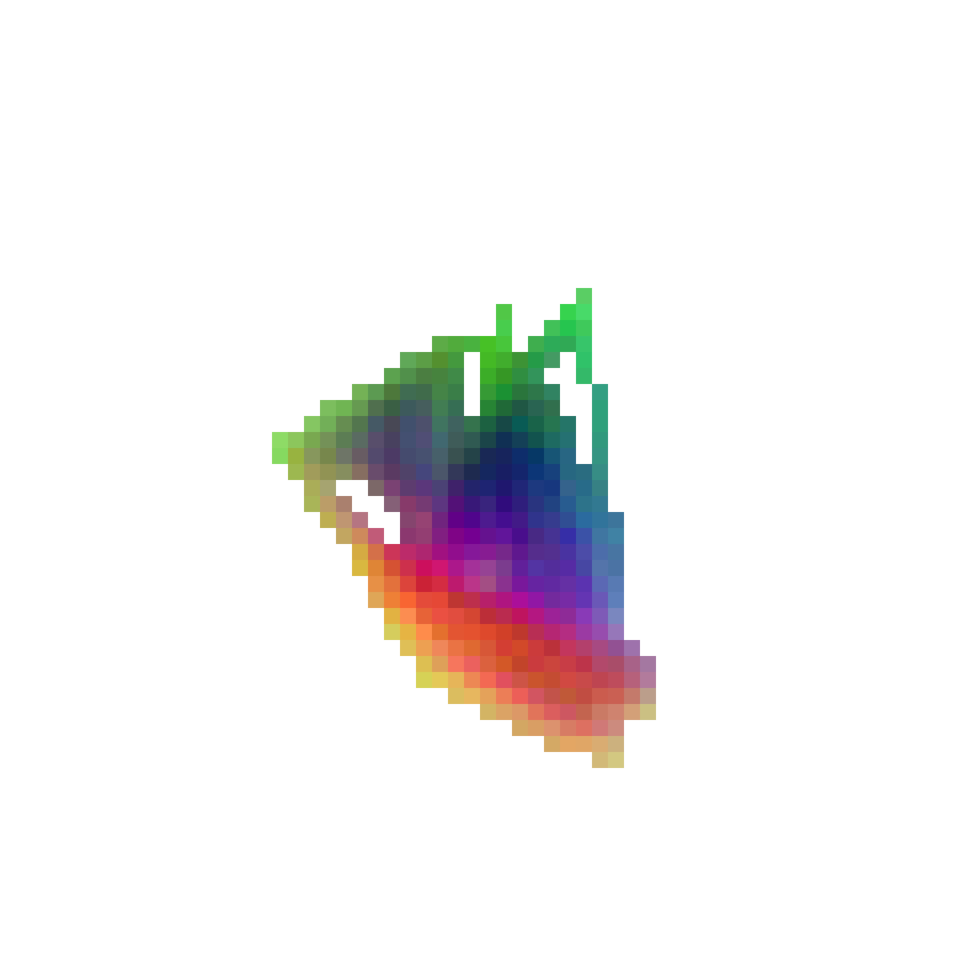}
    
    \zoomcrop{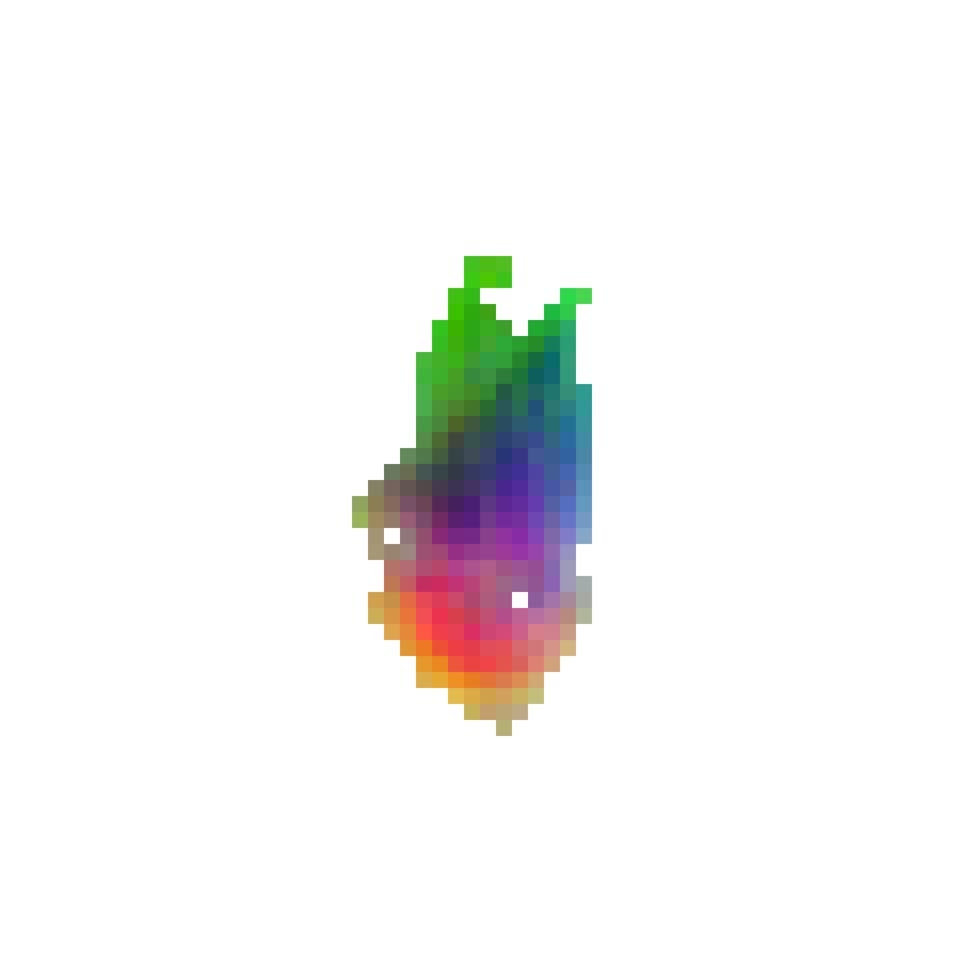}\hspace{0.5pt}%
    \zoomcrop{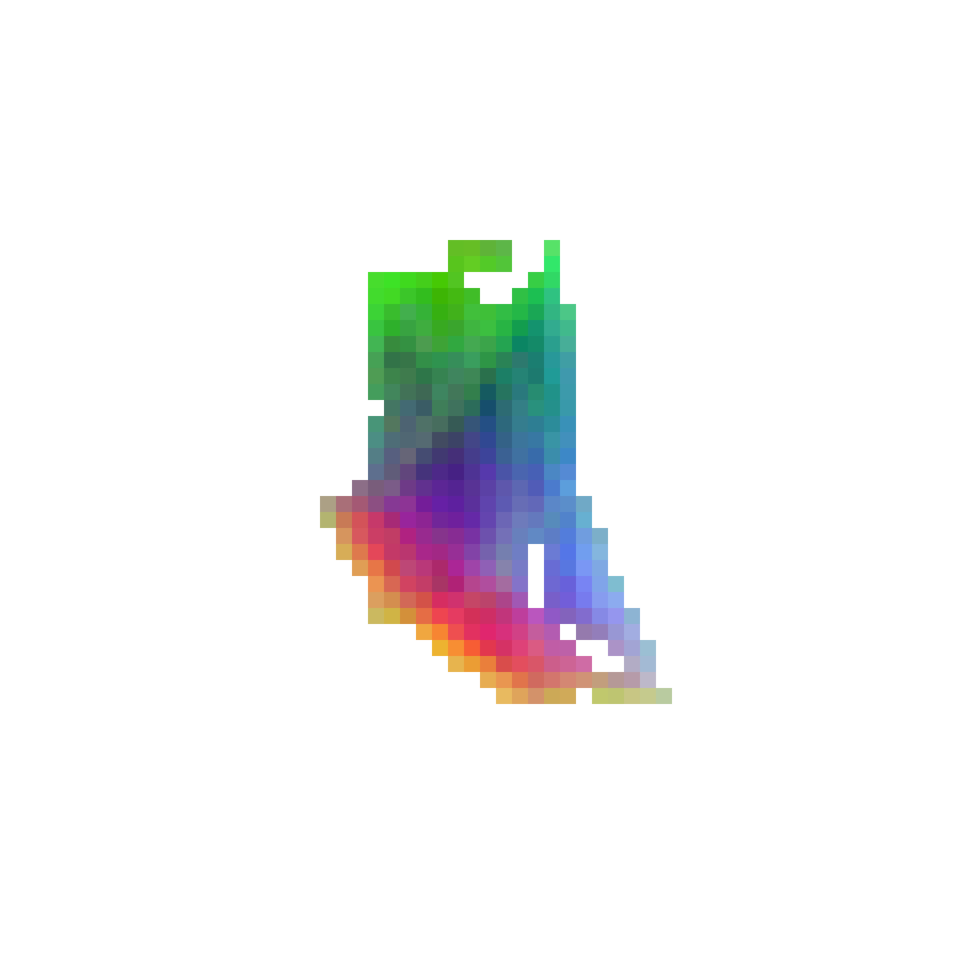}

    \vspace{4pt}
    
    \zoomcrop{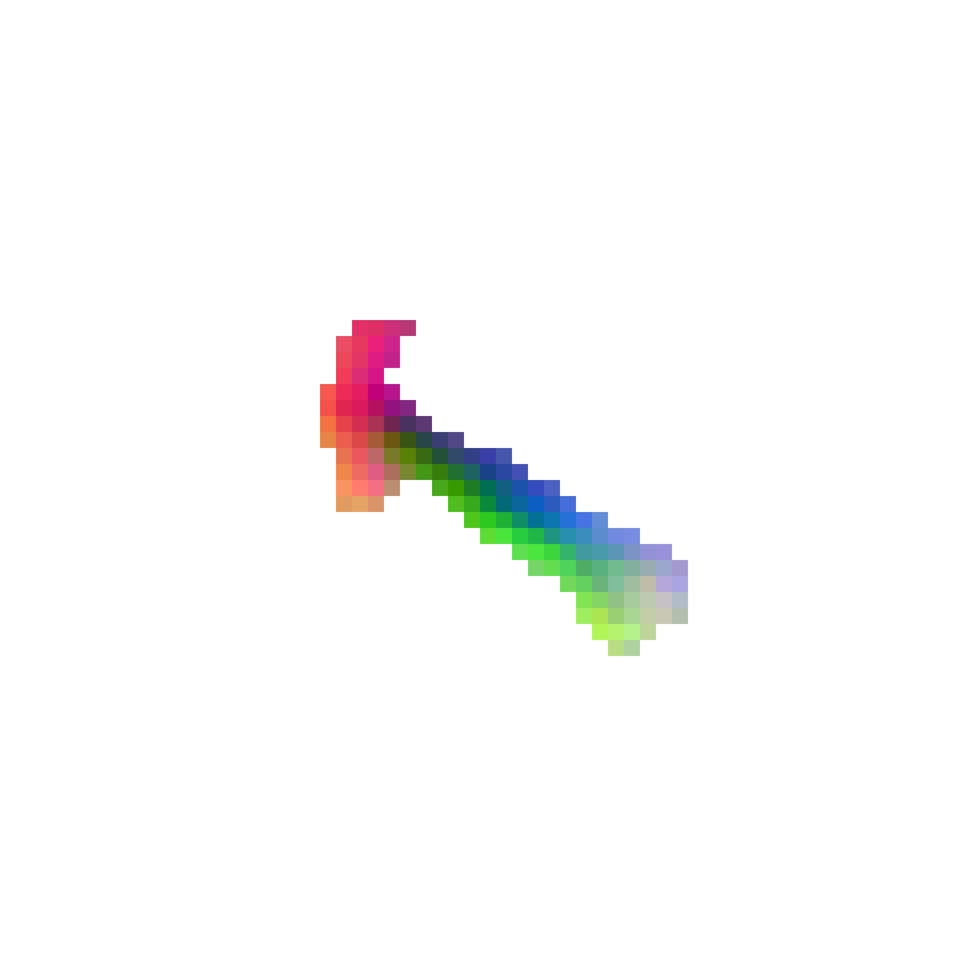}
    \zoomcrop{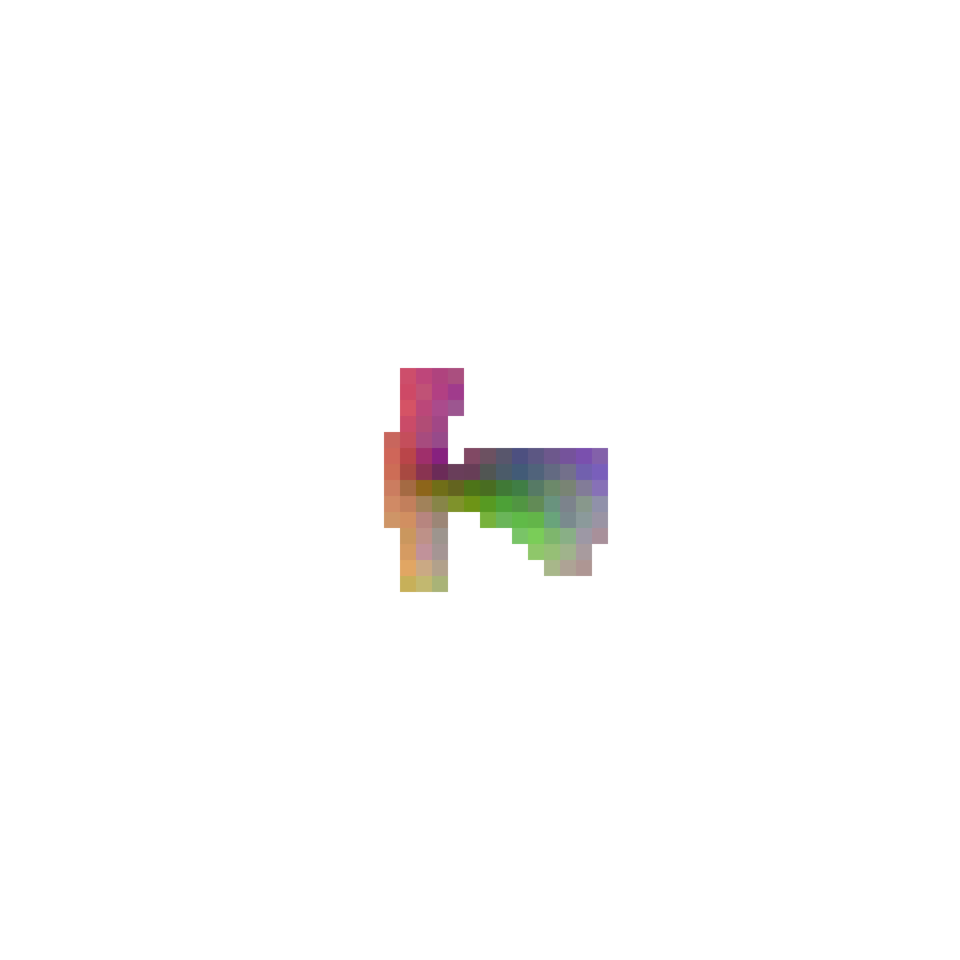}\hspace{0.5pt}%
    \zoomcrop{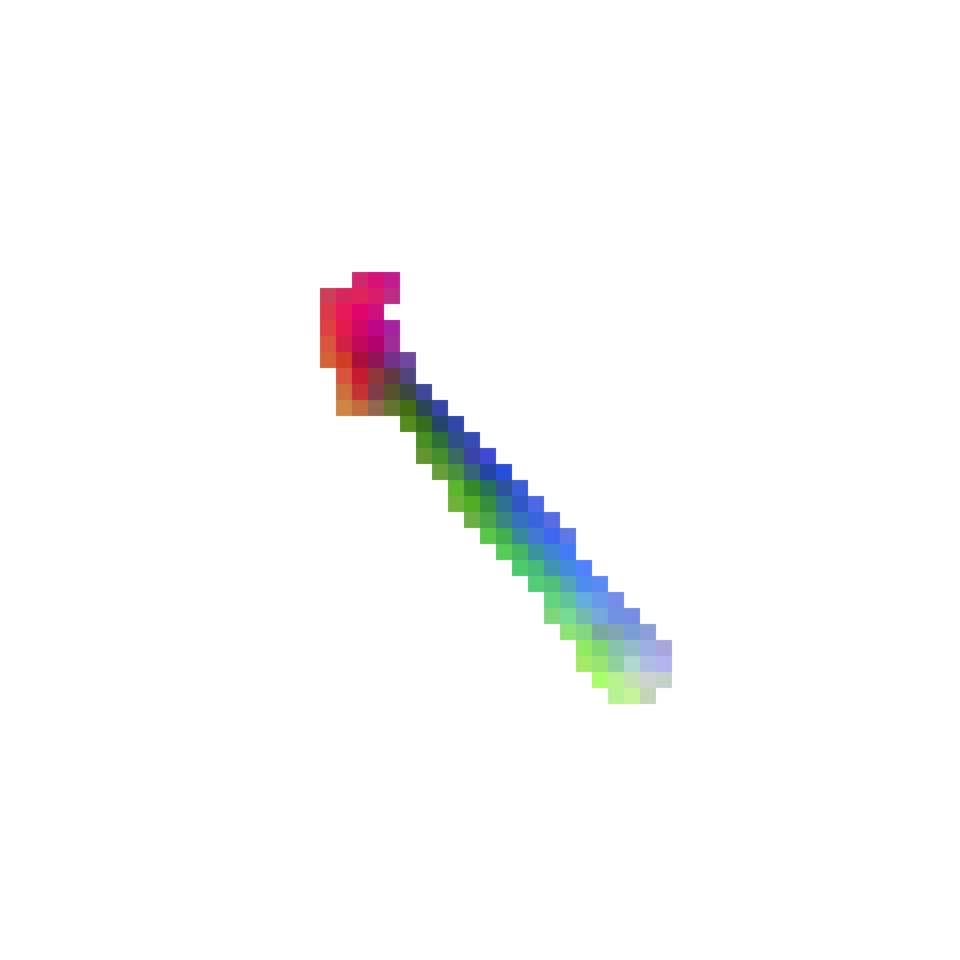}
    \captionsetup{skip=4pt}
    \caption{Ours}
  \end{subfigure}
  \vspace{-0.75em}
  \caption{
  \textbf{Feature visualization.} PCA visualization of learned features on randomly selected 3D objects from Objaverse~\cite{deitke2023objaverse}. Compared to (b) CLIP~\cite{radford2021learning_CLIP}, (c) FiT3D~\cite{yue2024improving_FiT3D}, and (d) MEF~\cite{you2024multiview_ME}, our method (e) not only generates consistently smoother and more coherent features with reduced noise but also accurately preserves semantic correspondences across multiple viewpoints.
  }
  \label{fig:appendix_quali_pca}
  \vspace{-0.75em}
  
\end{figure*}

\subsection{Model Size and Budget}
We utilize the CLIP~\cite{radford2021learning_CLIP} ViT-B/16 model as our vision-language backbone, which contains approximately 93 million parameters, closely comparable to the vanilla CLIP with about 87 million parameters. 
For parameter-efficient fine-tuning, we employ the Low-Rank Adaptation (LoRA)~\cite{hu2022lora} technique, which takes up roughly 6 million parameters (about 6.5\% of the total). All experiments are conducted on up to four NVIDIA A6000 GPUs, and our geometric distillation process takes approximately 1 hour and 20 minutes per model on a single NVIDIA A6000 GPU. Compared to prior methods such as FiT3D, which require up to three days of training on four A6000 GPUs due to costly optimizing 3D feature Gaussians for all training scene, our method significantly reduces computational cost while achieving superior performance.

\subsection{Experimental Setup and Hyperparameters}
We use the AdamW optimizer~\cite{loshchilov2017decoupled_adamw} with a learning rate of $1\times10^{-5}$, and a train LoRA for up to 500 training epochs with early-stopping across all experiments. LoRA adapters with rank $r=4$ are applied to intermediate self attention layers in the CLIP model baseline. For the relative depth supervision, we add four LoRA layers to the 4th-7th attention layers, along with adapters following~\cite{chen2022adaptformer}. The loss components are equally weighted: $\lambda_{\text{match}} = 1.0$, $\lambda_{\text{depth}} = 1.0$, and $\lambda_{\text{cost}} = 1.0$. Additionally, we apply temperature annealing to the cost volume alignment loss $\mathcal{L}_{\text{cost}}$ as described in~\Cref{eq:softmax_annealing_cost}, linearly decreasing $\tau$ from 1.0 to 0.5 during training. These hyperparameters were selected based on empirical tuning on ScanNet++ validation split and held consistent across all datasets to ensure fair comparison. We did not perform extensive hyperparameter search, and observed no significant sensitivity to small variations. 

For view sampling during geometric distillation on ScanNet++, we randomly sample 10,000 views across 100 scenes, then subsequently select 100 random pairs of views that share overlapping 3D regions. This sampling results in a dataset size equivalent to the Objaverse view pairs used in MEF~\cite{you2024multiview_ME}.

\subsection{Descriptive Statistics}
All results reported in the main paper and appendix represent the mean values over the full test set. For classification and tracking tasks, we use metrics such as PCK, Jaccard index, and positional accuracy at multiple thresholds. For depth estimation and semantic segmentation, we report RMSE, relative error, mIoU, and mAcc. We do not report error bars or variances, but all evaluations are deterministic and based on a single run unless otherwise specified. Our results are comparable to prior works under the same evaluation protocols and dataset splits.

\subsection{Parameters for Packages}
We rely on several well-established libraries and packages throughout our pipeline. For model implementation and training, we use PyTorch along with the HuggingFace Transformers~\cite{wolf2019huggingface} and PEFT (Parameter Efficient Fine Tuning) libraries to incorporate LoRA into the CLIP backbone. For vision tasks such as depth estimation and segmentation, we use torchvision and mmsegmentation-based tools for data pre-processing and evaluation. NLP evaluation metrics including BLEU, ROUGE, METEOR, and CIDEr are computed using standard implementations from the NLTK and COCOEval toolkits. All packages are used with default parameters unless otherwise specified. No additional tuning or modification was made to external evaluation functions.

\begin{figure*}
    \centering
    \includegraphics[width=0.95\linewidth]{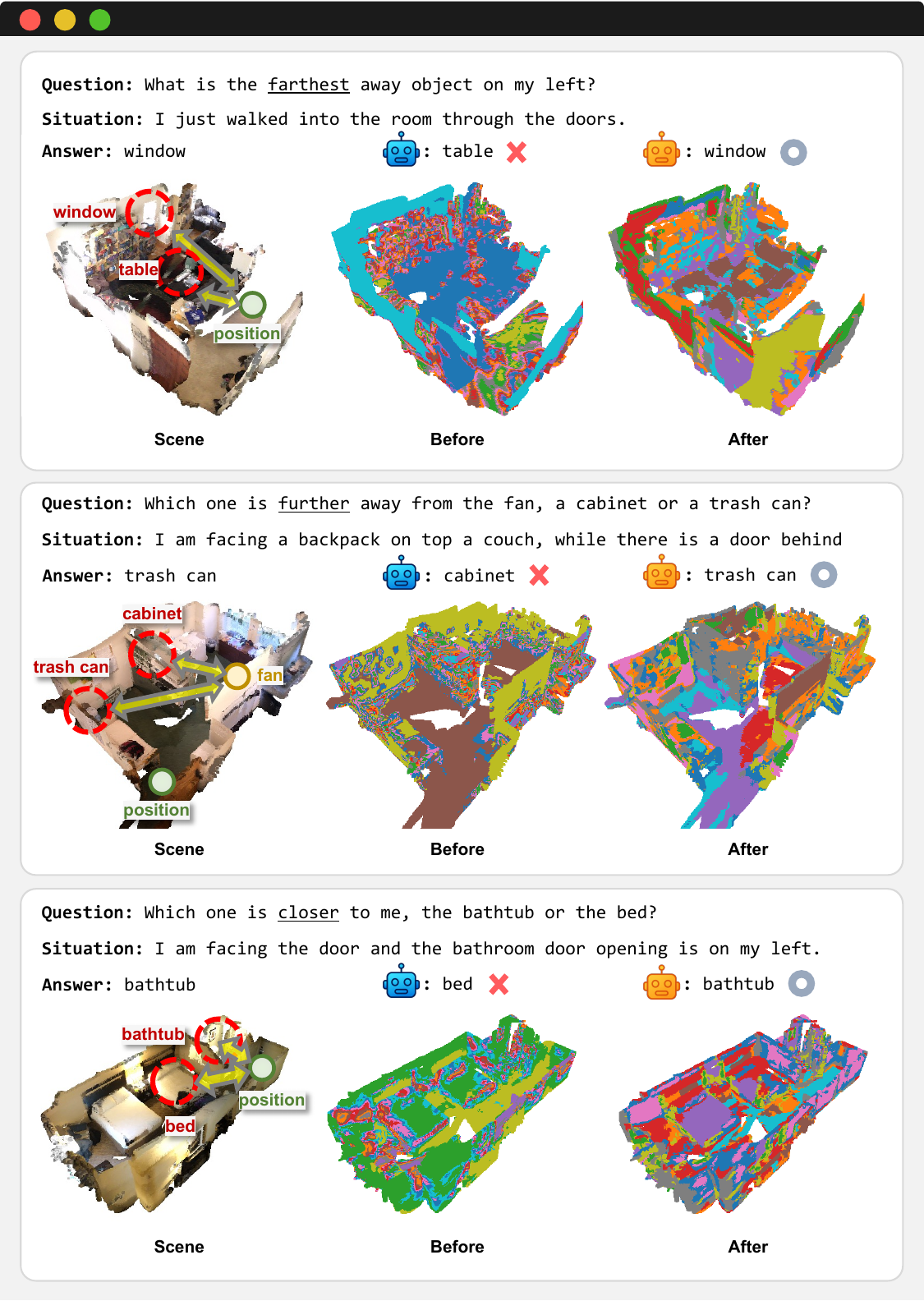}
    \caption{
       \textbf{Qualitative examples of 3D VQA on SQA3D.} Visualization of feature clustering for 3D scenes before and after our geometric distillation, following the protocol of Lexicon3D~\cite{man2024lexicon3d}. The 2D CLIP features and fine-tuned 2D CLIP features are lifted into 3D space and clustered using k-means. Each example presents a challenging VQA scenario, asking about relative object positions (e.g., ``farthest,'' ``further,'' ``closer''). Compared to vanilla CLIP (``Before''), our distilled features (``After'') offer clearer 3D spatial distinction and improved vision-language understanding for given 3D scenes.
    }
    \label{fig:appendix_3dvqa_examples}
    \vspace{-0.5em}
\end{figure*}

\section{AI Assistants In Research Or Writing}

\subsection{Information About Use Of AI Assistants}

We acknowledge the use of ChatGPT-4o~\cite{achiam2023gpt} for grammatical correction and style improvement during the writing of this paper.
However, all technical content, experiment design, and conceptual development were performed solely by the authors. No AI-generated content was used for core research contributions or evaluations.


\newcommand{\zoomwidecroptwo}[1]{%
  \includegraphics[width=\linewidth,
    trim=0pt 0pt 0pt 0pt,clip]{#1}}

\begin{figure*}[t!]
  \centering
  \begin{minipage}[t]{0.98\textwidth}
    \centering
    \subcaptionbox{Ground Truth}[0.322\linewidth]{%
      \zoomwidecroptwo{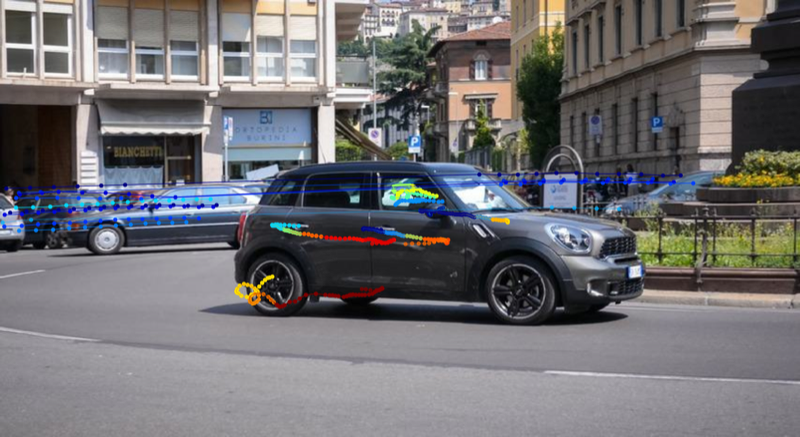}\par
      \vspace{2pt}
      \zoomwidecroptwo{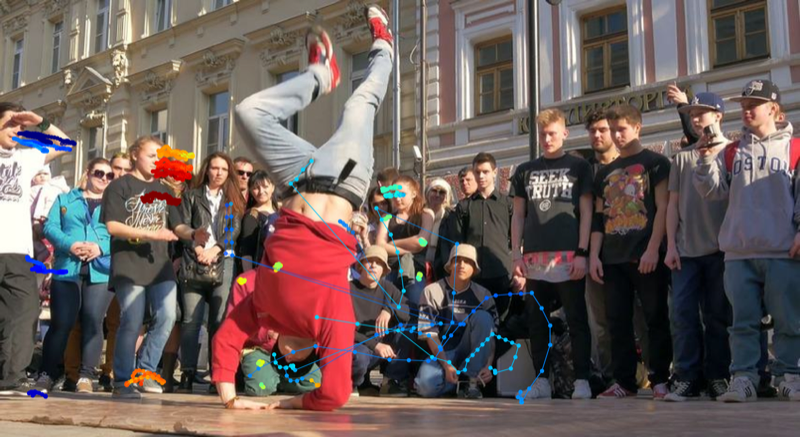}

      \vspace{2pt}
      \zoomwidecroptwo{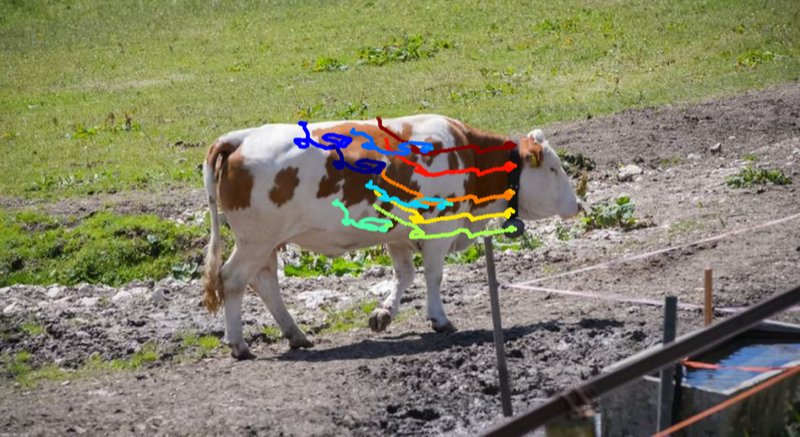}

      \vspace{2pt}
      \zoomwidecroptwo{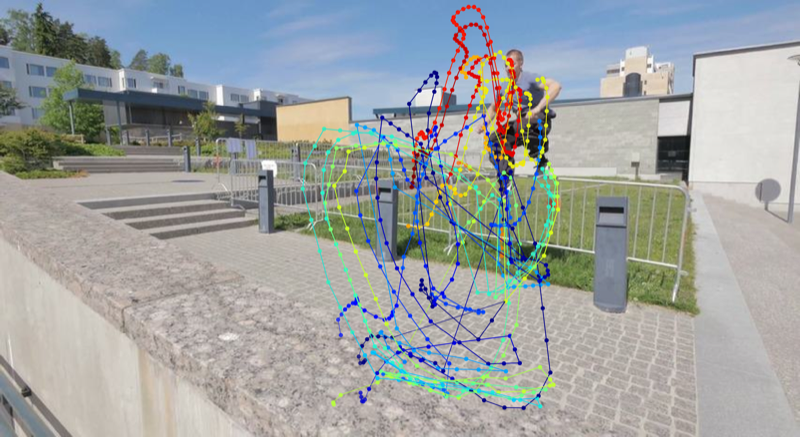}
      }
    \hspace{0.25pt}%
    \subcaptionbox{MEF}[0.322\linewidth]{%
      \zoomwidecroptwo{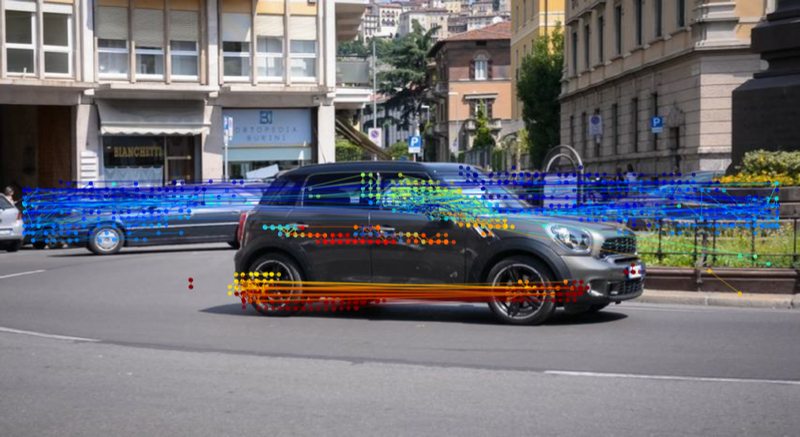}\par
      \vspace{2pt}
      \zoomwidecroptwo{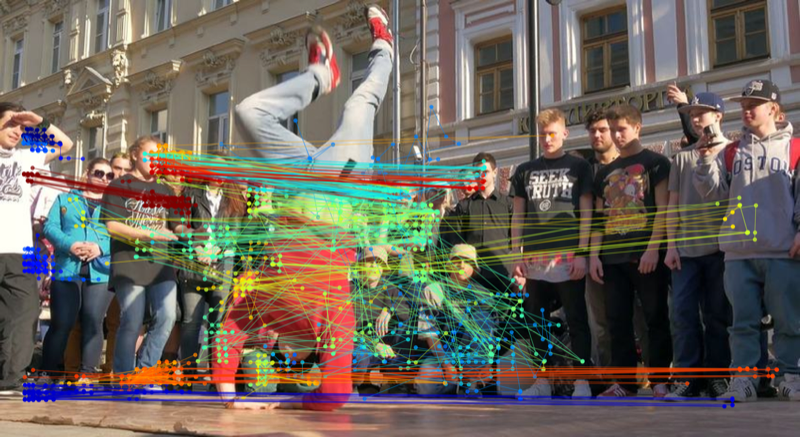}
      
      \vspace{2pt}
      \zoomwidecroptwo{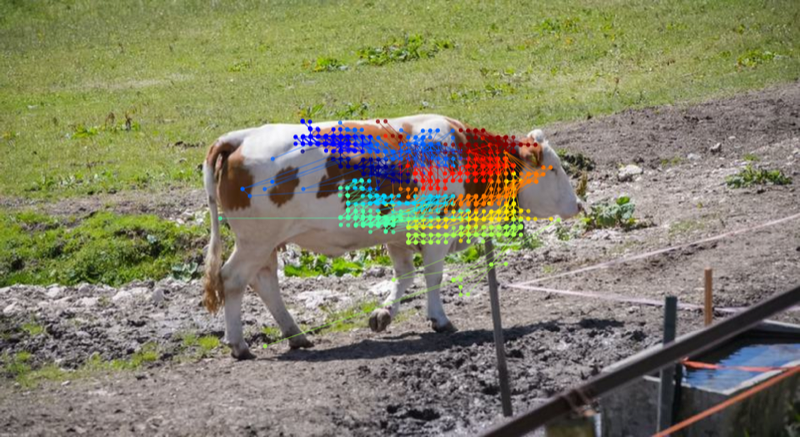}

      \vspace{2pt}
      \zoomwidecroptwo{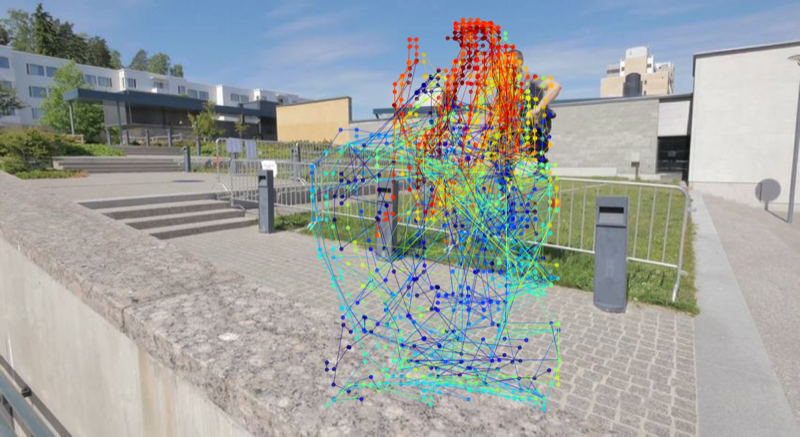}
    }
    \hspace{0.25pt}%
    \subcaptionbox{Ours}[0.322\linewidth]{%
      \zoomwidecroptwo{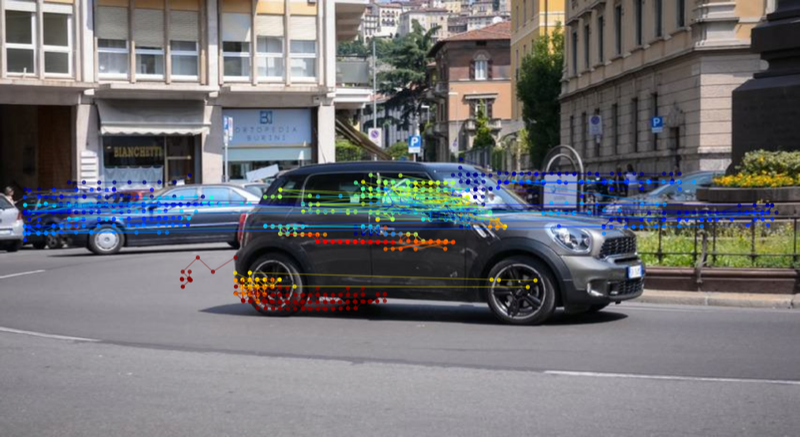}\par
      \vspace{2pt}
      \zoomwidecroptwo{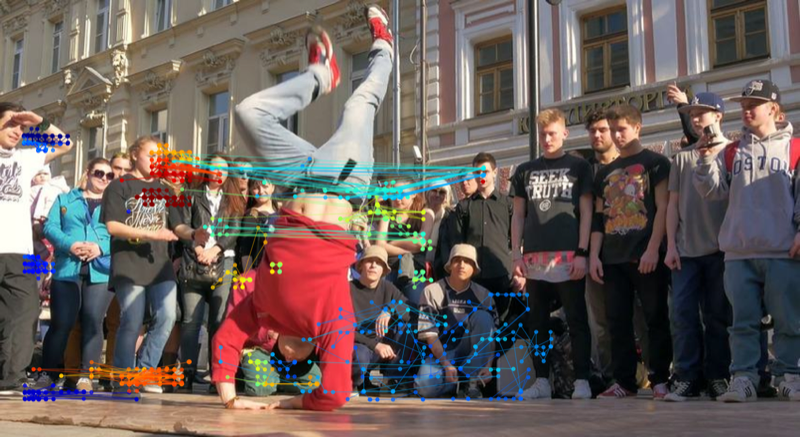}

      \vspace{2pt}
      \zoomwidecroptwo{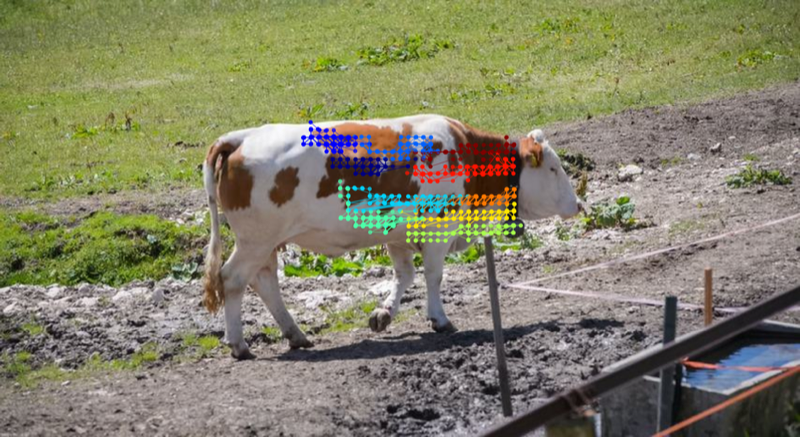}

      \vspace{2pt}
      \zoomwidecroptwo{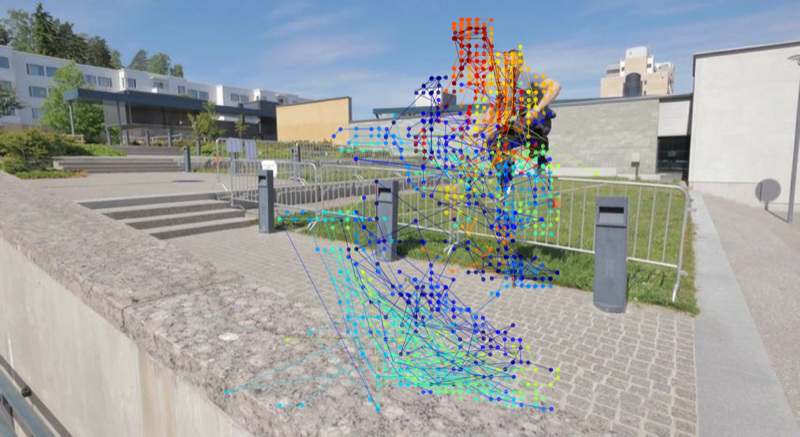}
    }
    \vspace{-0.6em}
        \caption{
        \textbf{Additional qualitative results on video tracking.} Visualization of predicted trajectories compared to (a) ground truth, (b) MEF~\cite{you2024multiview_ME}, and (c) ours. Our method provides more accurate and coherent object tracking, which significantly reduces incorrect correspondences and aligns better with ground-truth trajectories.
    }
    \label{fig:appendix_quali_video_tracking}
    \vspace{-0.5em}
  \end{minipage}
\end{figure*}

\section{Additional Qualitative Evaluation}

\subsection{Feature Visualization} 

To qualitatively analyze the effectiveness of our geometric distillation, we visualize PCA projections of features extracted from randomly sampled 3D objects in Objaverse. We compute a PCA between the patches of the images from the multi-view images of the same object and visualize their first 3 components. As illustrated in~\Cref{fig:appendix_quali_pca}, existing methods such as vanilla CLIP, FiT3D, and MEF produce noisy or inconsistent feature distributions across multiple views. In contrast, our method generates significantly smoother and more coherent feature maps that consistently preserve semantic correspondence across various viewpoints. This visualization confirms that our approach successfully injects robust multi-view geometric consistency into VLM features, which enables precise and noise-less representation of object parts and their spatial relationships.

\subsection{More Qualitative Results} 

We provide additional qualitative comparisons for video tracking performance on the TAP-Vid-DAVIS dataset (described in~\Cref{subsec:3d_visual_eval}) in~\Cref{fig:appendix_quali_video_tracking}. Compared to MEF~\cite{you2024multiview_ME}, our method produces notably cleaner and more accurate tracking results, which closely align with the ground-truth trajectories. Specifically, in the first row of~\Cref{fig:appendix_quali_video_tracking}, MEF struggles to accurately track the trajectory of the rear wheel, confusing it with the front wheel of the car. In contrast, our approach clearly distinguishes and consistently tracks object parts. These results show that our method effectively enhances consistency to viewpoint changes and object motion.

\subsection{Example Results of 3D VQA} 

As summarized in~\Cref{fig:appendix_3dvqa_examples}. we provide example results of 3D visual question answering 
evaluation on the SQA3D dataset following~\Cref{subsec:exp_3dvqa},. Specifically, we visualize features from vanilla CLIP and our fine-tuned CLIP obtained through geometric distillation. For visualization, we first lift the 2D CLIP features into their corresponding 3D scenes and apply k-means clustering. Our distilled features demonstrate clearer spatial coherence and improved geometric consistency compared to vanilla CLIP features. Consequently, our model exhibits superior spatial reasoning capabilities, which accurately identify relative object distances as required by challenging VQA questions, especially determining which object is farther or closer. For instance, while vanilla CLIP incorrectly identifies spatial relationships due to ambiguous feature representations, our method correctly interprets the precise spatial context, including spatially complex questions.

\begin{table*}[t!]
    \centering
    \caption{
        Absolute vs.\ relative depth loss in 3D correspondence understanding after fine-tuning on ScanNet++.
    }
    \label{tab:appendix_abs_depth}
    \vspace{-0.5em}
    \small
    \resizebox{\textwidth}{!}{
    \begin{threeparttable}
    \begin{tabular}{c|cccccc|cc|ccc}
    \toprule
    \multirow{3}{*}{Method} &
    \multicolumn{6}{c|}{\textbf{Semantic Correspondence}} &
    \multicolumn{2}{c|}{\textbf{Video Tracking}} &
    \multicolumn{3}{c}{\textbf{Pose Estimation}} \\
    \cmidrule(lr){2-7}\cmidrule(lr){8-9}\cmidrule(lr){10-12}
      & \multicolumn{3}{c}{Different Views} &
        \multicolumn{3}{c|}{Same Views} &
        \multirow{2}{*}{Jacc.} & \multirow{2}{*}{Avg.\ Pts} &
        \multicolumn{3}{c}{Thresholds} \\ 
    \cmidrule(lr){2-4}\cmidrule(lr){5-7}\cmidrule(lr){10-12}
      & 0.05 & 0.10 & 0.15 & 0.05 & 0.10 & 0.15 &  &  &
        1cm–1deg & 3cm–3deg & 5cm–5deg \\
    \midrule
    Abs. &
      27.04 & 41.33 & 50.37 & 37.45 & 57.63 & 66.58 &
      39.27 & 57.27 &
      9.46 & 42.04 & 60.93 \\
    \textbf{Rel.} &
      \textbf{28.48} & \textbf{43.07} & \textbf{53.55} &
      \textbf{42.16} & \textbf{61.57} & \textbf{72.16} &
      \textbf{40.09} & \textbf{57.75} &
      \textbf{10.96} & \textbf{44.93} & \textbf{63.65} \\
    \bottomrule
    \end{tabular}
    \end{threeparttable}}
\end{table*}

\section{Additional Ablation Study}

\subsection{Comparison of Absolute and 
Relative Depth Understanding}

We perform an additional analysis comparing the effects of absolute and relative depth losses on 3D correspondence understanding. Specifically, we fine-tune models on ScanNet++ using either absolute depth loss or our proposed relative depth loss, and evaluate them across the 3D correspondence tasks described in~\Cref{subsec:3d_visual_eval}. For absolute depth loss, we implement log-scale depth regression, which directly predicts depth values.
Given predicted depth $\hat{d}_p$ and ground-truth depth $\tilde{d}_p$ at keypoint $p$ for a single view, the absolute depth loss $\mathcal{L}_{\texttt{abs\_depth}}$ is computed as:
{\vspace{-1.0em}
    \begin{equation}   
    \mathcal{L}_{\texttt{abs\_depth}} = \frac{1}{|\mathcal{P}|}\sum_{p \in \mathcal{P}} |\hat{d}_p - s \cdot \tilde{d}_p|,
    \quad s = \frac{D_{\texttt{max}}^{\texttt{pred}}}{D_{\texttt{max}}^{\texttt{gt}}}
\end{equation}
}
where $D_{\texttt{max}}^{\texttt{pred}}$ and $D_{\texttt{max}}^{\texttt{gt}}$ denote the maximum depth from predictions and ground-truth, respectively, and $s$ is the scale factor ensuring that predictions match the range of the ground-truth.

As shown in~\Cref{tab:appendix_abs_depth}, the relative depth loss consistently outperforms absolute depth across all metrics. For semantic correspondence, it significantly improves PCK@0.05 from 27.04\% to 28.48\% (different views) and from 37.45\% to 42.16\% (same views). Similarly, relative depth supervision enhances video tracking, increasing the average Jaccard index from 39.27\% to 40.09\%, and boosts precise pose estimation accuracy at the 1cm–1deg threshold from 9.46\% to 10.96\%.

These results indicate that explicitly modeling relative depth relationships, rather than absolute depth values, yields more generalizable geometric representations. Additionally, it reduces the risk of overfitting to the depth distribution of the training dataset.

\begin{table*}[t!]
  \centering
    \caption{Ablation study of loss components on 3D correspondence understanding after finetuning on ScanNet++.}
    \vspace{-1.0em}
  \label{tab:appendix_ablation_scannetpp}
  \small
  \resizebox{\textwidth}{!}{
  \begin{threeparttable}
  \begin{tabular}{ccc|cccccc|cc|ccc}
    \toprule
    \multicolumn{3}{c|}{\textbf{Loss Components}} &
    \multicolumn{6}{c|}{\textbf{Semantic Correspondence}} &
    \multicolumn{2}{c|}{\textbf{Video Tracking}} &
    \multicolumn{3}{c}{\textbf{Pose Estimation}} \\
    \cmidrule(lr){1-3}\cmidrule(lr){4-9}\cmidrule(lr){10-11}\cmidrule(lr){12-14}
    \multirow{2}{*}{$\mathcal{L}_{\texttt{match}}$} &
    \multirow{2}{*}{$\mathcal{L}_{\texttt{depth}}$} &
    \multirow{2}{*}{$\mathcal{L}_{\texttt{cost}}$} &
      \multicolumn{3}{c}{Different Views} &
      \multicolumn{3}{c|}{Same Views} &
      \multirow{2}{*}{Jaccard} & \multirow{2}{*}{Avg.\ Pts} &
      \multicolumn{3}{c}{Accuracy within Thresholds} \\
    \cmidrule(lr){4-6}\cmidrule(lr){7-9}\cmidrule(lr){12-14}
      & & & 0.05 & 0.10 & 0.15 & 0.05 & 0.10 & 0.15 & & & 
      1cm–1deg & 3cm–3deg & 5cm–5deg \\
    \midrule
        \yesmark & \nomark & \nomark  & 26.32 & 41.76 & 50.72 & 37.45 & 58.30 & 68.15 & 37.78 & \underline{57.45} & 9.61 & 44.77 & 63.52 \\
     \yesmark & \yesmark &\nomark & \underline{27.25} & \textbf{43.43} & \underline{52.18} & \underline{38.82} & \underline{60.20} & \underline{69.64} & \underline{38.26} & 56.43 & \underline{10.80} & \textbf{47.40} & \textbf{64.93} \\
     \yesmark &  \yesmark &  \yesmark &  \textbf{28.48} & \underline{43.07} & \textbf{53.55} & \textbf{42.16} & \textbf{61.57} & \textbf{72.16} & \textbf{40.09} & \textbf{57.75} & \textbf{10.96} & \underline{44.93} & \underline{63.65} \\
    \bottomrule
  \end{tabular}
  \end{threeparttable}}

\end{table*}

\begin{table*}[t!]
  \centering
  \caption{Comparison of our VGGT and MASt3R-based methods on 3D correspondence understanding. }
  \label{tab:appendix_vggt_mast3r}
\vspace{-1.0em}
  \small
  \resizebox{\textwidth}{!}{
  \begin{threeparttable}
  \begin{tabular}{lcc|cccccc|cc|ccc}
    \toprule
    \multicolumn{3}{c|}{Model} &
    \multicolumn{6}{c|}{\textbf{Semantic Correspondence}} &
    \multicolumn{2}{c|}{\textbf{Video Tracking}} &
    \multicolumn{3}{c}{\textbf{Pose Estimation}} \\ 
    \cmidrule(lr){1-3}\cmidrule(lr){4-9}\cmidrule(lr){10-11}\cmidrule(lr){12-14}
        \multirow{2}{*}{Method} & \multirow{2}{*}{Teacher} & \multirow{2}{*}{Dataset} &
        \multicolumn{3}{c}{Different Views} &
        \multicolumn{3}{c|}{Same Views} &
        \multirow{2}{*}{Jaccard} & \multirow{2}{*}{Pos.\ Acc.} &
        \multirow{2}{*}{1cm-1deg} & \multirow{2}{*}{3cm-3deg} & \multirow{2}{*}{5cm-5deg} \\
        \cmidrule(lr){4-6}\cmidrule(lr){7-9}
      & & & 0.05 & 0.10 & 0.15 & 0.05 & 0.10 & 0.15 & & &\\
    \midrule
    CLIP (Vanilla) & — & — &
      16.61 & 26.96 & 37.64 & 18.23 & 32.27 & 43.01 &
      27.73 & 42.59 &
      2.50 & 19.32 & 33.11 \\
    \multirow{2}{*}{Ours (VGGT)} 
      & \multirow{2}{*}{VGGT} & Objaverse &
        19.84 & 32.79 & 44.24 & 25.44 & 42.48 & 55.18 &
        36.77 & 52.68 &
        6.94 & 34.37 & 51.83 \\[2pt]
      & & ScanNet++ &
        24.22 & 39.52 & 48.34 &
        30.79 & 53.03 & 63.26 &
        \underline{37.28} & 54.22 &
        8.15 & 38.75 & 57.55 \\
    \multirow{2}{*}{Ours (MASt3R)} 
      & \multirow{2}{*}{MASt3R} & Objaverse &
        \underline{25.87} & \underline{39.85} & \underline{50.21} &
        \underline{36.77} & \underline{56.61} & \underline{67.93} &
        35.60 & \underline{54.65} &
        \underline{8.50} & \underline{39.30} & \underline{57.68} \\[2pt]
      & & ScanNet++ &
        \textbf{28.48} & \textbf{43.07} & \textbf{53.55} &
        \textbf{42.16} & \textbf{61.57} & \textbf{72.16} &
        \textbf{40.09} & \textbf{57.75} &
        \textbf{10.96} & \textbf{44.93} & \textbf{63.65} \\
    \bottomrule
  \end{tabular}
  \end{threeparttable}
}
\vspace{-1.0em}
\end{table*}

\subsection{Ablation on Loss Components with Different Training Dataset}

To further investigate the generalization of each loss component in our geometric distillation, we conduct an additional ablation study by fine-tuning on the real-world ScanNet++ dataset, complementing our earlier analysis performed on Objaverse as in~\Cref{subsec:ablation_study}). Specifically, we evaluate the effects of the matching loss $\mathcal{L}_{\texttt{match}}$, relative depth loss $\mathcal{L}_{\texttt{depth}}$, and cost volume alignment loss $\mathcal{L}_{\texttt{cost}}$ across the downstream 3D correspondence tasks described in~\Cref{subsec:3d_visual_eval}.

As shown in~\Cref{tab:appendix_ablation_scannetpp}, adding the relative depth loss $\mathcal{L}_{\texttt{depth}}$ significantly enhances semantic correspondence, increasing PCK@0.10 from 41.76\% to 43.43\% (different views), and improving pose estimation accuracy at the strict 1cm–1deg threshold from 9.61\% to 10.80\%. Incorporating the cost volume alignment loss $\mathcal{L}_{\texttt{cost}}$ further strengthens performance, which yields substantial gains across most metrics. Specifically, semantic correspondence at PCK@0.05 notably increases from 26.32\% to 28.48\% (different views) and from 37.45\% to 42.16\% (same views). Additionally, video tracking accuracy measured by the average Jaccard index improves from 37.78\% to 40.09\%, and pose estimation achieves the highest accuracy of 10.96\% at 1cm–1deg threshold.

These results confirm that each loss component meaningfully contributes to enhancing cross-view consistency and spatial understanding. Particularly, the cost volume alignment loss $\mathcal{L}_{\texttt{cost}}$ improves the precision of representations, which significantly benefits performance on the most stringent evaluation metrics.

\subsection{Comparison of MASt3R and VGGT as a Teacher Model}

We conduct additional experiments to compare the effectiveness of different pretrained 3D foundation models, MASt3R and VGGT, used as teacher models in our geometric distillation method.
Specifically, we evaluate their performance across multiple downstream 3D correspondence tasks as summarized in~\Cref{tab:appendix_vggt_mast3r}.

Both MASt3R and VGGT-based models substantially outperform the vanilla CLIP baseline, and this demonstrates the effectiveness of our geometric distillation approach. However, we observe consistent differences between the two teachers. Overall, MASt3R consistently generates superior results compared to VGGT, particularly when fine-tuned on real-world ScanNet++ data. For example, on ScanNet++, MASt3R achieves significantly better semantic correspondence accuracy (PCK@0.05 of 28.48\% vs. 24.22\% in different-view scenarios and 42.16\% vs. 30.79\% in same-view scenarios), enhanced video tracking performance (average Jaccard index 40.09\% vs. 37.28\%), and improved pose estimation accuracy (10.96\% vs. 8.15\% at 1cm–1deg threshold). 

We attribute this difference in performance partly to the operational characteristics of each teacher model. Specifically, VGGT requires selecting an anchor viewpoint as user input to estimate dense correspondences across other views, so that it potentially introduces noise or inaccuracies. In contrast, MASt3R directly predicts dense and consistent semantic correspondences without requiring explicit selection of anchor points, which results in more reliable geometric guidance. Thus, while both models effectively enhance the geometric understanding of VLMs, MASt3R provides more precise and robust geometric priors in our experiments.

\section{Failure Cases}

Although our geometric distillation method significantly enhances the VLM representations, we identify limitations under certain challenging scenarios, also shared by MEF~\cite{you2024multiview_ME}. Specifically, our approach heavily relies on accurate geometric priors from pretrained 3D foundation models. Consequently, when input views have minimal or no overlapping 3D regions, these foundation models may fail to accurately infer or reconstruct the underlying geometry. Such failures can propagate erroneous geometric guidance into our distilled VLM features, which may degrade its performance on downstream tasks. This limitation might be alleviated through improved sampling strategies that explicitly consider shared viewing regions, as well as by enhancing the single-image 3D inference capability of the underlying 3D foundation models.

We believe that addressing these limitations is an important future direction. Potential improvements may include utilizing more powerful 3D foundation models trained on diverse, large-scale multi-view datasets or integrating explicit uncertainty estimation to mitigate the impact of unreliable geometric guidance.




    



\end{document}